%% file: acl_latex.tex
\pdfoutput=1

\documentclass[11pt]{article}

\usepackage[preprint]{acl}

\usepackage{times}
\usepackage{latexsym}

\usepackage[T1]{fontenc}

\usepackage[utf8]{inputenc}

\usepackage{microtype}

\usepackage{inconsolata}

\usepackage{graphicx}

\usepackage{booktabs}

\usepackage{multirow}
\usepackage{booktabs}
\usepackage{colortbl}
\usepackage{xcolor}

\usepackage{adjustbox}
\usepackage{makecell}
\usepackage{tabularx}
\usepackage{amsfonts} 

\usepackage{amsmath}
\usepackage{subcaption}

\usepackage[utf8]{inputenc}
\usepackage{tcolorbox}
\usepackage{tabularx}
\usepackage{lipsum}
\usepackage{enumitem}
\usepackage{courier}

\title{Uncovering the Impact of Chain-of-Thought Reasoning for Direct Preference Optimization: Lessons from Text-to-SQL}

\author{
    \textbf{Hanbing Liu\textsuperscript{1}\thanks{Equal contribution.}},
    \textbf{Haoyang Li\textsuperscript{2,3}\footnotemark[1]},
    \textbf{Xiaokang Zhang\textsuperscript{2,3}},
    \textbf{Ruotong Chen\textsuperscript{2}}, \\ 
    \textbf{Haiyong Xu\textsuperscript{5}},
    \textbf{Tian Tian\textsuperscript{5}},
    \textbf{Qi Qi\textsuperscript{1}},
    \textbf{Jing Zhang\textsuperscript{2,4}}\thanks{Corresponding author.}
    \\
    \textsuperscript{1}Gaoling School of Artificial Intelligence, Renmin University of China, Beijing, China, \\
    \textsuperscript{2}School of Information, Renmin University of China, Beijing, China, \\
    \textsuperscript{3}Key Laboratory of Data Engineering and Knowledge Engineering, Beijing, China, \\
    \textsuperscript{4}Engineering Research Center of Database and Business Intelligence, Beijing, China, \\
    \textsuperscript{5}China Mobile Information Technology Center
    \\
    {\fontfamily{zi4}\selectfont\{liuhanbing, lihaoyang.cs, zhang-jing\}@ruc.edu.cn}
}

\usepackage{CJKutf8}
\begin{document}
\maketitle
\begin{abstract}
Direct Preference Optimization (DPO) has proven effective in complex reasoning tasks like math word problems and code generation. However, when applied to Text-to-SQL datasets, it often fails to improve performance and can even degrade it. Our investigation reveals the root cause: unlike math and code tasks, which naturally integrate Chain-of-Thought (CoT) reasoning with DPO, Text-to-SQL datasets typically include only final answers (gold SQL queries) without detailed CoT solutions. By augmenting Text-to-SQL datasets with synthetic CoT solutions, we achieve, for the first time, consistent and significant performance improvements using DPO.

Our analysis shows that CoT reasoning is crucial for unlocking DPO’s potential, as it mitigates reward hacking, strengthens discriminative capabilities, and improves scalability. These findings offer valuable insights for building more robust Text-to-SQL models. To support further research, we publicly release the code and CoT-enhanced datasets
~\footnote{\url{https://github.com/RUCKBReasoning/DPO_Text2SQL}}.




\end{abstract}

\section{Introduction}
Text-to-SQL has recently gained significant attention in natural language processing and database research~\citep{li2024codes, wang2020rat-sql, DBLP:journals/pvldb/FuLWLTS23catsql, pourreza2024din-sql}. It translates natural language questions into SQL queries, allowing non-experts to easily access data, making it a valuable tool for business intelligence, data exploration, and other data-centric applications.

With large language models (LLMs), two main approaches have emerged for solving Text-to-SQL: prompting-based methods~\citep{talaei2024chess, pourreza2024din-sql, pourreza2024chase-sql}  and supervised fine-tuning (SFT) methods~\citep{pourreza2024dts-sql, li2023resdsql, li2024codes}. Prompting-based methods often rely on powerful closed-source LLMs, making them costly and slow, and raising data privacy concerns. In contrast, SFT trains open-source, deployable LLMs using benchmark datasets like Spider and Bird. However, SFT performance is often limited by the scarcity of high-quality training data, which is expensive and time-consuming to create.

\begin{figure}[t!]
  \vspace{\baselineskip}
  \centering
  \includegraphics[width=0.98\linewidth]{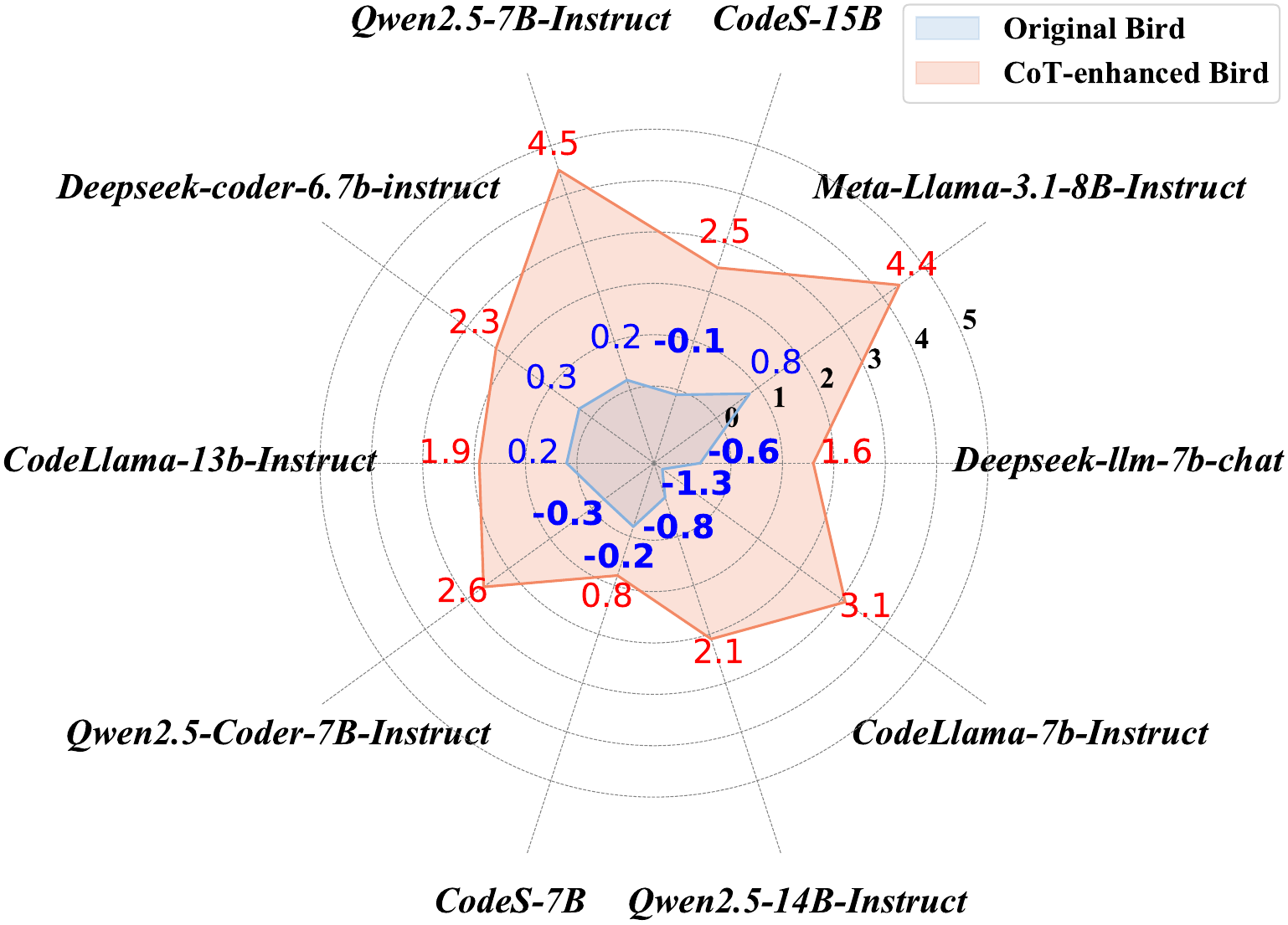}
  \caption{Model performance gains (greedy decoding) achieved by DPO over SFT (Improved Execution Accuracy, \%). Chain-of-thought reasoning is crucial for unlocking DPO's potential, ensuring its effectiveness and stability. }
  \label{fig:dpo_changes}
\end{figure}

Recent studies in complex reasoning tasks, such as math word problems~\cite{DBLP:conf/emnlp/XuLLHLZWZDZ0D24chatglm} and code generation~\cite{DBLP:journals/corr/abs-2406-06887plum}, have demonstrated that preference optimization algorithms (\emph{e.g.}, DPO~\cite{rafailov2024dpo}, KTO~\cite{DBLP:journals/corr/abs-2402-01306kto}, SimPO~\cite{DBLP:journals/corr/abs-2405-14734simpo}) can significantly enhance SFT models. These algorithms leverage preference data pairs to enable models to distinguish between correct and incorrect responses, addressing the limitations of simple SFT. Despite the proven success of preference optimization techniques, recent works in Text-to-SQL have rarely adopted these methods to improve the Text-to-SQL capabilities of LLMs. This raises a critical question: \textit{How much improvement can preference optimization bring to the Text-to-SQL task?}

\textbf{Preliminary Experiments.} To answer this question, we conduct initial experiments on Bird~\cite{li2024bird}, a challenging cross-domain Text-to-SQL benchmark. Each data sample consists of a <question, database, SQL query> triplet. Text-to-SQL models receive the question and database information (\emph{e.g.}, table names, column names, data types, etc.) and generate the target SQL query. To ensure the universality of our findings, we evaluate 10 open-source LLMs, ranging from 6.7B to 15B parameters. For preference optimization, we employ DPO, a widely adopted technique used in cutting-edge LLMs like LLaMA3~\cite{dubey2024llama3}, Qwen2.5~\cite{qwenreport}, and Mixtral~\cite{mixtral}.

Specifically, we follow the standard DPO training pipeline, which consists of three key steps:
(1) \textbf{SFT}: The base LLM is first fine-tuned on Bird's training set.
(2) \textbf{Preference Pair Construction}: Using the SFT model, multiple SQL queries are sampled for each training sample. Correct and incorrect queries are identified through database execution to create preference pairs.
(3) \textbf{DPO Training}: Finally, the SFT model is further trained on these preference pairs using the DPO loss, resulting in the final DPO model.

\textbf{Observations.} The ``Original Bird'' area in Figure~\ref{fig:dpo_changes} illustrates the performance gains introduced by DPO, measured as the improvement in execution accuracy between the DPO model and the SFT model with greedy search inference. Surprisingly, the results reveal that DPO does not consistently improve performance; in fact, it leads to performance degradation for 6 out of the 10 evaluated LLMs. To make preference optimization effective for Text-to-SQL, we additionally explore several strategies, including hyperparameter tuning~\cite{rafailov2024dpo}, integrating SFT loss~\cite{ouyang2022rlhf}, replacing DPO with KTO~\cite{DBLP:journals/corr/abs-2402-01306kto}, and using a small model to construct preference data~\cite{yang2024sense}. However, as shown in Appendix~\ref{apx:dpotricks}, these attempts still result in limited performance improvements ($<$1.5\%). 

\textbf{Hypothesis.} After extensive but unsuccessful algorithmic exploration, we hypothesize that the suboptimal performance of DPO in the Text-to-SQL task is primarily due to a critical yet often-overlooked factor: the quality of the data. By analyzing datasets for complex reasoning tasks, such as MATH~\cite{DBLP:conf/nips/HendrycksBKABTS21MathBenchmark}, GSM8K~\cite{DBLP:journals/corr/abs-2110-14168gsm8k}, CodeUltraFeedback~\cite{codeultrafeedback}, Orca-Math~\cite{orcamath}, and DART-Math~\cite{dartmath}, we observe that these datasets provide not only final answers but also chain-of-thought (CoT)-styled solutions with detailed reasoning steps. These CoT solutions bridge the gap between input questions and final answers, enabling LLMs to achieve better generalization and interpretability during SFT and DPO training. In contrast, Text-to-SQL datasets like Bird~\cite{li2024bird}, Spider~\cite{yu2018spider}, WikiSQL~\cite{wikisql}, and ScienceBenchmark~\cite{sciencebenchmark} only provide final answers (\emph{i.e.}, gold SQL queries), forcing SFT and DPO to rely solely on SQL queries as training labels. This discrepancy leads us to propose a hypothesis: \textit{The effectiveness of DPO is likely attributed to the use of CoT, a crucial factor that is often overlooked.}

\textbf{Verification.} 
To test this hypothesis, we introduce a pipeline to study how CoT affects DPO's performance in the Text-to-SQL task. We use an LLM-based CoT synthesizer to efficiently generate step-by-step CoT solutions for Text-to-SQL datasets with minimal human effort. The synthesizer takes the database information, question, and gold SQL query as input. Then, using the same settings as earlier experiments, we apply SFT and DPO to the CoT-enhanced Bird dataset. As shown in Figure~\ref{fig:dpo_changes}, adding CoT significantly improves DPO's performance across all 10 evaluated LLMs. Additionally, we extend our evaluations to other Text-to-SQL benchmarks, including Spider, Spider-DK~\cite{gan2021spiderdk}, Spider-Syn~\cite{gan2021spidersyn}, Spider-Realistic~\cite{deng2021spiderrealitic}, and Dr.Spider~\cite{DBLP:conf/iclr/Changdrspider}. Consistent trends are observed across these benchmarks.

\textbf{Analysis.} To understand why CoT reasoning is essential for unlocking DPO's potential, we conduct a comprehensive analysis and make three key observations. First, introducing CoT significantly reduces reward hacking during DPO training, ensuring stable and effective performance. Second, CoT enhances DPO's effectiveness as an implicit reward model, improving its ability to discriminate between correct and incorrect responses. Finally, CoT increases DPO's scalability, both in terms of the number of preference data and inference-time sampling budgets.

Our contributions are summarized as follows:
\begin{itemize}[leftmargin=1.0em, itemsep=0.1em, parsep=0em, topsep=0em]
    \item  We conduct extensive experiments on Text-to-SQL datasets to evaluate DPO within the standard training pipeline. Contrary to prior studies, we find that DPO does not consistently improve performance and can sometimes degrade it. However, by augmenting these datasets with synthetic CoT solutions, we achieve stable and significant performance improvements with DPO for the first time. As existing works overlook the critical data issue in Text-to-SQL, our findings provide important insights for effectively integrating DPO into Text-to-SQL pipelines.
    \item We also provide a comprehensive analysis to understand why CoT reasoning is essential for DPO. Our findings reveal that incorporating CoT mitigates reward hacking, strengthens discriminative ability, and enhances scalability.
\end{itemize}

\section{Related Work}

\input{RelatedWork}

\section{Pipeline}
\input{sections/Pipeline}

\section{Experiment Setup}
\subsection{Datasets}

\textbf{Common Benchmark:} Spider~\citep{yu2018spider} is a widely used Text-to-SQL dataset comprising a training set of 7,000 samples and a development set of 1,034 samples. This dataset covers 200 databases across 138 diverse domains.

\textbf{Challenging Benchmark:} Bird~\citep{li2024bird} presents a more challenging benchmark, featuring a training set of 9,428 samples and a development set of 1,534 samples. It includes 95 large databases across 37 professional domains. In contrast to Spider, Bird offers a more realistic scenario that aligns with real-world applications. 



\textbf{Robustness Benchmarks:} Spider-DK~\citep{gan2021spiderdk}, Spider-Syn~\citep{gan2021spidersyn}, and Spider-Realistic~\citep{deng2021spiderrealitic} are three widely adopted robustness evaluation sets that modify the development set of Spider to simulate real-world scenarios. Another significant derivative, Dr.Spider~\citep{DBLP:conf/iclr/Changdrspider}, creates 17 distinct robustness evaluation sets by comprehensively perturbing the Spider development set across 3 aspects: questions, databases, and SQL queries. 


\subsection{Evaluation Metrics}
For all benchmarks, we use the execution accuracy (EX) metric~\citep{yu2018spider} to evaluate the accuracy of the model's predictions. EX measures whether the predicted and gold SQL queries produce identical execution results on the given database. For Spider's development set, we additionally employ a more robust metric, test-suite accuracy (TS)~\citep{DBLP:conf/emnlp/ZhongYK20testsuite}, which extends EX by evaluating whether the predicted SQL query consistently passes the EX evaluation across multiple test-suite database instances.

\subsection{Inference Strategy}
Given a Text-to-SQL model, we explore three inference strategies: \textbf{(a) Greedy}: Use greedy decoding with a temperature of 0 to generate a response. \textbf{(b) Pass@1}: Sample a response with a temperature of 1.0. To ensure stability, we repeat this process 16 times and report the average scores. \textbf{(c) Maj@K}: Sample $K$ responses with a temperature of 1.0 and conduct majority voting based on the execution results of the predicted SQL queries. The final prediction is selected from the most-voted group.





\subsection{Implementation Details}
We select 10 base models from various model families, including Deepseek~\citep{bi2024deepseekllm, guo2024deepseekcoder}, Qwen~\citep{yang2024qwen, hui2024qwencoder}, Llama~\cite{dubey2024llama3, roziere2023codellama}, and CodeS~\citep{li2024codes}. These models cover different specialties (general-purpose, code- or SQL-specific) and range from 6.7B to 15B parameters. For each LLM, we conduct SFT and DPO using either the original training dataset (Vanilla) or the CoT-enhanced dataset (Syn CoT). More implementation details are listed in Appendix~\ref{apx:imp_details}. Details about training data can be found in Appendix~\ref{apx:datadetails}.


\section{Experimental Results}

\input{sections/MainResult}

\section{Why Does DPO Benefit From CoT?}

\input{sections/CoTHelpDPO}
\section{Prospects}
Our trained DPO model can be integrated into existing multi-agent Text-to-SQL frameworks. As a proof of concept, we adopt Qwen2.5-7B-Instruct as the base model and integrate it into two Text-to-SQL frameworks: DTS-SQL~\citep{pourreza2024dts-sql} and C3-SQL~\citep{dong2023c3}. Implementation details are provided in Appendix~\ref{apx:application}. Results in Table~\ref{tab:OtherPipeline} demonstrate that our model could improve performance over existing frameworks.

\input{Tables/Transferability}
\section{Conclusion}
In this work, we demonstrate, for the first time, consistent and effective performance improvements using DPO on the Text-to-SQL task, enabled by synthetic CoT solutions. Through comprehensive experiments and detailed analyses, we show that CoT reasoning is essential for unlocking DPO's potential, as it mitigates reward hacking, enhances the model's discriminative capabilities, and improves scalability during DPO training. We believe these findings will inspire researchers and practitioners to develop more robust and effective open-source Text-to-SQL models.



\section{Limitations}
The databases provided by the Bird dataset are typically large, leading to significant time consumption when executing SQL queries. This slows down the collection of preference data, as all sampled SQL queries must be executed on the databases to obtain feedback. To address this, we modify Bird's evaluation script to enable parallel execution of SQL queries using multi-processing, significantly accelerating the preference data construction process. However, resource contention among multiple processes can lead to SQL execution timeouts, causing correct predicted SQL queries to be incorrectly classified as incorrect, thereby introducing false negatives. This inaccuracy in feedback signals could potentially impact DPO training.

In contrast, the Spider dataset presents a different challenge due to the simplicity of its database values. Relying on execution results to distinguish between correct and incorrect SQL queries can lead to false positives~\cite{DBLP:conf/emnlp/ZhongYK20testsuite}, resulting in potentially unreliable feedback signals for DPO training.

\bibliography{alignment, custom, text-to-sql}

\appendix

\section{Preliminary Experiments on DPO Tricks}\label{apx:dpotricks}
Before exploring chain-of-thought reasoning, we extensively experiment with applying various DPO tricks to the original Bird dataset. This included hyperparameter tuning, using different loss variants, and exploring alternative preference data construction strategies. Specifically, for hyperparameter tuning, we evaluate different values of the $\beta$ parameter in DPO, testing 0.05 and 0.2 (the default value is 0.1). For loss variants, we experiment with augmenting the DPO loss by incorporating the SFT loss for correct responses (i.e., the correct sampled SQL queries). Additionally, we explore replacing DPO with its variant, KTO~\citep{DBLP:journals/corr/abs-2402-01306kto}. For preference data construction strategies, we follow SENSE~\cite{yang2024sense} to fine-tune a small-scale language model, Deepseek-coder-1.3b-instruct, on Bird's training set to collect preference data. The results, summarized in Table~\ref{tab:dpotrick}, show that none of these approaches yield significant performance improvements for DPO.


\definecolor{darkgreen}{RGB}{0,150,0}
\begin{table}[ht]
    \centering
\begin{adjustbox}{max width=\columnwidth}
    \begin{tabular}{ >{\centering\arraybackslash}p{0.4\columnwidth} | >{\centering\arraybackslash}p{0.4\columnwidth} }
        \toprule
         \multicolumn{2}{c}{\textbf{DPO Tricks (SFT$=58.8\%$)}} \\
         \midrule
         \multicolumn{2}{c}{\texttt{Hyper-parameter Tuning}} \\ \midrule
        $\beta=0.05$ & $\beta=0.2$ \\
        57.6\% (\textbf{\textcolor{darkgreen}{-1.2\%}}) & 57.8\% (\textbf{\textcolor{darkgreen}{-1.0\%}}) \\
        \midrule
         \multicolumn{2}{c}{\texttt{Loss Variants}} \\ \midrule 
         +SFT Loss & KTO \\
         60.0\% (\textbf{\textcolor{red}{+1.2\%}}) & 59.8\% (\textbf{\textcolor{red}{+1.0\%}}) \\
         \midrule
         \multicolumn{2}{c}{\texttt{Data Construction Strategies}} \\ \midrule
        \multicolumn{2}{c}{SENSE \citep{yang2024sense}} \\ 
        \multicolumn{2}{c}{58.4\% (\textbf{\textcolor{darkgreen}{-0.4\%}})} \\
        \bottomrule
    \end{tabular}
\end{adjustbox}
    \caption{Even with tricks applied, DPO still struggles to improve model performance on the Bird dataset (greedy decoding). The base model is Qwen2.5-7B-Instruct.}
    \label{tab:dpotrick}
\end{table}

\section{Chain-of-Thought Solutions} \label{apx:cotSynthesis}
In this section, we present the prompts used for Chain-of-Thought (CoT) reasoning synthesis and provide qualitative examples of the model's step-by-step Text-to-SQL responses.

\subsection{Prompt for Synthesis}
We carefully design the prompts used for CoT synthesis, as shown in Table~\ref{tab:prompt}. In our template, the gold SQL from the dataset is provided as a reference answer. This design enables the model to generate diverse reasoning paths during sampling while striving to maintain the correctness of the synthesized outputs.

\subsection{Synthesized Chain-of-Thought}
A synthesized Chain-of-Thought solution, generated by the \verb|gpt-4o-mini-2024-07-18| on an instance from the Bird Train dataset, is illustrated in Table~\ref{tab:synthesisCoT}. The response begins with an analysis of the input question, followed by a step-by-step breakdown of the SQL generation process. After generating the SQL, the model further provides explanations for each component of the SQL query.

\subsection{Samples From Syn CoT Models}
We select a sample question from the Bird development set and compare the responses generated during the SFT stage and the DPO stage. The SFT-generated response is shown in Table~\ref{tab:cotResponseSFT}, while the DPO-generated response is presented in the Table~\ref{tab:cotResponseDPO}. Notably, DPO corrected an entity mismatch error present in the SFT response.

\section{Direct Preference Optimization}\label{apx:dpo}
We provide a brief overview of direct preference optimization (DPO) and demonstrate how to utilize the trained DPO model to calculate the implicit reward of a response, as well as the credit assignment for each token within that response.
\subsection{Learning Objective}
For Reinforcement Learning with Human Feedback (RLHF) incorporating KL penalty, the learning objective is defined as \citep{ouyang2022rlhf}:
\begin{align*}
\max_{\pi_{\theta}}\mathbb{E}_{x\sim\mathcal{D},y\sim\pi_{\theta}(y|x)} & \left[r_{\phi}(x,y)\right] \\
 -\beta\mathbb{D}_{\mathrm{KL}} & \left[\pi_{\theta}(y\mid x)\mid\mid\pi_{\mathrm{ref}}(y\mid x)\right]
\end{align*}

Here, $\pi_{\mathrm{ref}}$ and $\pi_{\theta}$ represent the initial model distribution and the optimized policy, respectively, while $r_{\phi}$ denotes a parameterized reward model.

Direct Preference Optimization (DPO) reformulates the objective by replacing the reward function with a differentiable form, reflecting the relationship between the optimal policy and the reward function. This leads to a new objective:
\begin{align*}
\mathbb{E}_{(x,y_{w},y_{l})\sim\mathcal{D}}\left[\log \sigma\left(\beta \log\frac{\pi_{\theta}(y_{w}\mid x)}{\pi_{\mathrm{ref}}(y_{w}\mid x)}\right.\right. \\
\left.\left.  -\beta\log\frac{\pi_{\theta}(y_{l}\mid x)}{\pi_{\mathrm{ref}}(y_{l}\mid x)} \right)\right]
\end{align*}

In this formulation, $y_w$ denotes the chosen output and $y_l$ denotes the rejected output. The parameter $\beta$ controls the penalty strength imposed by the KL divergence. By collecting pairwise preference data, the model can be optimized using supervised fine-tuning, achieving a performance comparable to RLHF \citep{rafailov2024dpo}.

\subsection{Implicit Reward}\label{apx:implicit_reward}

DPO implicitly encodes a reward model within the generative model. The reward of a given input-output pair $(x, y)$ can be derived as:

$$r_{\theta}(x, y) = \beta \log \frac{\pi_\theta(y\mid x)}{\pi_{\mathrm{ref}}(y\mid x)}$$

As the DPO training progresses, the optimized model simultaneously becomes a better generative model and a more refined reward model. After training, the implicit reward model, which is derived via conditional likelihood, can be independently used as a reward function \citep{lambert2024rewardbench, chen2024bootstrappinglanguagemodelsdpo, DBLP:journals/corr/abs-2402-06457vstar}.

\subsection{Token-level Credit Assignment}\label{apx:token_reward}

The implicit reward scores the entire output as a whole. By decomposing the conditional probability that featuring autoregressive generation process, the reward can be re-expressed as:
$$r_{\theta}(x,y) = \sum_{t=1}^{T}\beta \log \frac{\pi_\theta(y_t\mid x,y_{1:t-1})}{\pi_{\mathrm{ref}}(y_t\mid x,y_{1:t-1})}$$

This decomposition allows for the calculation of token-level rewards, as the model score for each token can be identified separately. Although DPO training uses supervision at the full-sequence level, evidence has shown that the model can generalize compositionality to some extent, allowing it to distribute the reward signal to key tokens \citep{rafailov2024dpo2}. This facilitates credit assignment across the output sequence. The resulting dense reward can be utilized for further training and optimization \citep{zhong2024dpo-ppo}.

\section{Implementation Details}\label{apx:imp_details}

The computational environment used in our experiments is equipped with a 64-core Intel(R) Xeon(R) Platinum 8358 CPU @ 2.60GHz and 8 NVIDIA A800 GPUs with 80GB of memory each, running CUDA version 12.1.1.

Training is conducted using Llama Factory \citep{zheng2024llamafactory}, with FlashAttention 2.0 \citep{dao2023flashattention} and DeepSpeed ZeRO-3 \citep{rasley2020deepspeed} enabled. For models larger than 10B parameters, the DPO stage utilizes CPU offloading for both model parameters and optimizer states. During inferenece, the trained models are hosted with vLLM \citep{kwon2023vllm}.

For training, we employ full-parameter fine-tuning with bf16 mixed-precision \citep{micikevicius2018bf16}. The optimizer used was AdamW \citep{DBLP:conf/iclr/LoshchilovH19adamw} with default parameters (\(\beta_1 = 0.9\), \(\beta_2 = 0.99\)). A cosine decay learning rate schedule and a linear warmup over the first 5\% of training steps are also applied. The context window of models is set to 4096 tokens.

Across all training phases, we adopt a consistent batch size of 64. The learning rates for the SFT and DPO phases are set to \(1 \times 10^{-5}\) and \(1 \times 10^{-6}\), respectively, for models smaller than 10B, and \(7 \times 10^{-6}\) and \(7 \times 10^{-7}\), respectively, for models larger than 10B. The \(\beta\) parameter for DPO is set to 0.1. 

All models are trained for 4 epochs during the SFT phase, and the best checkpoint is selected to serve as the reference model in DPO, based on the maj@K metric on the development set. Training is then continued for 8 epochs during the DPO phase.

Unless otherwise specified, during inference time across all stages (including chain-of-thought synthesis), the sampling budget is set to 16, with the $temperature$ and $topK$ parameters set to 1.0 and 32, respectively.

\section{Training Data Details}\label{apx:datadetails}
\noindent\textbf{Quantity.}
In Bird dataset, the Vanilla SFT data consists of $9,428$ instances, while the CoT SFT data consists of $9,428 \times K$ instances (for Table~\ref{tab:model-comparison}, $K=16$, as we generate 16 CoT solutions for each training sample in the original dataset). 

The size of the DPO training data is smaller than that of the SFT training data and is model-dependent. This is because, during the construction of the preference data, we exclude data samples for which all SFT model-generated CoT solutions are either entirely correct or entirely incorrect. As a result, both Vanilla and Syn CoT preference datasets contain approximately 1.5k-2.5k preference pairs (e.g., Qwen-7b-Instruct Syn CoT has 1,546 pairs). The relationship between sample budgets and the quantity of DPO training data is illustrated in Figure~\ref{fig:scalePrefFull} and Figure~\ref{fig:scalePrefLog}.

\noindent\textbf{Construction of Input-Output Sequences and Their Average Length.}
Input prompt has an average length of $965$ tokens, which is the same for Vanilla and Syn CoT settings. The input prompt includes not only the question but also database information, such as table and column names, primary and foreign key relationships, and potentially useful database values. Following CodeS~\citep{li2024codes}, we first use a schema item classifier to identify the most relevant tables and columns based on the question. To improve recall accuracy, we replace the backbone model of the classifier from RoBERTa-Large (355M)~\citep{liu2019roberta} to XLM-RoBERTa-XL (3.5B)~\citep{goyal2021xlroberta}. We then use the ``coarse-to-fine'' database value-matching approach to retrieve question-related values from the database. The retrieved tables, columns, values, and remaining primary and foreign keys form the database prompt.

The output label of Vanilla (gold SQL in the Bird dataset) has an average length of $44$ tokens, while the synthesized chain-of-thought solutions have an average length of $404$ tokens. Reported token numbers is measured by the tokenizer of Qwen-7b-Instruct.

\section{Data Quality Ablation}\label{apx:qualityablation}

From Table~\ref{tab:model-comparison}, we can see that there are many cases where Syn CoT SFT has already surpassed Vanilla SFT, thus, it is natural to doubt that the benefit could be brought about by the potent proprietary model. In this section, we first analyze the ability of GPT-4o-mini on Text-to-SQL, then, we replace GPT-4o-mini with smaller open-sourced models to synthesize CoT reasoning paths, therefore further confirm that it is the chain-of-thought style solution path itself that enhance the effect of DPO.

\noindent \textbf{Ability of the Synthesizers.} We evaluate GPT-4o-mini's capability on Bird development set, as well as other open-sourced models that we will use as synthesizers. The result is shown in Table~\ref{tab:cotquality}.

\definecolor{darkgreen}{RGB}{0,150,0}
\begin{table}[ht]
    \centering
\begin{adjustbox}{max width=\columnwidth}
    \begin{tabular}{c | c | c | c }
        \toprule
         \multirow{2}{*}{\textbf{Model}} &  \multicolumn{3}{c}{\textbf{Bird Dev}} \\ \cline{2-4}
         & ZS-CoT & FS-SQL & FS-CoT \\ \midrule
        GPT-4o-mini & 50.1 & 55.3 & 53.0 \\
        Qwen(1.5B) & 15.6 & 23.1 & 18.9 \\
        Qwen(7B) & 41.8 & 47.6 & 43.1 \\
        Qwen(32B) & 52.9 & 55.3 & 53.0 \\
        \bottomrule
    \end{tabular}
\end{adjustbox}
    \caption{Performance of models used for chain-of-thought solution synthesis under different prompting strategies (Greddy). \textbf{Qwen (XB)}: Qwen2.5-XB-Instruct; \textbf{ZS-CoT}: Zero-shot CoT; \textbf{FS-SQL}: 3-Shot examples randomly chosen from Bird Train; \textbf{FS-CoT}: 3-Shot examples randomly chosen from its own synthesized CoT solutions on Bird Train. }
    \label{tab:cotquality}
\end{table}
As a comparison, Qwen2.5-7B-Instruct, after fine-tuning on original Bird Train data, reaches 58.8, which is higher than the EX of any synthesizer, as shown in Table~\ref{tab:model-comparison}.

It is evident that directly prompting closed-source instruction-tuned models does not inherently offer an advantage in the Text-to-SQL task, consistent with findings from previous work \citep{li2024codes}. Closed-source models only perform well when incorporating with complex multi-agent designs \citep{pourreza2024chase-sql}. Therefore, in this paper, we use GPT-4o-mini merely as a substitute for manual effort, allowing us to quickly and cost-effectively obtain CoT solutions of Text-to-SQL data, rather than distilling capabilities from a powerful model.

\noindent \textbf{DPO with Different CoT Quailities.} As shown in Table~\ref{tab:cotquality}, CoT quality varies for different size of synthesizers. We then use each of them to synthesize CoT solutions, and train the corresponding Syn CoT model separately (Base model is Qwen2.5-7B-Instruct). The results are presented in Table~\ref{tab:opencotpdo}.

\definecolor{darkgreen}{RGB}{0,150,0}
\begin{table}[ht]
    \centering
\begin{adjustbox}{max width=\columnwidth}
    \begin{tabular}{c | c | c }
        \toprule
         \multirow{2}{*}{\textbf{Synthesizer}} &  \multicolumn{2}{c}{\textbf{Bird Dev}} \\ \cline{2-3}
         & SFT & DPO \\ \midrule
        GPT-4o-mini & 57.4 & 61.9 {\color{red}(+4.5)} \\
        Qwen(1.5B) & 40.9 & 59.1 {\color{red}(+18.2)} \\
        Qwen(7B) & 54.3 & 60.6 {\color{red}(+6.3)} \\
        Qwen(32B) & 59.3 & 62.5 {\color{red}(+3.2)} \\
        \bottomrule
    \end{tabular}
\end{adjustbox}
    \caption{Performance of models trained from CoT-enhanced data generated by different synthesizers (Greddy). \textbf{Qwen (XB)}: Qwen2.5-XB-Instruct. }
    \label{tab:opencotpdo}
\end{table}

Its evident that all these models achieve significant improvement in the DPO phase. Interestingly, the impact of CoT quality on model performance is significantly weakened after DPO.

\section{Results on Spider and Its Robustness Variants} \label{apx:spider}

In addition to the Bird benchmark, we also train and evaluate three representative base models on the Spider benchmark. We further assess the models trained on Spider with robustness test sets. 

However, during our experiments, we identified several issues with using CoT and DPO techniques on the Spider dataset:

\begin{enumerate}
    \item \textbf{Simplicity of SQL in Spider: Most SQL queries in Spider are quite simple.} A typical example is: \texttt{Q: How many concerts are there in year 2014 or 2015? A: SELECT count(*) FROM concert WHERE year = 2014 OR year = 2015}. smaller pre-trained language model methods have already achieved good results on these queries \citep{li2023resdsql}. Additionally, the best models on the Spider benchmark have reached human-level performance (91.2\%\footnote{from the Spider Leaderboard: \url{ttps://yale-lily.github.io/spider}}). It is widely acknowledged that CoT often does not perform better on simple questions, possibly due to overthinking issues \citep{sprague2024cotcotchainofthoughthelps}.
    \item \textbf{Inaccurate Feedback on Execution Results in Spider:} When constructing preference data, we rely on execution results on the database to judge the correctness of sampled SQL. The Bird dataset's databases are specially designed with massive rows, whereas Spider's databases have fewer rows (\#Row/DB: Bird ($549K$) vs. Spider ($2K$), $<0.5\%$ \citep{li2024bird}), leading to potential false positives during evaluation \citep{DBLP:conf/emnlp/ZhongYK20testsuite}. Although subsequent work has attempted to mitigate this by constructing multiple test suite (TS) databases, the Spider benchmark only provides TS databases for the development set, not for the training set, leading to inaccurate preference pair construction.
    \item \textbf{Small Scale of Constructible Preference Data on Spider:} Due to the first issue, most models on the Spider dataset achieve very high accuracy on the training set (Pass@16 $\geq 99\%$, compared to Bird's $\leq80\%$), resulting in a very small preference dataset ($0.1-0.3k$, with the SFT phase dataset being $7k$. As for comparison, DPO data on Bird is approximately $1.5-2.5k$).
\end{enumerate}

Based on these issues, we choose to focus primarily on the model's performance on the Bird dataset, one of the most challenging Text-to-SQL benchmarks that closely reflects real-world scenarios. Therefore, results related to Spider are included as a reference in the appendix due to constraints of limited space, as shown below.

\subsection{Spider}


The model performance on Spider's development set is summarized in Table~\ref{tab:spiderDev}. Despite the aforementioned challenges, the Syn-CoT model consistently achieved improvements during the DPO stage. 

\subsection{Spider Variants}

We select the best checkpoint according to the Spider development set and directly evaluate it on these robustness test sets. The results for Spider-Syn, Spider-Realistic, and Spider-DK are presented in Table~\ref{tab:spiderVariants}. The CoT model continues to demonstrate consistent performance improvements during the DPO stage, further confirming its strong generalization capabilities after DPO training.

Dr.~Spider is a more comprehensive and sophisticated robustness test set, which categorizes all perturbations into three major types: database (DB), natural language question (NLQ), and SQL query. It then further subdivides them into 17 subcategories. For each type of perturbation, dedicated test sets are constructed~\citep{DBLP:conf/iclr/Changdrspider}.

For DB perturbations, the results of Syn CoT models are shown in Table~\ref{tab:drSpiderDBCoT}, alongside the results of the vanilla model in Table~\ref{tab:drSpiderDBVanilla}. For NLQ perturbations, the results of Syn CoT models are shown in Table~\ref{tab:drSpiderNLQCoT}, alongside the results of the vanilla model in Table~\ref{tab:drSpiderNLQVanilla}. For SQL perturbations, the results of Syn CoT models are shown in Table~\ref{tab:drSpiderSQLCoT}, alongside the results of the vanilla model in Table~\ref{tab:drSpiderSQLVanilla}.

Except for a few specific cases, Syn CoT model still demonstrates consistent improvements during the DPO stage.

\section{Error Classifications} \label{apx:class}
\subsection{Description}
In this paper, we classify errors made by predicted SQL into 6 major categories with 17 variant types. Descriptions of each category or type are shown in Table~\ref{table:ErrorCategoryDescription}.

\subsection{Error Samples}
We provide samples for each error type in our classification criteria, for external knowledge, see Table~\ref{table:ErrorSampleEK}, for schema linking, see Table~\ref{table:ErrorSampleSchema}, for value retrieval, see Table~\ref{table:ErrorSampleValue}, for operation, see Table~\ref{table:ErrorSampleOperation}, for information, see Table~\ref{table:ErrorSampleInfo}, for syntax error, see Table~\ref{table:ErrorSampleSyntax}. These are selected from model predictions on the Bird development set.

\subsection{Classification Result}
\begin{figure*}[t]
  \centering
  \begin{subfigure}[t]{0.49\linewidth}
    \centering
    \includegraphics[width=\linewidth]{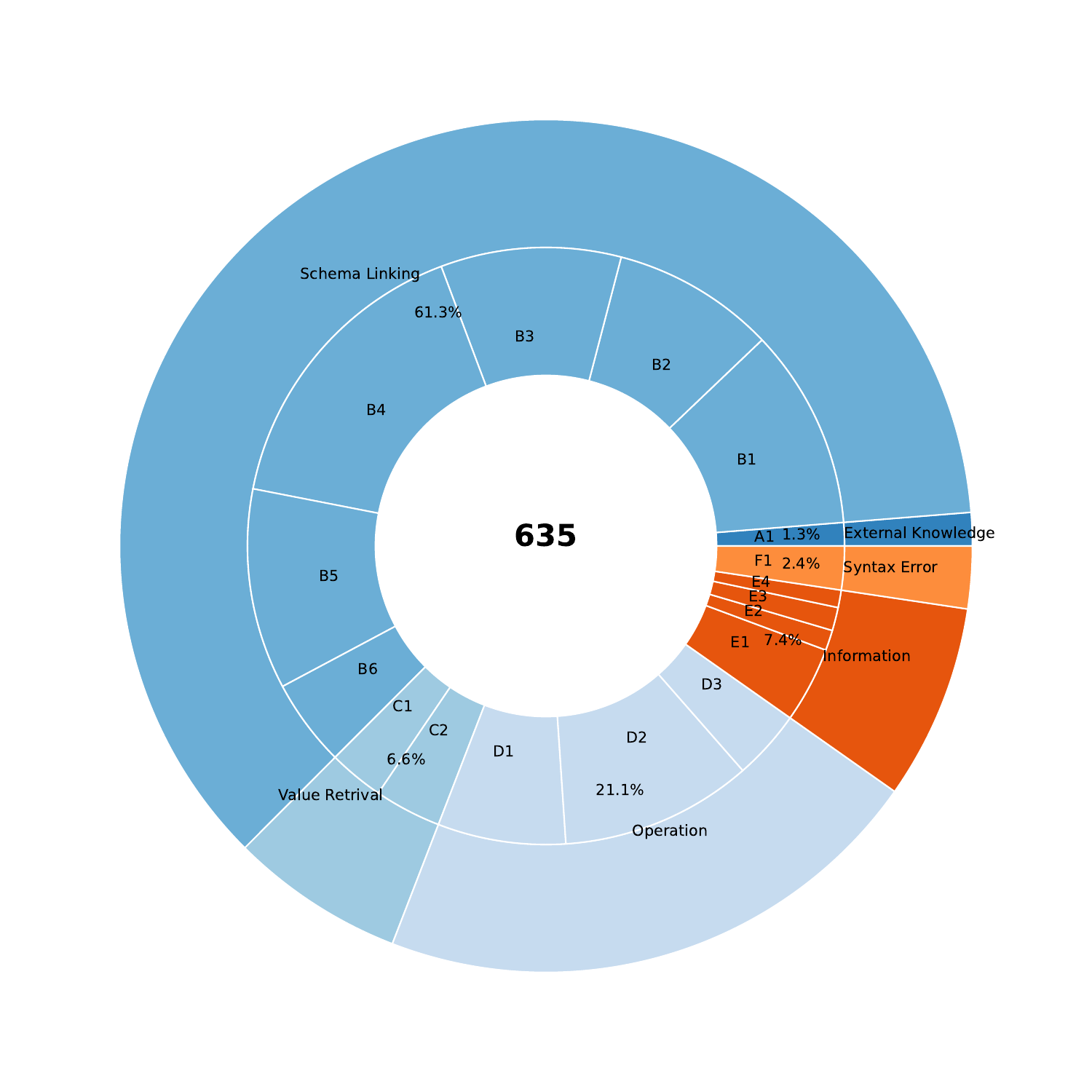}
    \subcaption{SFT}
  \end{subfigure}
  \hfill
  \begin{subfigure}[t]{0.49\linewidth}
    \centering
    \includegraphics[width=\linewidth]{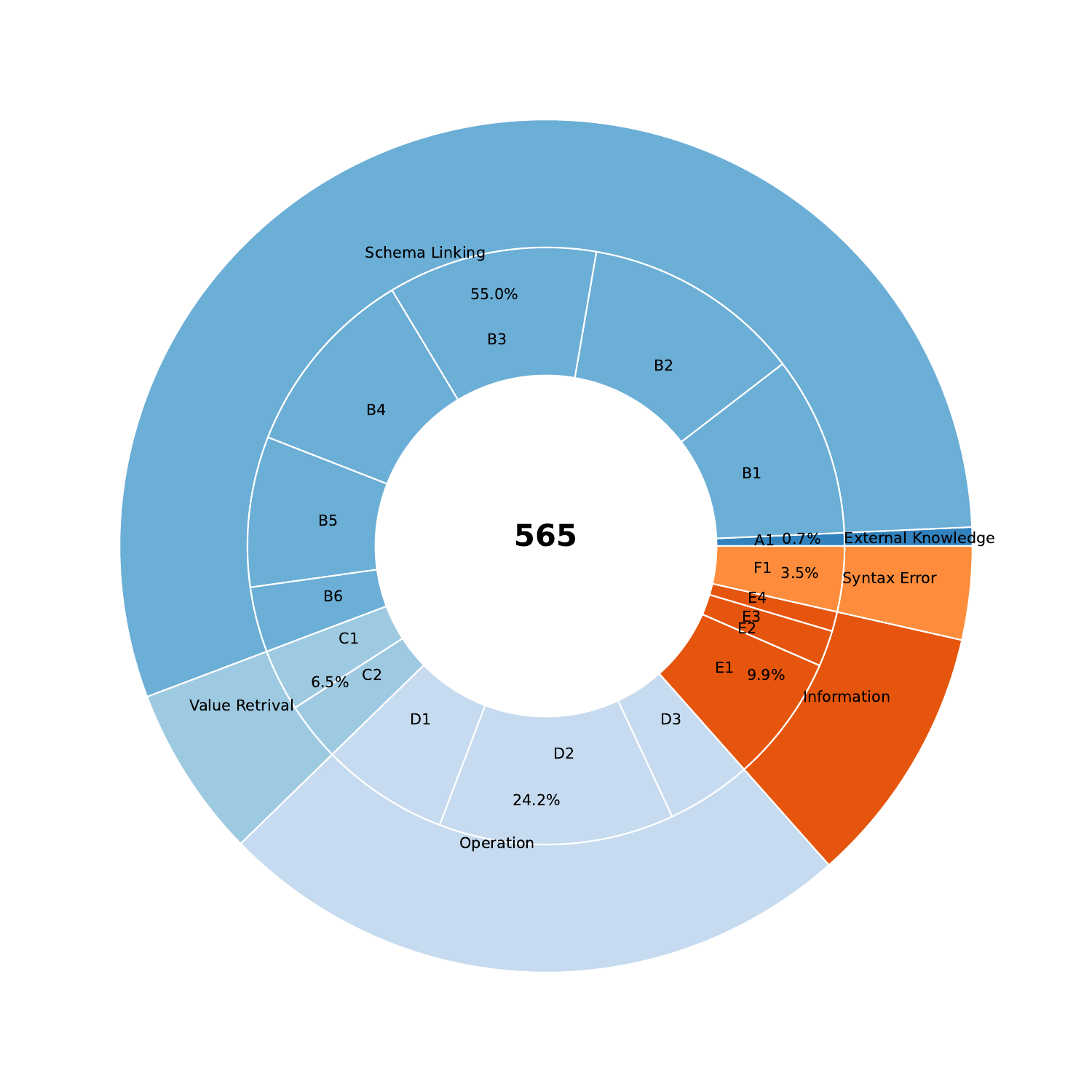}
    \subcaption{DPO}
  \end{subfigure}
  \caption{Error statistics of Syn CoT model. The percentage of each category and total error count are on the chart.}
  \label{tab:ErrorStatCoT}
\end{figure*}

\begin{figure*}[t]
  \centering
  \begin{subfigure}[t]{0.49\linewidth}
    \centering
    \includegraphics[width=\linewidth]{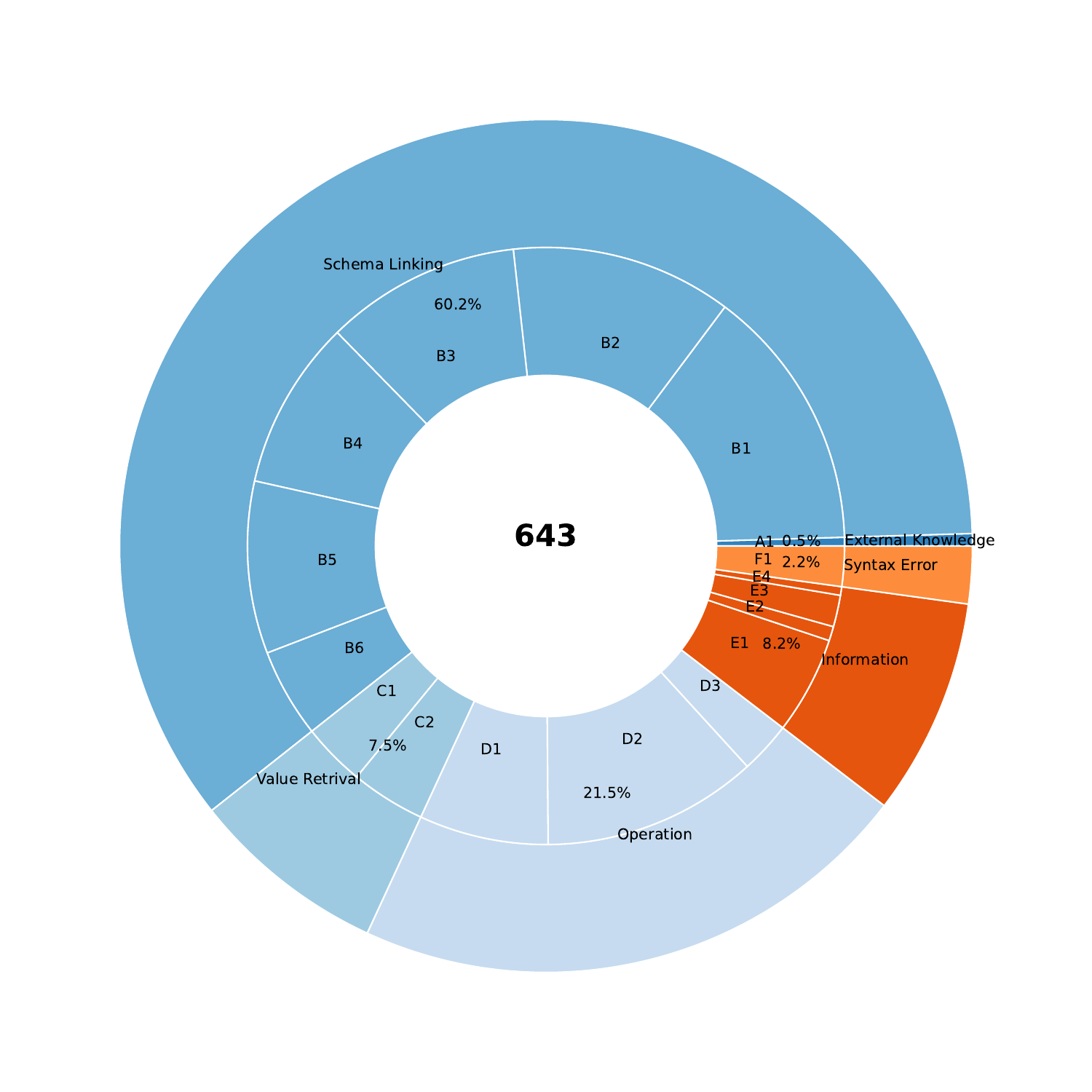}
    \subcaption{SFT}
  \end{subfigure}
  \hfill
  \begin{subfigure}[t]{0.49\linewidth}
    \centering
    \includegraphics[width=\linewidth]{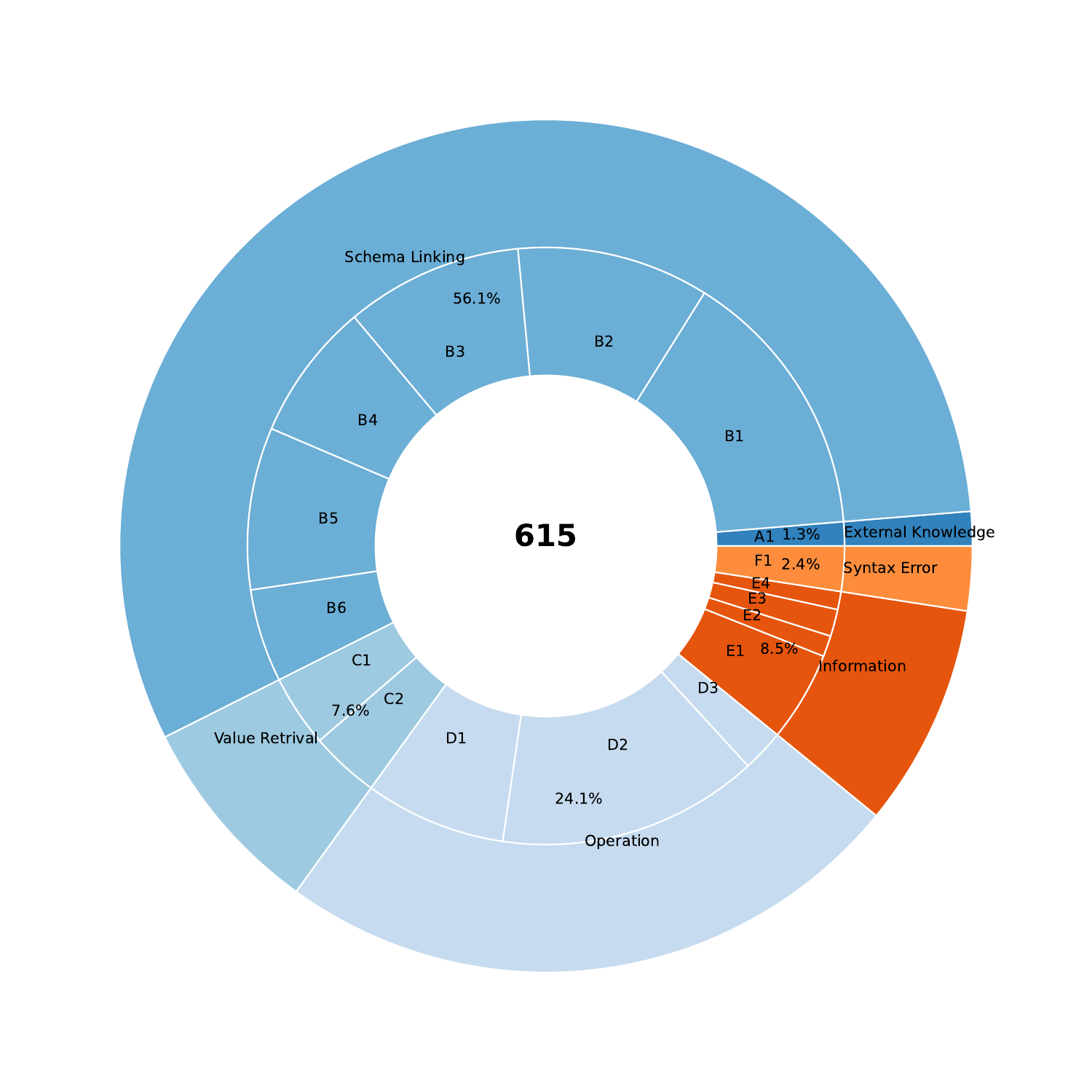}
    \subcaption{DPO}
  \end{subfigure}
  \caption{Error statistics of Vanilla model. The percentage of each category and total error count are on the chart.}
  \label{tab:ErrorStatVanilla}
\end{figure*}

The total number of errors and the proportion of each error type are presented in pie charts. Results of Syn CoT models is shown in Figure~\ref{tab:ErrorStatCoT}, and results of vanilla models is shown in Figure~\ref{tab:ErrorStatVanilla}.

A SQL query may commit multiple types of errors simultaneously. In our analysis, however, we only attribute each erroneous SQL to the most prominent type of mistake it made logically, for the convenience of analysis.

\section{More Analysis of DPO} \label{apx:analysis}

\subsection{Overall Effect}
\begin{figure*}[t]
  \centering
  \begin{subfigure}[t]{0.49\linewidth}
    \centering
    \includegraphics[width=\linewidth]{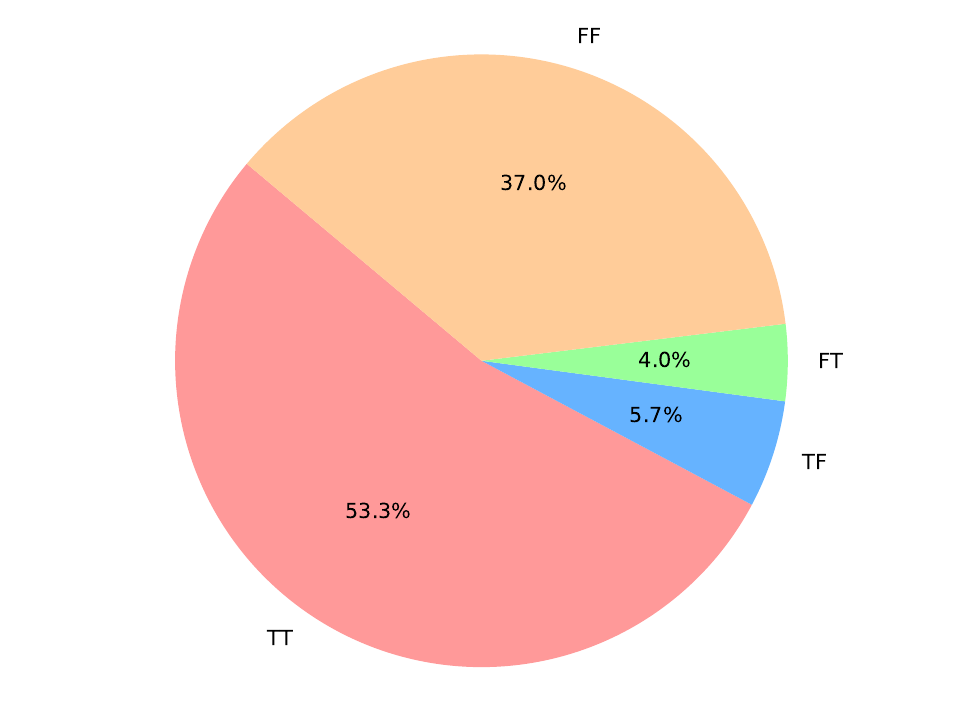}
    \subcaption{Vanilla}
  \end{subfigure}
  \hfill
  \begin{subfigure}[t]{0.49\linewidth}
    \centering
    \includegraphics[width=\linewidth]{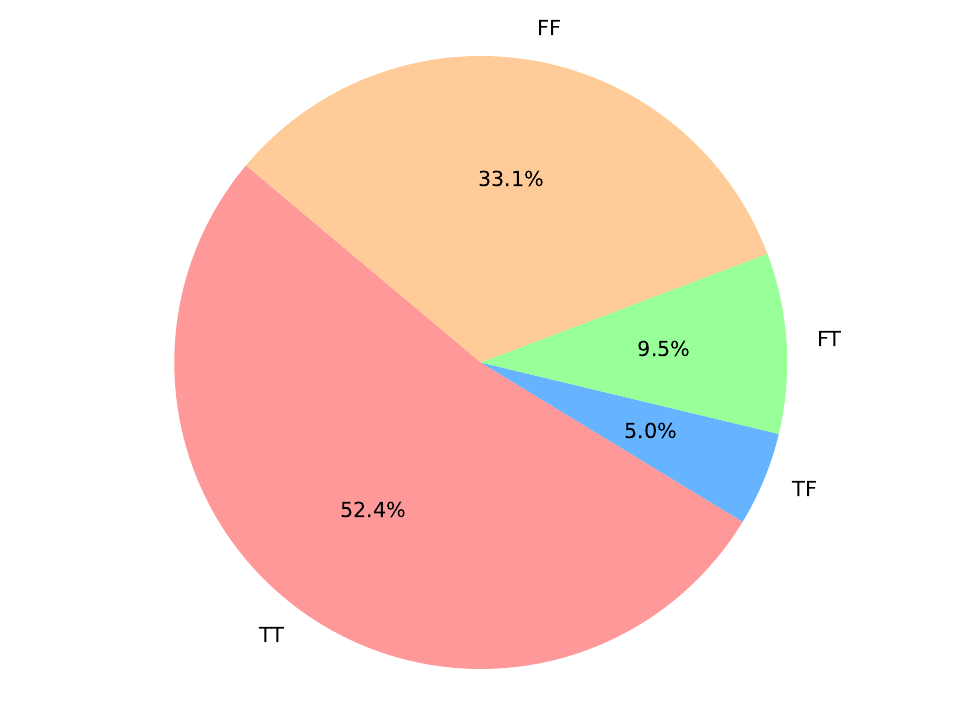}
    \subcaption{Syn CoT}
  \end{subfigure}
  \caption{Overall effect of DPO. T/F indicates an item either pass or fail, the first one is the status of SFT, while the second one refers to the status of DPO. For example, 'FT' represents instances that are corrected after DPO training. \textbf{Syn CoT outperforms Vanilla in DPO mainly by fixing more errors.}}
  \label{fig:overallDPO}
\end{figure*}

The changes in the correctness of model-generated outputs before and after the DPO stage are illustrated in Figure~\ref{fig:overallDPO}. The CoT model corrects a greater number of errors during the DPO stage, while the proportion of newly introduced errors remained comparable to that of the vanilla model.

\subsection{Effect on Difficulty Classes}
\begin{figure*}[t]
  \centering
  \begin{subfigure}[t]{0.49\linewidth}
    \centering
    \includegraphics[width=\linewidth]{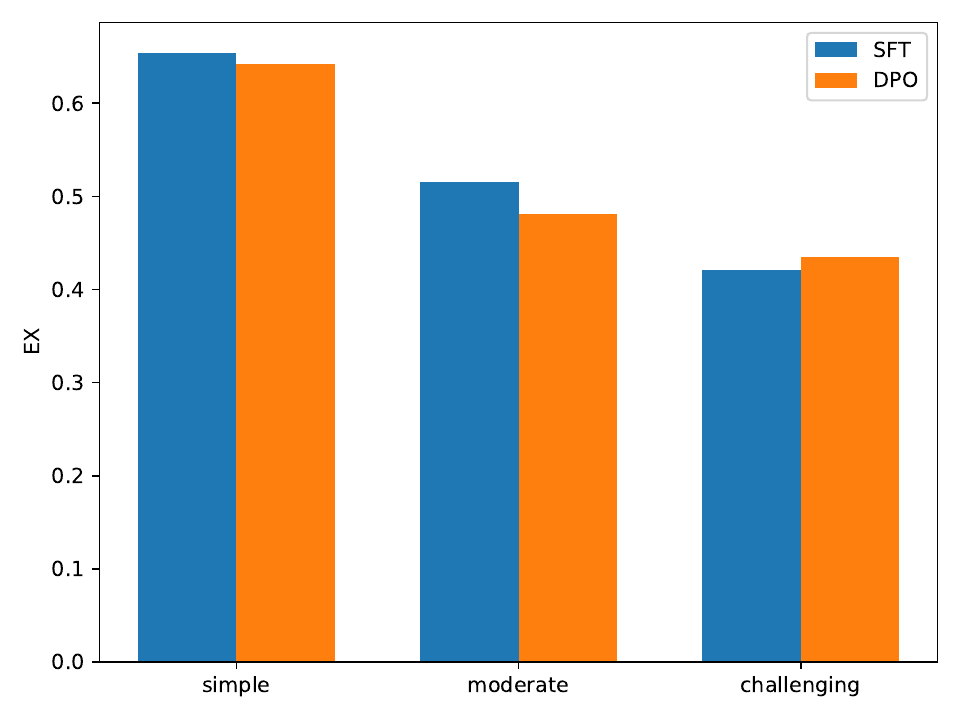}
    \subcaption{Vanilla}
  \end{subfigure}
  \hfill
  \begin{subfigure}[t]{0.49\linewidth}
    \centering
    \includegraphics[width=\linewidth]{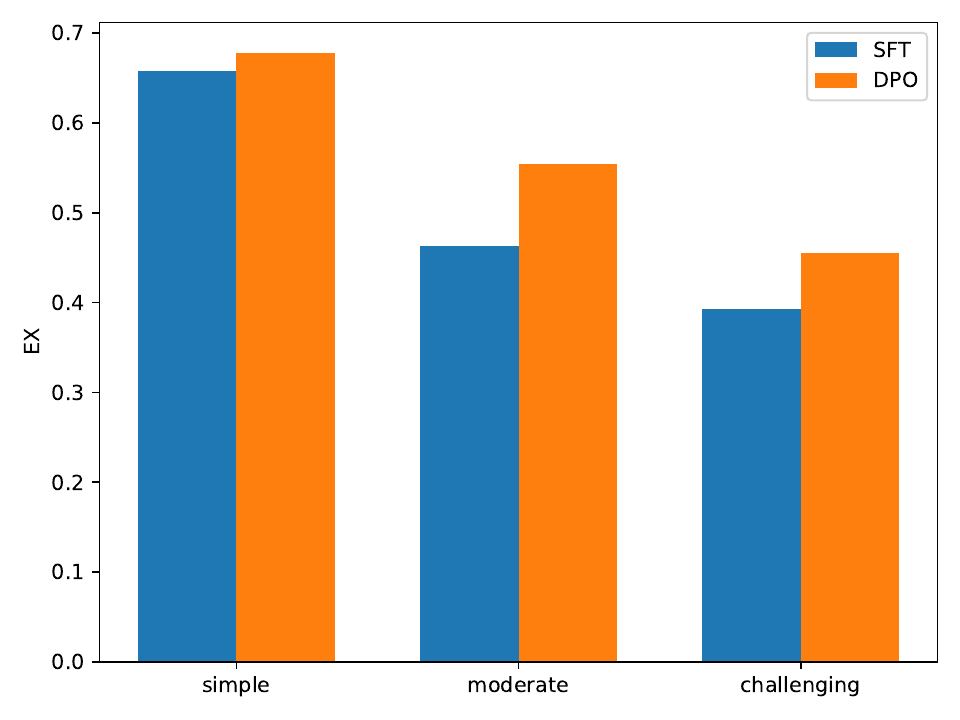}
    \subcaption{Syn CoT}
  \end{subfigure}
  \caption{Effect of DPO on different difficulty sets. Vanilla DPO struggles in every difficult set. \textbf{Syn CoT DPO mainly enhances model performance on moderate questions, then harder questions.} Surprisingly, even with Syn CoT, model performance on simple questions does not improve much.}
  \label{fig:difficultiesDPO}
\end{figure*}

We analyze the impact of DPO on questions of varying difficulty levels, as shown in Figure~\ref{fig:difficultiesDPO}. The CoT model exhibits significant improvements in performance on medium- and high-difficulty questions during the DPO stage, while the improvements on simpler questions are relatively limited.

\subsection{Fix Rate Difference}
To facilitate a detailed comparison of how the CoT model enhances DPO's error correction capabilities for different error types, we rank the error types by the increase in fix rates introduced by DPO. The results are presented in Table~\ref{tab:fixRateRank}.

\subsection{Emerging Errors}
The number of newly introduced errors during the DPO process is summarized in Table~\ref{tab:newErrorStat}. Overall, the distribution of emerging errors in the Syn CoT model is similar to that of the vanilla model. However, while CoT improves DPO's ability to address hallucination errors, it also leads to an increase in newly introduced hallucinations during the DPO stage.

\subsection{Transition Matrices}
\begin{figure*}[t]
  \centering
  \begin{subfigure}[t]{0.49\linewidth}
    \centering
    \includegraphics[width=\linewidth]{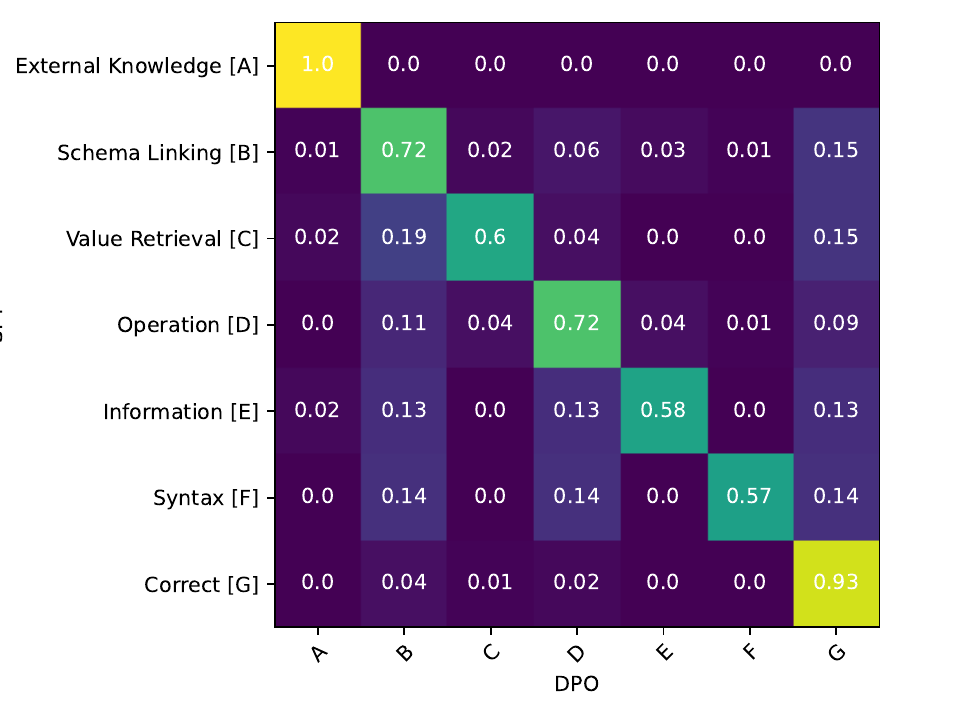}
    \subcaption{Vanilla}
  \end{subfigure}
  \hfill
  \begin{subfigure}[t]{0.49\linewidth}
    \centering
    \includegraphics[width=\linewidth]{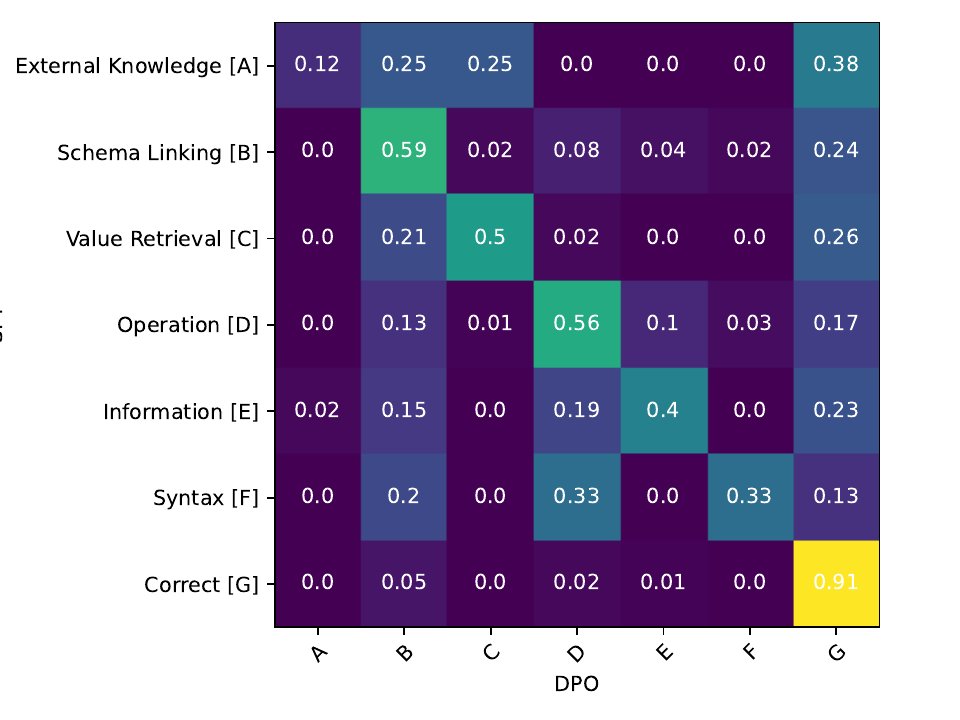}
    \subcaption{Syn CoT}
  \end{subfigure}
  \caption{Comparison of transition matrix among error categories. G indicates correct instances.}
  \label{fig:matrixTight}
\end{figure*}

\begin{figure*}[t]
  \centering
    \includegraphics[width=\linewidth]{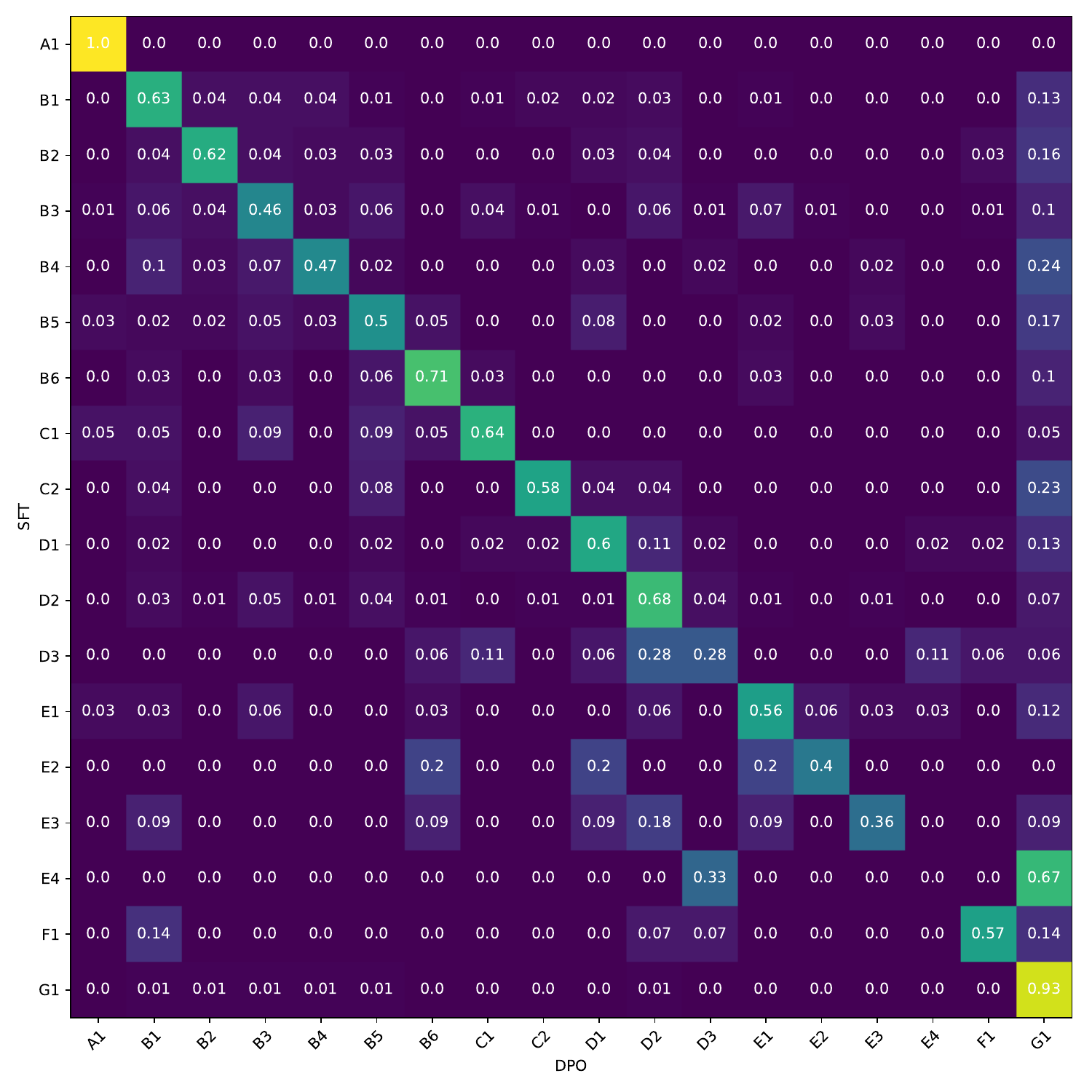}
  \caption{Vanilla full transition matrix of error types. G1 indicates correct instances.}
  \label{fig:matrixFullVanilla}
\end{figure*}

\begin{figure*}[t]
  \centering
    \includegraphics[width=\linewidth]{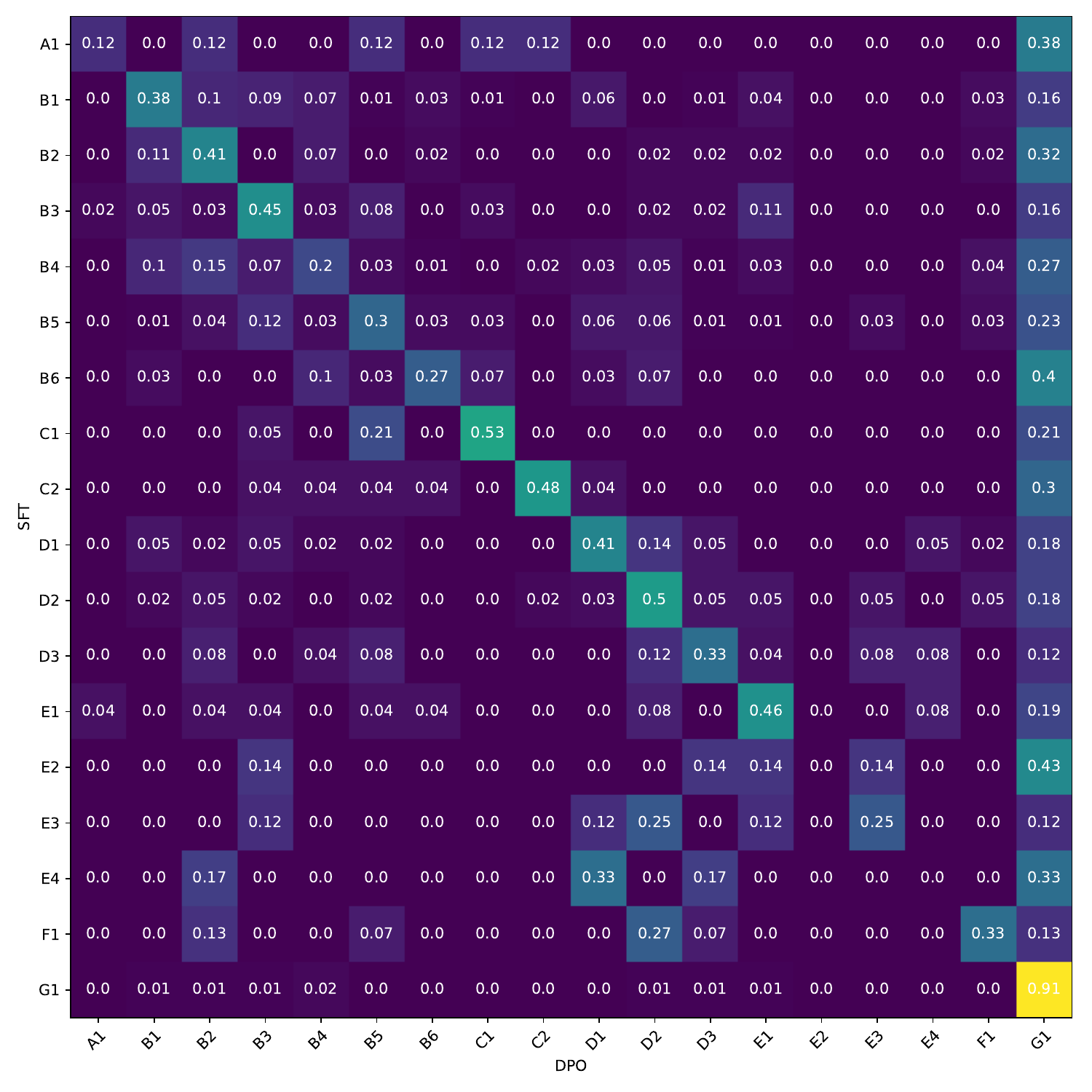}
  \caption{Syn CoT full transition matrix of error types. G1 indicates correct instances.}
  \label{fig:matrixFullCoT}
\end{figure*}

The transitions between error categories before and after DPO are depicted in Figure~\ref{fig:matrixTight}. For a more detailed view of these transitions, the vanilla model's error type transitions are shown in Figure~\ref{fig:matrixFullVanilla}, while those of the Syn CoT model are presented in Figure~\ref{fig:matrixFullCoT}.

\subsection{Weakness}
\textbf{Existing Text-to-SQL Methods Can Complement DPO's Weaknesses.} Despite improvements brought by CoT, DPO remains less effective at fixing certain error types. However, these align with core challenges that existing Text-to-SQL methods aim to address. For example, the model frequently fails to recall relevant \underline{Table} (Fix~15.9\%) and \underline{Column} (Fix~16.1\%), a key challenge of schema linking. Notable recent works include CHESS \citep{talaei2024chess} and E-SQL \citep{DBLP:journals/corr/abs-2409-16751esql}. 
\underline{Syntax Error} (Fix~13.3\%) is another tricky problem. \citet{dac-sql, magic-sql} have proposed post-generation execution and repair strategies to ensure executable returned SQL.

\section{Experiment Design} \label{apx:design}
\subsection{Evaluation Preference Dataset}

To ensure a fair comparison of the discriminative capabilities between the Vanilla and CoT models, we construct the mentioned evaluation preference dataset as follows. 

First, both SFT models are used to sample from the development set. For any data point where both models could generate paired data ($i.e.$, both could simultaneously sample a positive and a negative example), we use the sampling outcomes from the CoT model to randomly construct a preference pair. Subsequently, we extract the SQL portion of the pair to serve as the preference pair for the vanilla model, incorporating it into their respective evaluation sets. Through this construction process, we ensure that the databases, questions, and SQLs in the dataset for both models are identical.

\section{Prominent Reward Hacking Patterns}\label{sec:RHPatterns}
In this section, we provide other prominent reward hacking patterns of the DPO training process in the Vanilla setting from our observations, as illustrated in Table~\ref{tab:egRH1},~\ref{tab:egRH2},~\ref{tab:egRH3}.

\section{More Scaling Results} \label{apx:scaling}
\begin{figure*}[ht]
  \centering
  \begin{subfigure}[t]{0.31\linewidth}
    \centering
    \includegraphics[width=\linewidth]{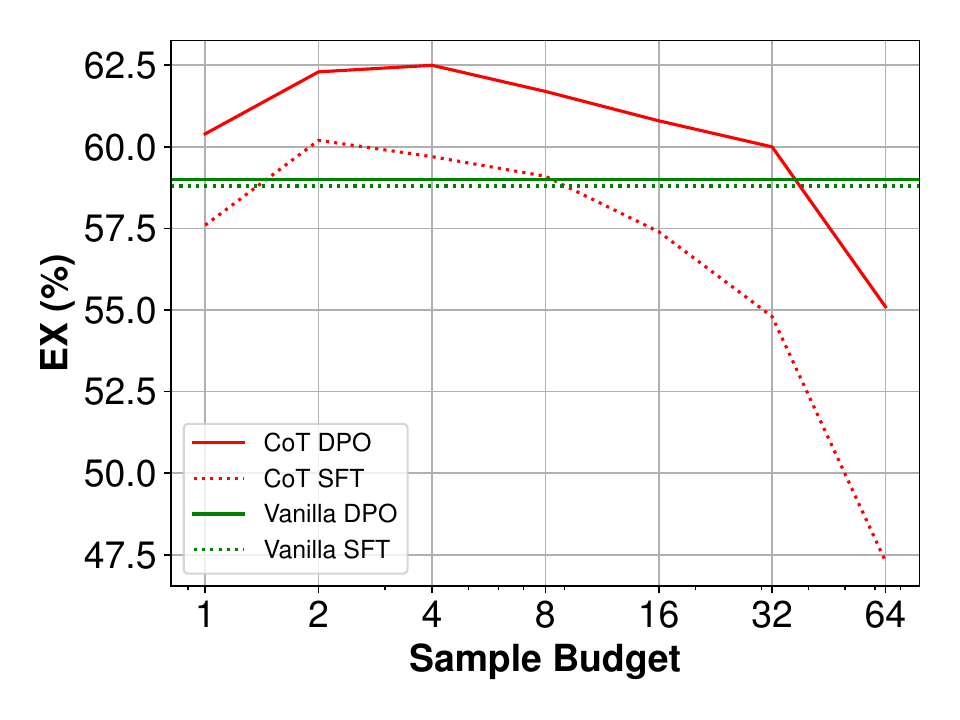}
    \subcaption{Greedy}
  \end{subfigure}
  \hfill
  \begin{subfigure}[t]{0.31\linewidth}
    \centering
    \includegraphics[width=\linewidth]{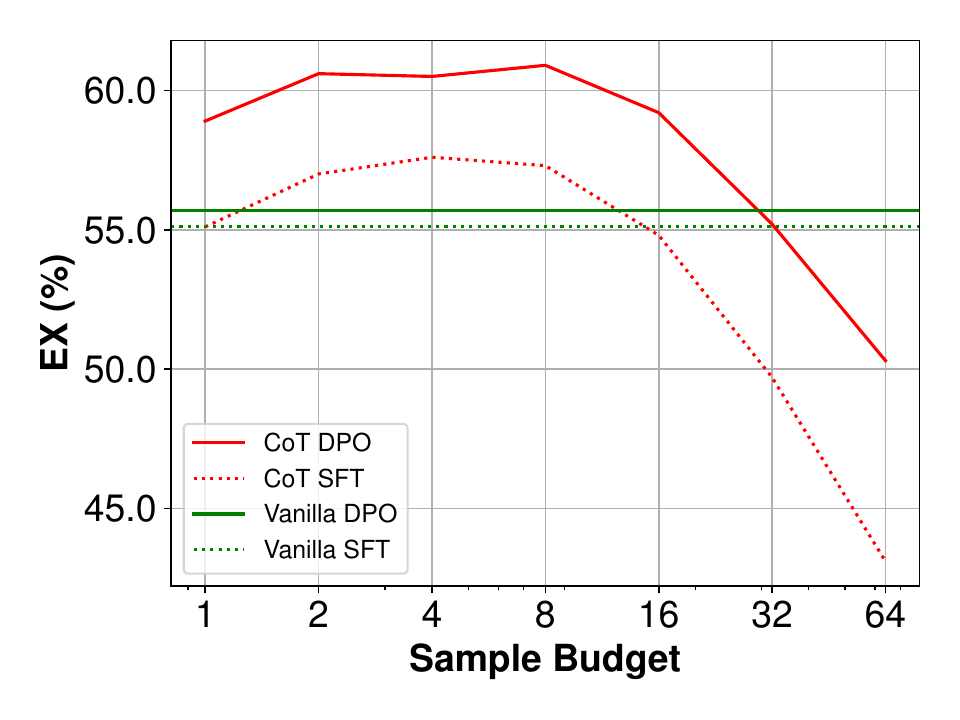}
    \subcaption{Pass@1}
  \end{subfigure}
  \hfill
  \begin{subfigure}[t]{0.31\linewidth}
    \centering
    \includegraphics[width=\linewidth]{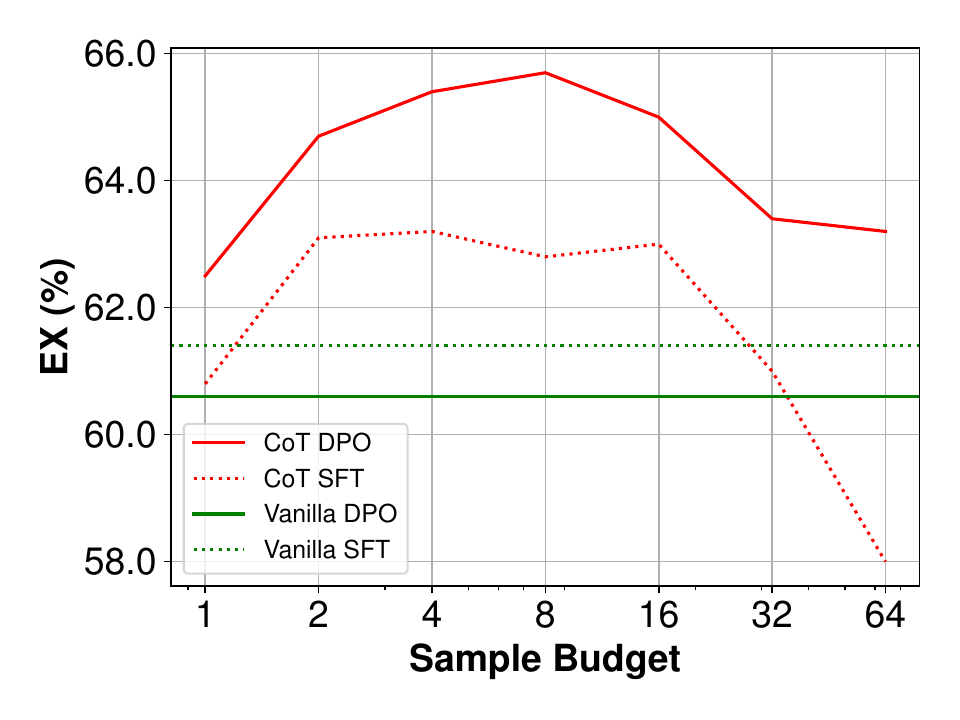}
    \subcaption{Maj@K}
  \end{subfigure}
  \caption{Model performance with different sample budget $K$ in Chain-of-Thought reasoning synthesis tested under different inference strategies. The base model is Qwen2.5-7B-Instruct. }
  \label{fig:scaleSynFull}
\end{figure*}

\begin{figure*}[ht]
  \centering
  \begin{subfigure}[t]{0.31\linewidth}
    \centering
    \includegraphics[width=\linewidth]{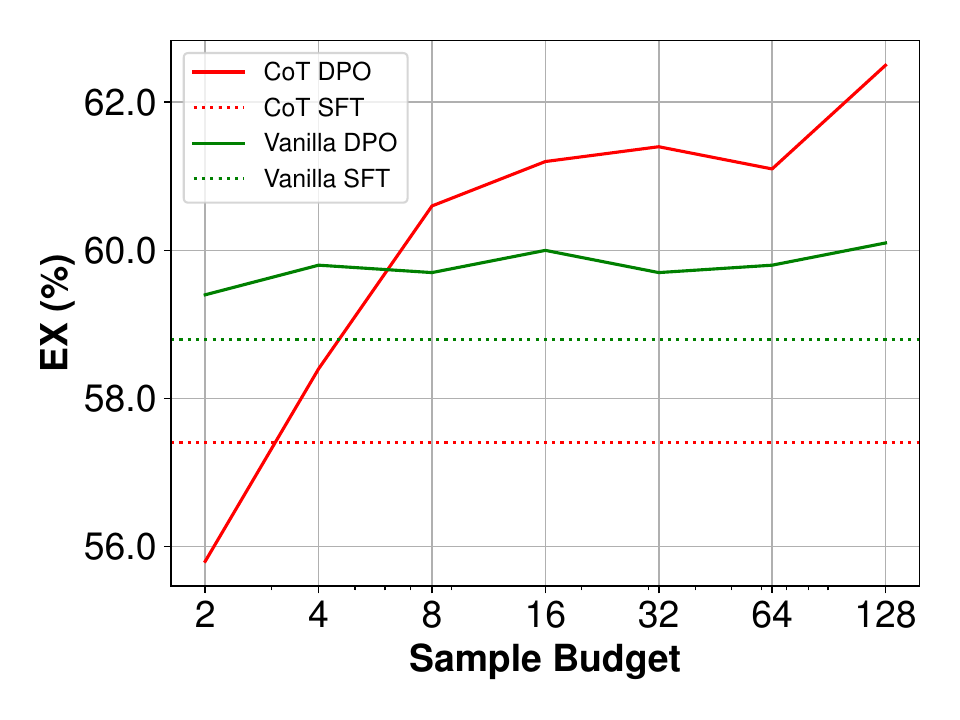}
    \subcaption{Greedy}
  \end{subfigure}
  \hfill
  \begin{subfigure}[t]{0.31\linewidth}
    \centering
    \includegraphics[width=\linewidth]{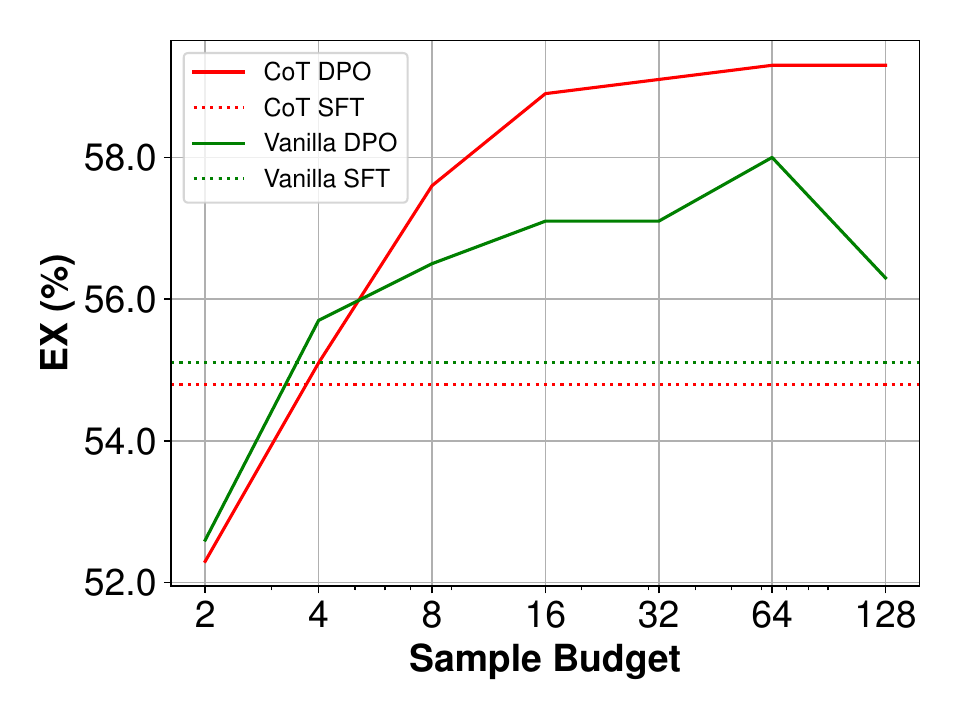}
    \subcaption{Pass@1}
  \end{subfigure}
  \hfill
  \begin{subfigure}[t]{0.31\linewidth}
    \centering
    \includegraphics[width=\linewidth]{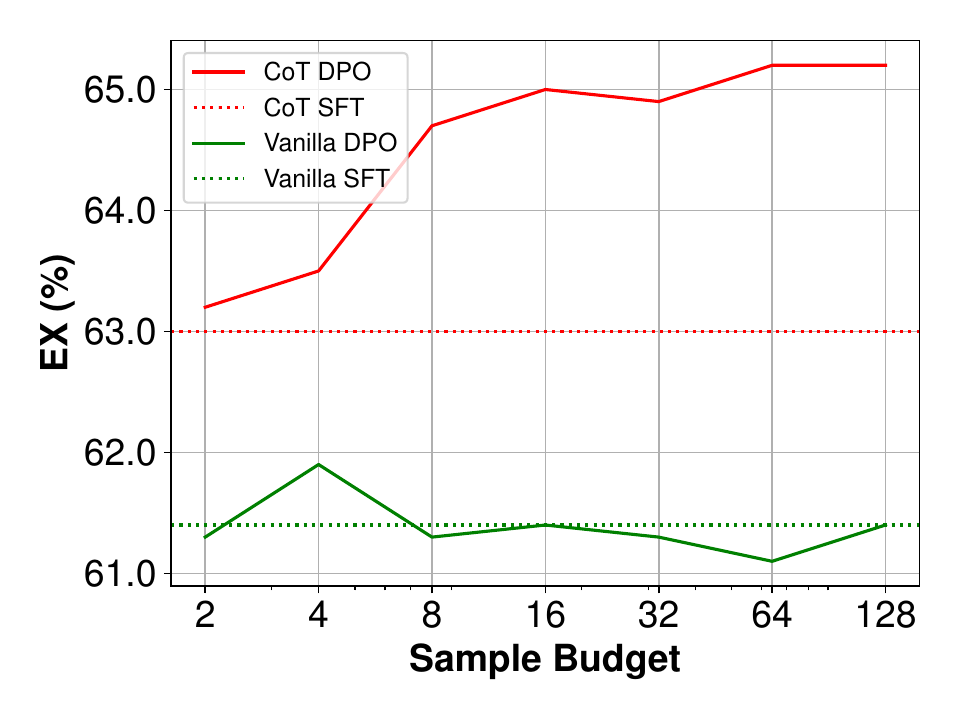}
    \subcaption{Maj@16}
  \end{subfigure}
  \caption{Model performance with different sample budgets in preference data collection tested under different inference strategies. The base model is Qwen2.5-7B-Instruct.}
  \label{fig:scalePrefFull}
\end{figure*}

Scaling behavior of performance on the CoT synthesis budget and sample budget of preference data collection under all inference strategies are complemented in Figure~\ref{fig:scaleSynFull} and Figure~\ref{fig:scalePrefFull}, respectively.

\begin{figure*}[ht]
  \centering
  \begin{subfigure}[t]{0.31\linewidth}
    \centering
    \includegraphics[width=\linewidth]{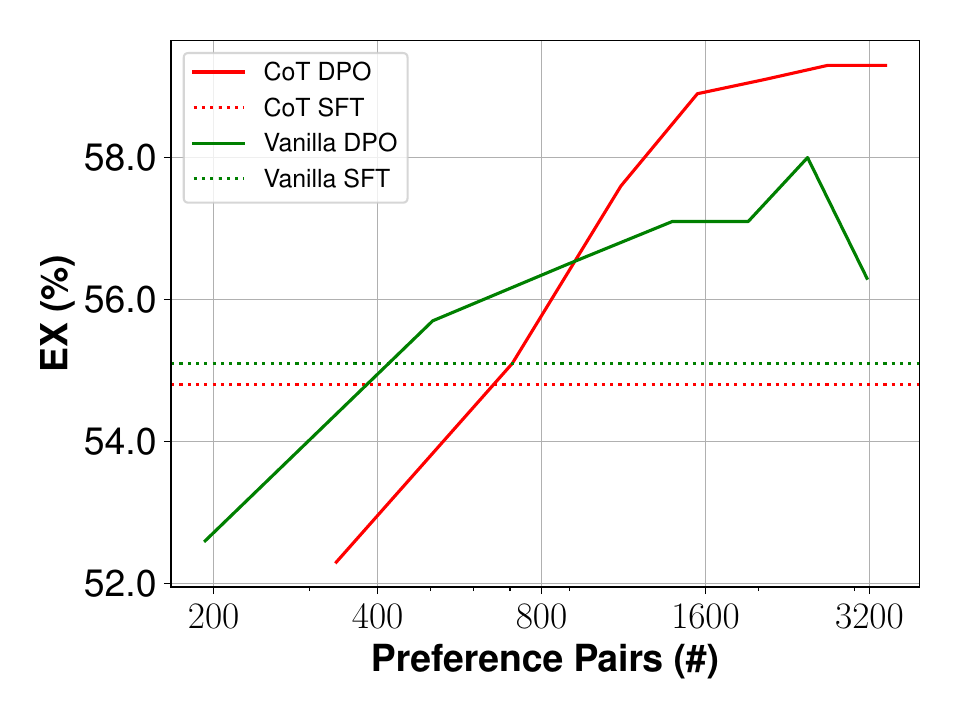}
    \subcaption{Greedy}
  \end{subfigure}
  \hfill
  \begin{subfigure}[t]{0.31\linewidth}
    \centering
    \includegraphics[width=\linewidth]{figures/PreferenceDataLog2.pdf}
    \subcaption{Pass@1}
  \end{subfigure}
  \hfill
  \begin{subfigure}[t]{0.31\linewidth}
    \centering
    \includegraphics[width=\linewidth]{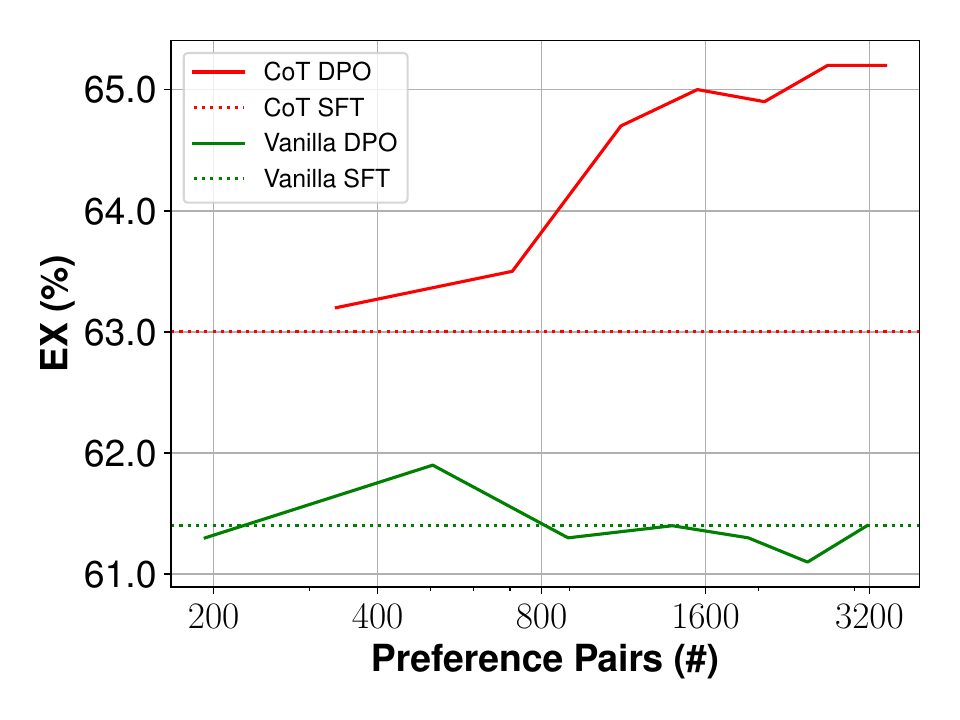}
    \subcaption{Maj@16}
  \end{subfigure}
  \caption{Model performance with different preference data sizes in DPO training tested under different inference strategies. The base model is Qwen2.5-7B-Instruct.}
  \label{fig:scalePrefLog}
\end{figure*}

It is noteworthy that performance saturation regarding sample budget in the preference data collection stage is mainly caused by the diminishing return of new preference pairs, as can be clearly seen from the log-scale plot Figure~\ref{fig:scalePrefLog}.

\section{Application Details}\label{apx:application}
\subsection{DTS-SQL}
DTS-SQL divides the Text-to-SQL task into two stages: Schema-Linking and SQL-Generation \citep{pourreza2024dts-sql}. Based on the code available in its repository, we construct the training and testing datasets for the SQL-Generation stage. (Since the original code is developed for the Spider dataset, we refer to the submitted \verb|DTS_SQL_BIRD_submission.py| file to account for incorporating Hints when constructing Prompts for the Bird dataset.) At this stage, we obtain a dataset without CoT. Subsequently, we utilize \texttt{gpt-4o-mini-2024-07-18} to generate 4 CoT paths, thereby creating two distinct datasets for training our two types of models.
\subsection{C3}
C3 stands for Clear Prompting, Calibration of Model Bias, and Consistency Output \citep{dong2023c3}. For the Clear Prompting and Calibration of Model Bias components, we use the C3 prompt templates, removing parts related to CoT. We run the Schema-Linking code to generate prompt inputs, which are paired with the correct SQL to form input-output pairs. Additionally, we use \texttt{gpt-4o-mini-2024-07-18} to generate 4 CoT paths, thereby creating two datasets for training our two models. After training the models, we test on Spider Dev using a Consistency Output subset of size 16, identical to Maj@16.

\input{Tables/ExamplePrompt}
\input{Tables/ExampleSynCoT}
\input{Tables/ExampleCoTSFT}
\input{Tables/ExampleCoTDPO}

\clearpage

\input{Tables/MainExp_Spider}
\input{Tables/MainExp_SpiderOther}
\input{Tables/MainExp_DrDB}
\input{Tables/MainExp_DrNLQ}
\input{Tables/MainExp_DrSQL}

\clearpage

\input{Tables/ErrorCategories}
\input{Tables/ErrorSamples}


\clearpage

\input{Tables/ErrorAnalysisRanking}
\input{Tables/ErrorAnalysisNewError}

\clearpage

\include{Tables/ExampleHackingPatterns}

\clearpage

\end{document}

%% file: RelatedWork.tex
\textbf{Text-to-SQL.}
The Text-to-SQL task aims to convert natural language questions into SQL queries for a given database.
With the emergence of large language models (LLMs) like GPT-4~\citep{openai2024@gpt4-turbo} and Gemini~\citep{DBLP:journals/corr/abs-2312-11805gemini}, the field has rapidly shifted towards leveraging LLMs for unified processing. These methods typically involve fine-tuning open-source language models~\cite{li2024codes, pourreza2024dts-sql, yang2024sense} or prompting closed-source LLMs through a multi-agent framework~\cite{pourreza2024din-sql, gao2024dali-sql, talaei2024chess, pourreza2024chase-sql, wang2024mac-sql}. Our work focuses on utilizing open-source LLMs and investigates the joint effectiveness of CoT reasoning and preference learning.





\textbf{Preference Optimization.} Preference optimization aims to align the model with the preferences of responses, which typically requires a compare-based training set.  
To this end, various methods have been proposed, from DPO~\citep{rafailov2024dpo} to its variants SimPO~\citep{DBLP:journals/corr/abs-2405-14734simpo}, KTO~\citep{DBLP:journals/corr/abs-2402-01306kto}, and IPO~\citep{DBLP:conf/aistats/AzarGPMRVC24ipo}. In this study, we employ DPO as our preference optimization algorithm. We note that only one study, SENSE~\citep{yang2024sense}, also employs DPO for optimizing Text-to-SQL models. However, their approach differs significantly from ours. Specifically, we follow the standard DPO pipeline, collecting preference data directly from the SFT model, whereas SENSE uses a small-scale model (1B parameters) to construct preference data for larger models, which may pose challenges in terms of transferability. Additionally, SENSE continues to use SQL queries as training labels, while this paper leverages CoT-style solutions.

\textbf{Learning from Execution Feedback.} Prior to the advent of LLMs, learning through execution feedback had already been widely applied to code-related tasks. CodeRL \citep{coderl} uses unit test results as rewards to optimize of pre-trained models through reinforcement learning. Other notable works include RLTF \citep{rltf}, StepCoder \citep{stepcoder}, and PseudoFeedback \citep{psudofeedback}. Nowadays, similar approaches are employed in the post-processing of general-purpose models to enhance their logical and coding capabilities \citep{qwenreport, dscoderv2}. However, existing general code models have not shown advantages in Text-to-SQL \citep{li2024codes}, and the substantial memory and CPU consumption of SQL execution makes online methods infeasible. We are the first to achieve stable and significant gains in Text-to-SQL by leveraging execution feedback and preference optimization.

%% file: sections/Pipeline.tex


\subsection{Overview}
\begin{figure}[t!]
  \centering
  \includegraphics[width=0.75\linewidth]{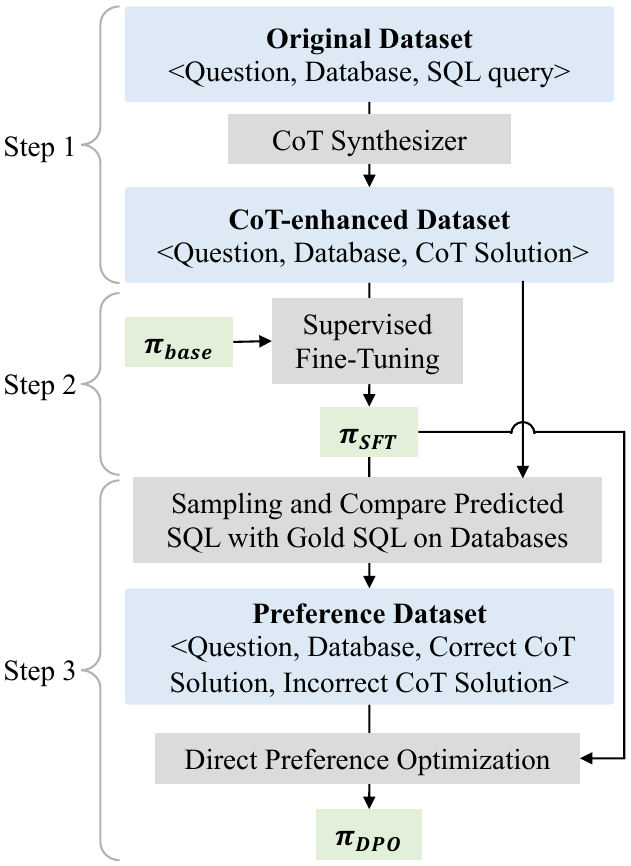}
  \caption{Overview of the proposed pipeline.}
  \label{fig:overview}
\end{figure}

Figure~\ref{fig:overview} illustrates our pipeline, which consists of three steps: (1) CoT synthesis, (2) supervised fine-tuning (SFT), and (3) direct preference optimization (DPO). Initially, we use an LLM as the CoT synthesizer to generate CoT solutions for a given Text-to-SQL dataset. With this CoT-enhanced dataset, we perform SFT and utilize feedback from databases to construct preference data pairs, followed by DPO. Further details are provided below.


\subsection{Chain-of-Thought Synthesis}
To minimize human annotation costs, we utilize an LLM as the CoT synthesizer to generate CoT solutions for existing Text-to-SQL datasets. Specifically, for each data sample <question, database, SQL qeury>, we employ GPT-4o-mini~\citep{openai2024@gpt4-mini} to generate $K$ diverse CoT solutions that demonstrate the step-by-step conversion of the question into the gold SQL query. Since the final answers (\emph{i.e.}, gold SQL queries) are provided to the CoT synthesizer, the generated CoT solutions are both reliable and accurate, making them suitable for subsequent SFT and DPO training. The prompts used for CoT synthesis and qualitative examples are detailed in Appendix~\ref{apx:cotSynthesis}.

\subsection{Supervised Fine-tuning}
Then, we perform supervised fine-tuning (SFT) using the CoT-enhanced dataset. The input prompt includes not only the question but also the database prompt, including information such as table names, column names, column data types, primary and foreign key relationships, etc. Details about the prompt construction can be found in Appendix~\ref{apx:datadetails}. The output sequences for SFT are LLM-synthesized CoT solutions. Formally, we denote the question as $q$, the database prompt as $d$, and the output CoT solution as $c$. The objective of SFT is guided by a conditional next-token prediction loss:
\begin{equation*}
\mathcal{L_{SFT}} = - \mathbb{E}_{(q,d,c) \sim D_{S}} [\log \pi_{base}(c \mid q, d)],
\end{equation*}
where $D_{S}$ denotes the CoT-enhanced training set, and $\pi_{base}$ refers the base model. After fine-tuning, we obtain model $\pi_{SFT}$, serving as the reference model for the subsequent DPO.





\subsection{Direct Preference Optimization}
Then, we apply DPO \citep{rafailov2024dpo} to further improve the Text-to-SQL capabilities of the SFT model. A brief overview of the DPO algorithm, including its learning objective, implicit reward mechanism, and token-level credit assignment, is provided in Appendix~\ref{apx:dpo}.



\textbf{Preference Dataset Construction.}
DPO requires a preference dataset to teach the model to distinguish between correct and incorrect responses. In this study, each preference data sample consists of a quadruple: <question, database, correct CoT solution, incorrect CoT solution>. To construct this dataset, we sample CoT solutions from the supervised fine-tuned (SFT) model and use database feedback to determine the correctness of each sampled CoT solution. Specifically, for each data sample in the training set, we generate $N$ distinct CoT solutions from the reference model $\pi_{SFT}$. We then extract predicted SQL queries from these solutions using regular expressions and execute both the predicted and gold SQL queries on the corresponding databases. A sampled CoT solution is labeled as correct only if the execution results (\emph{e.g.}, records that satisfy certain filter conditions) of the predicted SQL query completely match those of the gold SQL query; otherwise, it is labeled as incorrect. For data samples with multiple correct and incorrect CoT solutions, we randomly select one correct and one incorrect solution to form a preference pair for DPO training.




\textbf{DPO Learning Objective.}
The goal of DPO is to maximize the margin between the log-likelihood of the correct and incorrect responses while ensuring the model remains close with the reference policy. Formally, for a question $q$ and its corresponding database prompt $d$, let $c^{+}$ and $c^{-}$ represent the correct and incorrect CoT solutions, respectively. The DPO loss is defined as: 
\begin{equation*}
\scalebox{0.8}{$
\begin{aligned}
&\mathcal{L_{DPO}} = - \mathbb{E}_{(q,d,c^{+},c^{-}) \sim D_{P}} \\
&\log \sigma\left(\beta \log\frac{\pi_{DPO}(c^{+}\mid q,d)}{\pi_{SFT}(c^{+}\mid q,d)} -\beta\log\frac{\pi_{DPO}(c^{-} \mid q,d)}{\pi_{SFT}(c^{-}\mid q,d)}\right),
\end{aligned}$
}
\end{equation*}
where $\mathcal{D}_{P}$ is the preference dataset, $\sigma(\cdot)$ is the sigmoid function, $\beta$ is a hyperparameter that controls the penalty strength imposed by the KL divergence, $\pi_{DPO}$ represents the DPO model, which is initialized from the SFT model at the start of training.



%% file: sections/MainResult.tex
\subsection{Main Results}
\input{Tables/MainExp}
The results on the Bird benchmark are presented in Table~\ref{tab:model-comparison}. Results on the Spider benchmark and its robustness variants are deferred to Appendix~\ref{apx:spider}. 


\textbf{Vanilla Models Struggle to Achieve Performance Gains in the DPO Stage.} 
For Greedy and Maj@16, DPO gains are minimal or even negative, and improvements for the Pass@1 strategy are marginal as well. Models showing performance degradation after DPO tend to worsen as training progresses, indicating that directly applying DPO in the vanilla setting may impair performance.


\textbf{Models with Synthetic CoT Achieve Stable and Significant Gains in the DPO Stage.}
These gains are evident across all base models and inference strategies. Even when CoT models outperform vanilla models in the SFT stage, the performance gains from CoT remain consistently significant during the subsequent DPO stage. Moreover, this phenomenon persists even when replacing GPT-4o-mini with much weaker LLMs (e.g., Qwen2.5-1.5B-Instruct) to synthesize CoT solutions, confirming the benefit of the CoT solution style per se for DPO, as discussed in Appendix~\ref{apx:qualityablation}.



\textbf{Synthesized CoT plus DPO Exhibit Higher Performance Ceilings.} 
As shown in the $\Delta$EX column in Table~\ref{tab:model-comparison}, all base models trained with CoT-enhanced data through the SFT and DPO pipeline achieved higher performance ceilings. This indicates that integrating CoT synthesis with DPO is highly effective for the Text-to-SQL task, offering a promising new approach to developing improved Text-to-SQL models. With our straightforward pipeline, Qwen2.5-14B-Instruct achieves the second-best performance on the Bird development set among all open-source models, despite having significantly fewer parameters, as shown in Table~\ref{tab:BestModels}.

\input{Tables/BestLocalModel}

\subsection{Error Analysis}\label{sec:errorAnalysis}
\input{Tables/ErrorAnalysis}
To understand how DPO affects model generation and identify areas where CoT enhances DPO, we meticulously analyze errors made by Qwen2.5-7B-Instruct using the greedy decoding strategy. As shown in Table~\ref{table:ErrorAnalysis}, errors are classified into our pre-defined categories, with correction rates before and after DPO presented for each. Full explanations and examples of our classification criteria, along with detailed error statistics, are available in Appendix~\ref{apx:analysis}. We find that, except for Syntax Errors, synthesized CoT improves DPO's ability to correct errors across all other categories.



\textbf{DPO Excels at Correcting Errors Caused by Ignoring Details.} Error types with a correction rate exceeding 25\% during DPO are highlighted in bold. These errors primarily stem from the model's failure to pay sufficient attention to details. For instance, \underline{NULL/DISTINCT} (Fix 40.0\% with CoT) errors arise when the model overlooks missing or duplicate values in the relevant columns, leading to incorrect query results. Similarly, \underline{Column Sequence} (Fix 42.9\% with CoT) errors occur when the model does not return columns in the order specified in the question. 



\textbf{CoT Provides Logical Guidance to Enhance DPO's Error Correction.} Error types with significant improvement often benefit from CoT's reasoning nature. For example, \underline{JOIN} errors (Fix 32.1\%, +16.5\% with CoT) frequently occur when the required information resides in two tables that need to be joined via a third intermediate table. Without CoT, the model might attempt to directly join the two tables on an incorrect key. CoT's step-by-step logic clarifies these joins, aiding DPO in error correction. Conversely, for intuitive errors like \underline{Hallucination} (Fix 27.2\%, +3.5\% with CoT) and \underline{Date} (Fix 30.4\%, +7.3\% with CoT), CoT's impact is smaller. These errors stem from mismatches between SQL queries and database information, which vanilla DPO already can address effectively.
More analysis about DPO's effect with or without CoT reasoning can be found in Appendix~\ref{apx:analysis}.



%% file: Tables/MainExp.tex
\definecolor{darkgreen}{RGB}{0,150,0}
\begin{table*}[t!]
    \centering
\begin{adjustbox}{max width=0.85\textwidth}
    \begin{tabular}{c  c | c c | c c | c c | c}
        \toprule
         & \multirow{3}{*}{\textbf{Model}} & \multicolumn{6}{c}{\textbf{Bird Dev}} & \\ \cline{3-9}
         & & \multicolumn{2}{c}{Greedy} & \multicolumn{2}{c}{Pass@1} & \multicolumn{2}{c |}{Maj@16} & \multirow{2}{*}{$\Delta$EX} \\ \cline{3-8}
         & & SFT & DPO & SFT & DPO & SFT & DPO & \\ \midrule
        \multicolumn{9}{c}{\textbf{General Models}} \\ \midrule
        
        & Deepseek-llm-7b-chat & 51.8 & 51.2 (\textcolor{darkgreen}{-0.6})& 47.9  & 49.1 (\textcolor{red}{+1.3})& \text{54.5} & 54.3 (\textcolor{darkgreen}{-0.3}) & -\\ 
        & Meta-Llama-3.1-8B-Instruct & 59.0 & 59.8 (\textcolor{red}{+0.8}) & 56.1  & 57.2 (\textcolor{red}{+1.1}) & \text{61.4} & 60.8 (\textcolor{darkgreen}{-0.6}) & -\\ 
        & Qwen2.5-7B-Instruct & 58.8 & 59.0 (\textcolor{red}{+0.2}) & 55.1  & 55.7 (\textcolor{red}{+0.6}) & \text{61.4} & 60.6 (\textcolor{darkgreen}{-0.8}) & -\\ 
        \multirow{-4}{*}{\textbf{Vanilla}} & Qwen2.5-14B-Instruct & 64.3 & 63.5 (\textcolor{darkgreen}{-0.8}) & 62.3  & 62.6 (\textcolor{red}{+0.3}) &  64.6 & \text{65.1} (\textcolor{red}{+0.5}) & - \\
        
        \rowcolor{cyan!20}
        & Deepseek-llm-7b-chat & 54.3 & 55.9 (\textcolor{red}{+1.6})& 51.9  & 54.8 (\textcolor{red}{+2.9})& 59.1 & \text{61.0} (\textcolor{red}{+1.9}) & 54.5 $\rightarrow$ 61.0 (\textbf{\textcolor{red}{+6.5}})\\ 
        \rowcolor{cyan!20}
        & Meta-Llama-3.1-8B-Instruct & 56.8 & 61.2 (\textcolor{red}{+4.4}) & 57.5  & 59.0 (\textcolor{red}{+1.5}) & 60.2 & \text{61.9} (\textcolor{red}{+1.7}) & 61.4 $\rightarrow$ 61.9 (\textbf{\textcolor{red}{+0.5}}) \\ 
        \rowcolor{cyan!20}
        & Qwen2.5-7B-Instruct & 57.4 & 61.9 (\textcolor{red}{+4.5}) & 54.8  & 59.2 (\textcolor{red}{+4.4}) & 63.0 & \text{64.9} (\textcolor{red}{+1.9}) & 61.4 $\rightarrow$ 64.9 (\textbf{\textcolor{red}{+3.5}})\\ 
        \rowcolor{cyan!20}
        \multirow{-4}{*}{\textbf{Syn CoT}} & Qwen2.5-14B-Instruct & 63.2 & 65.3 (\textcolor{red}{+2.1}) & 61.8  & 64.7 (\textcolor{red}{+2.9}) &  65.4 & \text{67.1} (\textcolor{red}{+1.7}) & 64.6 $\rightarrow$ 67.1 (\textbf{\textcolor{red}{+2.5}})\\
        
        \midrule
        \multicolumn{9}{c}{\textbf{Coder Models}} \\ \midrule
        
        & Deepseek-coder-6.7b-instruct & 60.6 & 60.9 (\textcolor{red}{+0.3}) & 56.9 & 58.8 (\textcolor{red}{+1.9}) & 59.8 & \text{61.0} (\textcolor{red}{+1.2}) & - \\
        & CodeLlama-7b-Instruct-hf & 57.0 & 55.7 (\textcolor{darkgreen}{-1.3}) & 54.3 & 55.5 (\textcolor{red}{+1.2}) & \text{59.1} & 58.5 (\textcolor{darkgreen}{-0.6}) & - \\ 
        & CodeLlama-13b-Instruct-hf & 60.0 & 60.2 (\textcolor{red}{+0.2})& 56.7  & 57.9 (\textcolor{red}{+1.2})& 61.9 & \text{62.0} (\textcolor{red}{+0.1}) & - \\
        \multirow{-4}{*}{\textbf{Vanilla}} & Qwen2.5-Coder-7B-Instruct & 61.6 & 61.3 (\textcolor{darkgreen}{-0.3}) & 59.4 & 60.6 (\textcolor{red}{+1.2}) & 61.3 & \text{62.7} (\textcolor{red}{+1.4}) & - \\
        
        \rowcolor{cyan!20}
        & Deepseek-coder-6.7b-instruct & 61.5 & 63.8 (\textcolor{red}{+2.3}) & 59.9 & 62.3 (\textcolor{red}{+4.5}) & 64.3 & \text{65.4} (\textcolor{red}{+1.1}) & 59.8 $\rightarrow$ 65.4 (\textbf{\textcolor{red}{+5.6}}) \\
        \rowcolor{cyan!20}
        & CodeLlama-7b-Instruct-hf & 58.2 & 61.3 (\textcolor{red}{+3.1}) & 56.9 & 60.4 (\textcolor{red}{+3.5}) & 60.2 & \text{61.9} (\textcolor{red}{+1.7}) & 59.1 $\rightarrow$ 61.9 (\textbf{\textcolor{red}{+2.8}}) \\ 
        \rowcolor{cyan!20}
        & CodeLlama-13b-Instruct-hf & 62.0 & 63.9 (\textcolor{red}{+1.9})& 59.8  & 62.5 (\textcolor{red}{+2.7})& 63.6 & \text{65.8} (\textcolor{red}{+2.2}) & 61.9 $\rightarrow$ 65.8 (\textbf{\textcolor{red}{+3.9}})\\
        \rowcolor{cyan!20}
        \multirow{-4}{*}{\textbf{Syn CoT}} & Qwen2.5-Coder-7B-Instruct & 60.8 & 63.4 (\textcolor{red}{+2.6}) & 59.1 & 62.8 (\textcolor{red}{+3.7}) & 62.5 & \text{64.1} (\textcolor{red}{+1.6}) & 61.3 $\rightarrow$ 64.1 (\textbf{\textcolor{red}{+2.8}}) \\ \midrule
        
        \multicolumn{9}{c}{\textbf{SQL-Specialized Models}} \\ \midrule
        
        & CodeS-7B & 56.8 & 56.6 (\textcolor{darkgreen}{-0.2}) & 53.7 & 54.6 (\textcolor{red}{+0.9}) & \text{58.1} & 58.0 (\textcolor{darkgreen}{-0.1}) & -\\ 
        \multirow{-2}{*}{\textbf{Vanilla}} & CodeS-15B & 58.3 & 58.2 (\textcolor{darkgreen}{-0.1}) & 55.6 & 56.2 (\textcolor{red}{+0.6}) & \text{60.2} & 59.1 (\textcolor{darkgreen}{-1.1}) & -\\
        
        \rowcolor{cyan!20}
        & CodeS-7B & 56.7 & 57.5 (\textcolor{red}{+0.8}) & 54.2 & 55.3 (\textcolor{red}{+1.1}) & 60.2 & \text{61.7} (\textcolor{red}{+1.5}) & 58.1 $\rightarrow$ 61.7 (\textbf{\textcolor{red}{+2.6}}) \\ 
        \rowcolor{cyan!20}
        \multirow{-2}{*}{\textbf{Syn CoT}} & CodeS-15B & 58.6 & 61.1 (\textcolor{red}{+2.5}) & 56.6 & 60.5 (\textcolor{red}{+3.9}) & 62.4 & \text{63.2} (\textcolor{red}{+0.8}) & 60.2 $\rightarrow$ 63.2 (\textbf{\textcolor{red}{+3.0}})\\
        \bottomrule
    \end{tabular}
\end{adjustbox}
    \caption{Model performance on the Bird development set. \textbf{Vanilla}: SFT and DPO on the original Bird training set; \textbf{Syn CoT}: SFT and DPO on the CoT-enhanced training set; \textbf{$\Delta$EX}: The performance difference in EX between ``Syn CoT + DPO'' and ``Vanilla + SFT'' when using the same base model.}
    \label{tab:model-comparison}
\end{table*}

%% file: Tables/BestLocalModel.tex
\definecolor{darkgreen}{RGB}{0,150,0}
\begin{table}[h!]
    \centering
\begin{adjustbox}{max width=0.8\columnwidth}
    \begin{tabular}{l | c | c | c }
    \toprule
       \textbf{Model}  & \textbf{Date} & \textbf{Size} &  \textbf{EX} \\
    \midrule
        ExSL+granite-34b-code	& Oct 27, 2024	& 34B	& 72.43 \\
    \midrule
    \textbf{Qwen2.5-14B-Instruct} & \multirow{2}{*}{\textbf{Nov 12, 2024}}	& \multirow{2}{*}{\textbf{14B}}	& \multirow{2}{*}{\textbf{67.10}} \\ 
    \textbf{+ Syn CoT + DPO (Ours)}	& & & \\
    \midrule
    XiYanSQL-QwenCoder-32B & Jan 09, 2025 & 32B & 67.01 \\
    \midrule
    MSL-SQL+DeepSeek-V2.5 &	Oct 10, 2024 &	236B &	66.82 \\
    \midrule
    ExSL+granite-20b-code & May 14, 2024 & 20B & 65.38 \\
    \midrule
    SFT CodeS-15B &	Oct 12, 2023 &	15B & 58.47 \\
    \bottomrule
    \end{tabular}
\end{adjustbox}
\caption{Comparing with top open-source model-based Text-to-SQL methods on Bird development set.}
\label{tab:BestModels}
\end{table}

%% file: Tables/ErrorAnalysis.tex
\definecolor{darkgreen}{RGB}{0,150,0}
\begin{table*}[t]
\centering
\begin{adjustbox}{max width=0.8\textwidth}
\begin{tabular}{c p{3.5cm} l | c | c | c}
\toprule
\textbf{Category} & \textbf{Description} & \textbf{Type} & \textbf{Vanilla DPO Fix (\%)} & \textbf{Syn CoT DPO Fix (\%)} & \textbf{$\Delta$(\%)} \\
\midrule
\multirow{1}{*}{External Knowledge} 
    & Neglect of hints
    & \textbf{[A1] EK} 
    & 0.0 (0/3) 
    & \textbf{37.5 (3/8)} 
    & \textcolor{red}{+37.5}\\
\midrule
\multirow{6}{*}{Schema Linking} 
    & \multirow{6}{=}{Fails to match the question with its concerning table and columns}
    & [B1] Table 
    & 13.0 (12/92) 
    & 15.9 (11/69) 
    & \textcolor{red}{+2.9}
    \\
    & & \textbf{[B2] JOIN}
    & 15.6 (12/77)
    & \textbf{32.1 (18/56)}
    & \textcolor{red}{+16.5} \\
    & & [B3] Column
    & 10.3 (7/68)
    & 16.1 (10/62)
    & \textcolor{red}{+5.8} \\
    & & \textbf{[B4] Hallucination}
    & 23.7 (14/59)
    & \textbf{27.2 (28/102)}
    & \textcolor{red}{+3.5} \\
    & & [B5] Condition
    & 16.7 (10/60)
    & 23.2 (16/69) 
    & \textcolor{red}{+6.5} \\
    & & \textbf{[B6] NULL/DISTINCT}
    & 9.7 (3/31)
    & \textbf{40.0 (12/30)}
    & \textcolor{red}{+30.3} \\
\midrule
\multirow{2}{*}{Value Retrieval}
    & \multirow{2}{=}{Mismatch of condition with its storage format}
    & [C1] String/Number 
    & 4.5 (1/22) 
    & 21.1 (4/19) 
    & \textcolor{red}{+16.6}
    \\
    & & \textbf{[C2] Date}
    & 23.1 (6/26)
    & \textbf{30.4 (7/23)}
    & \textcolor{red}{+7.3} \\
\midrule
\multirow{3}{*}{Operation}
    & \multirow{3}{=}{Misunderstands required operation in the question.}
    & [D1] Mathematical Formula 
    & 13.3 (6/45)
    & 18.2 (8/44) 
    & \textcolor{red}{+4.9}
    \\
    & & [D2] Aggregation
    & 6.7 (5/75)
    & 18.2 (12/66) 
    & \textcolor{red}{+11.5} \\
    & & [D3] Complex Operation
    & 5.6 (1/18)
    & 12.5 (3/24) 
    & \textcolor{red}{+6.9} \\
\midrule
\multirow{4}{*}{Information}
    & \multirow{4}{=}{Fails to organize information in the right way}
    & [E1] Redundant/Incomplete 
    & 11.8 (4/34) 
    & 19.2 (5/26) 
    & \textcolor{red}{+7.4}
    \\
    & & \textbf{[E2] Column Sequence}
    & 0 (0/5)
    & \textbf{42.9 (3/7)} & \textcolor{red}{+42.9}
    \\
    & & [E3] ORDER BY/LIMIT
    & 9.1 (1/11)
    & 12.5 (1/8) 
    & \textcolor{red}{+3.4} \\
    & & \textbf{[E4] Format}
    & \textbf{66.7 (2/3)}
    & \textbf{33.3 (2/6)} & \textcolor{darkgreen}{-33.4}
    \\
\midrule
Syntax Error 
    & Inexecutatble SQL 
    & [F1] Syntax 
    & 14.3 (2/14) 
    & 13.3 (2/15)
    & \textcolor{darkgreen}{-1.0}\\
\bottomrule
\end{tabular}
\end{adjustbox}
\caption{Comparison of Vanilla and Syn CoT DPO correction capability across error types on Bird development set (greedy). The base model is Qwen2.5-7B-Instruct.}
\label{table:ErrorAnalysis}
\end{table*}

%% file: sections/CoTHelpDPO.tex
To further investigate why CoT reasoning is essential for DPO, we conduct a series of analyses using Qwen2.5-7B-Instruct on the Bird benchmark. 

\subsection{Better Discriminative Ability}

\begin{figure*}[ht]
    \centering
    \begin{minipage}{0.31\linewidth} 
        \centering
    \includegraphics[width=\columnwidth]{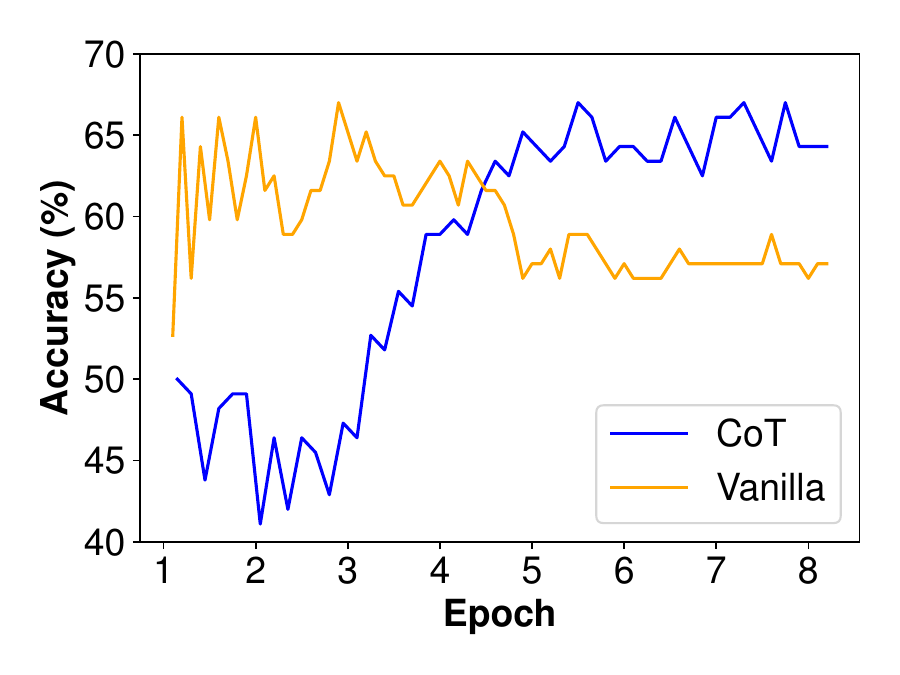}
  \caption{Comparison of model's discriminative ability during DPO (measured by classification accuracy on curated evaluation set). }
  \label{fig:discriminator}
    \end{minipage}
    \hfill 
    \begin{minipage}{0.62\linewidth} 
        \centering
          \begin{subfigure}[ht]{0.49\columnwidth}
            \centering
            \includegraphics[width=\columnwidth]{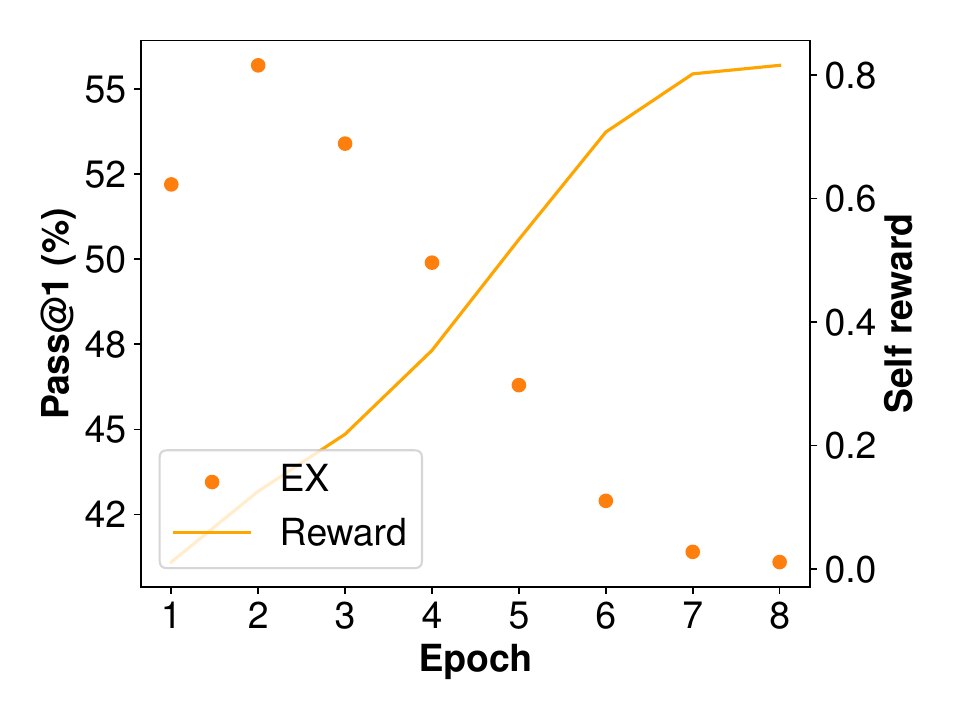}
            \subcaption{Vanilla}
            \label{fig:selfreward-a1}
          \end{subfigure}
          \hfill
          \begin{subfigure}[ht]{0.49\columnwidth}
            \centering
            \includegraphics[width=\columnwidth]{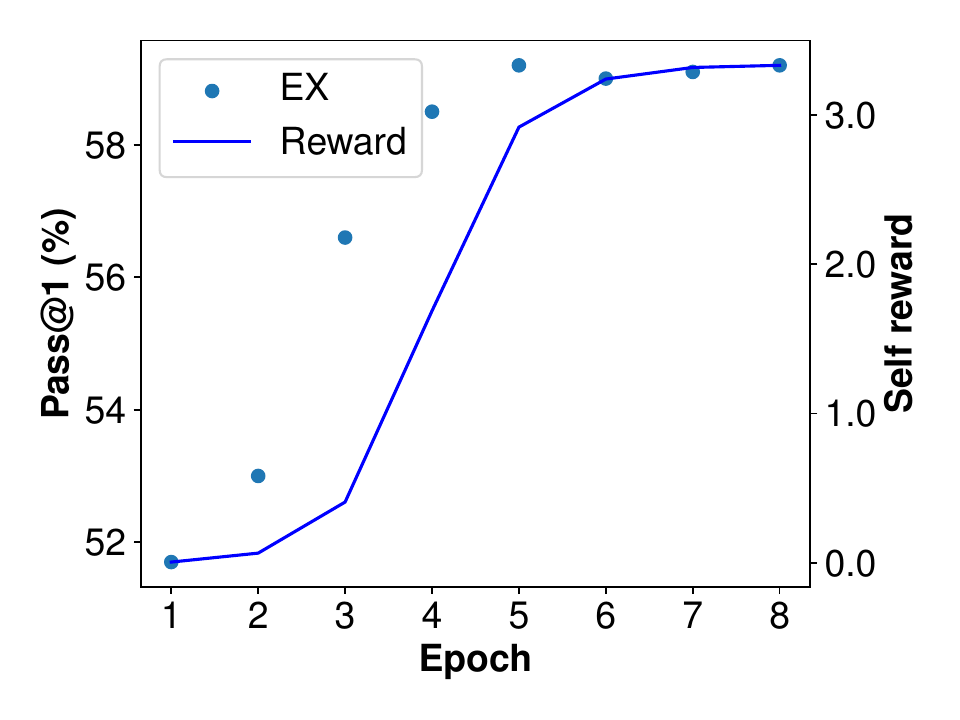}
            \subcaption{Syn CoT}
            \label{fig:selfreward-b}
          \end{subfigure}
          \caption{Comparison of model's self-assessed performance (average implicit reward policy model given to its own roll-outs) and real performance (EX) on Bird development set (Pass@1) during DPO training.}
          \label{fig:selfreward}
    \end{minipage}
\end{figure*}

\begin{figure}[t]
  \centering
  \begin{subfigure}[t]{0.493\linewidth}
    \centering
    \includegraphics[width=\linewidth]{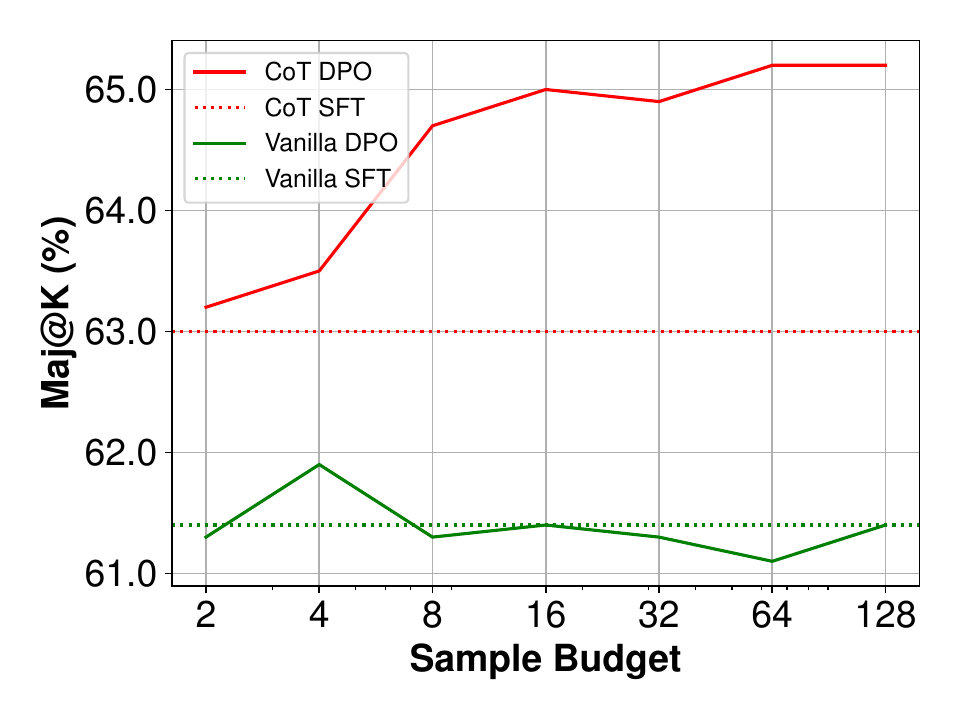}
    \subcaption{Preference Data}
    \label{fig:scaleMaj-b}
  \end{subfigure}
  \hfill
  \begin{subfigure}[t]{0.493\linewidth}
    \centering
    \includegraphics[width=\linewidth]{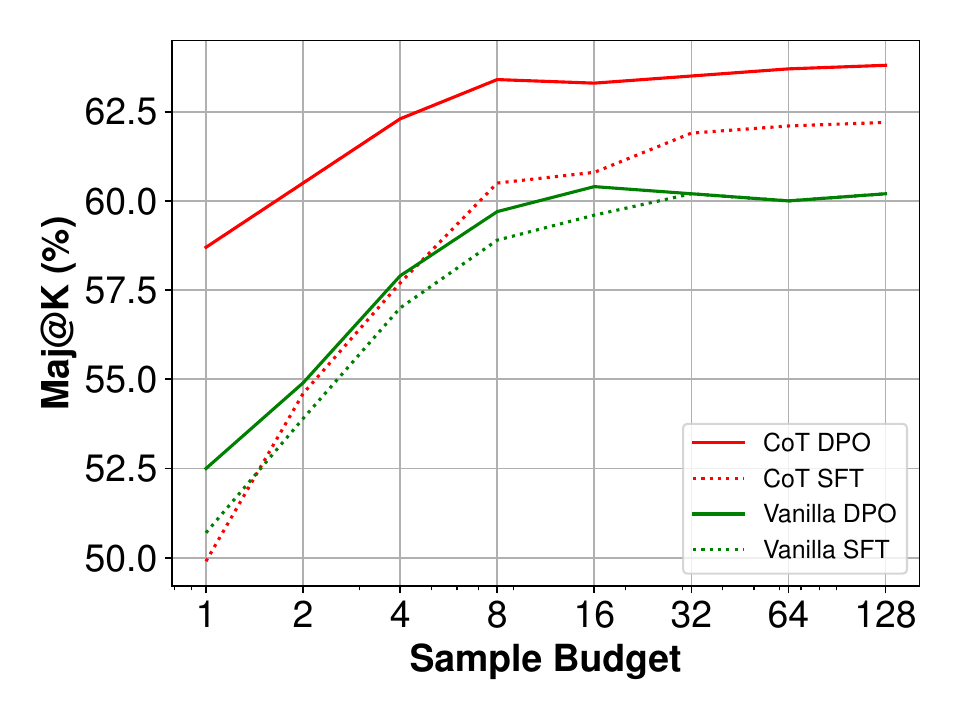}
    \subcaption{Inference-time}
    \label{fig:scaleMaj-c}
  \end{subfigure}
  \caption{Model performance with different sample budget $K$ in each stage (Maj@K). Qwen2.5-7B-Instruct is used as the base model.}
  \label{fig:scaleMaj}
\end{figure}


We design an evaluation preference dataset to compare the discriminative capabilities of Vanilla and Syn CoT models after DPO. Construction details are in Appendix~\ref{apx:design}. During DPO training, we assess the model's ability to select the correct response by classification accuracy based on implicit reward. The results in Figure~\ref{fig:discriminator} show that \textbf{CoT reasoning enables DPO models to achieve more stable and superior discriminative ability}.



\input{Tables/TrainRobGreedy}

\subsection{Better Training Stability}
From a different perspective, we examine the discrepancy between the models' self-evaluation rewards for their generated outputs and the actual execution accuracy on the development set. Specifically, at the end of each epoch during DPO training, we sample responses using the current checkpoint and calculate the average implicit reward, as defined in Appendix~\ref{apx:implicit_reward}. Figure~\ref{fig:selfreward} illustrates the results for both the Syn CoT and Vanilla models.

\textbf{Vanilla Model is Susceptible to Reward Hacking.} As shown in Figure~\ref{fig:selfreward} (a), during DPO training, the performance of the Vanilla model initially peaks but then drops. Despite this drop, its self-reward scores continue to rise, indicating that DPO's underlying reward model mistakenly believes its outputs are improving, thereby demonstrating the phenomenon of reward hacking.


\textbf{In Contrast, The Self-reward of Syn CoT Model Reflects Its Actual Performance.} As shown in Figure~\ref{fig:selfreward} (b), In the early stages of DPO training, as the model's capabilities improve, its self-reward scores also increase. In the later stages, when the model's performance saturates, the self-reward scores stabilize rather than exhibiting the reward-hacking observed in the Vanilla model. This suggests that the Syn CoT model can recognize when its outputs are no longer improving and avoids overestimating its performance.

\textbf{Output Statistics.} We compare the statistical changes in the outputs of the two models (Vanilla vs. Syn CoT) after DPO in Table~\ref{tab:RewardHackingStat}. For the Vanilla model, the average output length increased significantly (+26.6\%) after DPO, accompanied by a substantial rise in the proportion of invalid SQL queries (+17.6\%).
In contrast, the Syn CoT model exhibited minimal changes in SQL length (+2\%), and a reduced non-executable rate (-2.5\%).

\input{Tables/RewardHackingStat}

\begin{figure}[htbp]
  \centering
  \includegraphics[width=0.9\columnwidth]{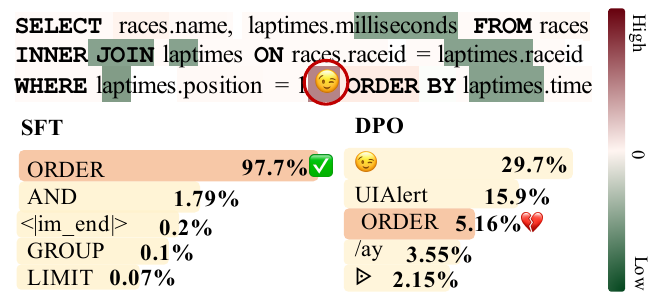}
  \caption{A case of reward hacking from Bird development set. The background color indicates token rewards.}
  \label{fig:RewardHackingCase}
\end{figure}

\textbf{A Case of Reward Hacking in Vanilla + DPO.} Furthermore, we show a classical non-executable output generated by the Vanilla DPO model in Figure~\ref{fig:RewardHackingCase}. In this example, the model assigns an exceptionally high reward to an incorrect token, even though this token does not appear in the training dataset. This indicates a clear occurrence of reward hacking. Other prominent reward hacking patterns are presented in Appendix~\ref{sec:RHPatterns}.

\textbf{Syn CoT Enhances DPO Robustness to Hyperparameter $\beta$.} As shown in Table~\ref{table:RobBeta}, varying the hyperparameter $\beta$ reveals that the DPO training of the vanilla model is highly sensitive to changes in $\beta$. In contrast, the Syn CoT model demonstrates notable robustness, consistently delivering strong performance despite these perturbations.

\subsection{Better Scalability}

In this section, we discuss the scalability of our proposed pipeline to provide guidance for best practices. We report the Maj@K results in Figure~\ref{fig:scaleMaj}, and results for Greedy and Pass@16 are provided in Appendix~\ref{apx:scaling}.
\textbf{Preference Data Collection (Figure~\ref{fig:scaleMaj-b}).} For vanilla models, collecting additional preference data does not enhance performance during the DPO stage. In contrast, increasing the sample budget for Syn CoT models consistently results in continuous performance improvements with DPO. 
\textbf{Inference-time Computation (Figure~\ref{fig:scaleMaj-c}).} 
By increasing the inference budget, both vanilla and Syn CoT models show improved performance, although the gains eventually stabilize. When the budget is limited, DPO offers a significant advantage for Syn CoT models over Vanilla models. For instance, ``CoT + DPO'' with 2 samples achieves the same performance as the ``Vanilla + DPO'' with 16 samples.

%% file: Tables/TrainRobGreedy.tex
\definecolor{darkgreen}{RGB}{0,150,0}
\begin{table}[htb]
    \centering
\begin{adjustbox}{max width=\columnwidth}
    \begin{tabular}{c | c | c c c }
    \toprule
    Model & SFT & $\beta=0.05$ & $\beta=0.1$ & $\beta=0.2$ \\
    \midrule
    Vanilla & 58.8 & 57.6 (\textbf{\textcolor{darkgreen}{-1.2}}) & 59.0 (\textbf{\textcolor{red}{+0.2}}) & 57.8 (\textbf{\textcolor{darkgreen}{-1.0}})\\
    Syn CoT & 57.4 & 61.6 (\textbf{\textcolor{red}{+4.2}}) & 61.9 (\textbf{\textcolor{red}{+4.5}})& 61.9 (\textbf{\textcolor{red}{+4.5}}) \\
    \bottomrule
    \end{tabular}
\end{adjustbox}
\caption{DPO performance across different $\beta$ values (greedy). The base model is Qwen2.5-7B-Instruct.}
\label{table:RobBeta}
\end{table}

%% file: Tables/RewardHackingStat.tex
\definecolor{darkgreen}{RGB}{0,150,0}
\begin{table}[htb]
    \centering
\begin{adjustbox}{max width=0.8\columnwidth}
    \begin{tabular}{c | c  c  c  c }
    \toprule
       \multirow{2}{*}{\textbf{Metric}} & \multicolumn{2}{c}{\textbf{Vanilla}} & \multicolumn{2}{c}{\textbf{Syn CoT}} \\
       & SFT & DPO & SFT & DPO \\
    \midrule
        \textbf{SQL Character Number} & 173 & 219 & 178 & 182 \\
       \textbf{Non-executable SQL(\%)} & 2.0 & 19.6 & 7.6 & 5.1 \\
    \bottomrule
    \end{tabular}
\end{adjustbox}
\caption{Statistics on predicted SQL queries.}
\label{tab:RewardHackingStat}
\end{table}

%% file: Tables/Transferability.tex
\definecolor{darkgreen}{RGB}{0,150,0}
\begin{table}[h!]
    \centering
\begin{adjustbox}{max width=0.9\columnwidth}
    \begin{tabular}{l l l l}
    \toprule
        \multicolumn{2}{c}{\textbf{Bird}} & \multicolumn{2}{c}{\textbf{Spider}} \\
    \midrule
        \textbf{Method} & \textbf{EX} & \textbf{Method} & \textbf{EX} \\
    \midrule
        DTS-SQL & 47.7 & C3-SQL & 80.5\\
        \quad + Syn CoT + DPO & 55.9 (\textbf{\textcolor{red}{+8.2}}) & \quad + Syn CoT + DPO & 81.9 (\textbf{\textcolor{red}{+1.4}}) \\
    \bottomrule
    \end{tabular}
\end{adjustbox}
\caption{Performance improvements of two Text-to-SQL frameworks when incorporating our trained models.}
\label{tab:OtherPipeline}
\end{table}

%% file: Tables/ExamplePrompt.tex
\begin{table*}[htbp]
\centering
\begin{tcolorbox}[
    colback=white, 
    colframe=black, 
    title=System, 
    fonttitle=\bfseries, 
    width=\textwidth, 
    boxrule=1pt, 
    rounded corners, 
]

\begin{tabularx}{\textwidth}{X}
You are a senior data analyst who expertise at structural query language (SQL). Given a question made by front end employees and targeted database schema, you are asked to translate that question into SQLite query with detailed explanation.
\newline
\newline
Additionally, the input will be accompanied with a reference solution from your colleagues, which may or may not be correct. This extra information intents to help you to formulate your answer,  and you are asked not to mention reference solution in any form. 
\newline
\newline
To facilitate SQL extraction with regular expression, the SQL in your answer should be expressed in a Markdown code block with proper highlight. For example, \\
\verb|`|\verb|`|\verb|`| SQL \\
    SELECT * FROM database; \\
\verb|`|\verb|`|\verb|`|
\end{tabularx}
\end{tcolorbox}
\caption{The system message we use when prompting \texttt{gpt-4o-mini-2024-07-18} to synthesize Chain-of-Thought reasoning solutions for publicly available Text-to-SQL datasets.}
\label{tab:prompt}
\end{table*}

%% file: Tables/ExampleSynCoT.tex
\begin{table*}[htbp]
\centering
\begin{tcolorbox}[
    colback=white, 
    colframe=black, 
    title=Response, 
    fonttitle=\bfseries, 
    width=\textwidth, 
    boxrule=1pt, 
    rounded corners, 
]

\begin{tabularx}{\textwidth}{X}
\textbf{Database Prompt:} ... \\
\textbf{Question:} Among the payments made by Mary Smith, how many of them are over 4.99?\\ 
\textbf{External Knowledge:} over 4.99 refers to amount > 4.99 \\
\textbf{Gold SQL:} SELECT count(payment.amount) FROM payment INNER JOIN customer ON payment.customer\_id = customer.customer\_id WHERE customer.first\_name = 'MARY' AND customer.last\_name = 'SMITH' AND payment.amount > 4.99 \\
\hline
\\ \textbf{Synthesized Chain-of-Thought Solution} \\

To answer the question about how many payments made by Mary Smith are greater than 4.99, we will need to join the \verb|`|payment\verb|`| table with the \verb|`|customer\verb|`| table. The query will count the number of payments that match Mary Smith's first and last name while ensuring the payment amount exceeds 4.99.
\\ \\
Here’s how the SQL query can be constructed:
\\
1. **Join the Tables**: We will perform INNER JOIN between the \verb|`|payment\verb|`| and \verb|`|customer\verb|`| tables on the \verb|`|customer\_id\verb|`| column. This links payments to the customers who made them.

2. **Filter Conditions**: We will add conditions to filter for only those rows where the customer's first name is 'MARY' and last name is 'SMITH'.

3. **Amount Condition**: Additionally, we will include a condition to check if the payment amount is greater than 4.99.

4. **Count the Results**: Finally, we will use the COUNT function to return the total number of matching records.
\\ \\
Here is the SQL query that fulfills the requirements:
\\ 
\verb|`|\verb|`|\verb|`|SQL \\
SELECT COUNT(payment.amount) \\ 
FROM payment \\ 
INNER JOIN customer ON payment.customer\_id = customer.customer\_id \\
WHERE customer.first\_name = 'MARY' \\
AND customer.last\_name = 'SMITH' \\
AND payment.amount > 4.99; \\
\verb|`|\verb|`|\verb|`|
\\ \\ 
**Explanation**: \\
- \verb|`|SELECT COUNT(payment.amount)\verb|`|: This part specifies that we want to count the number of payments that meet the conditions specified. \\
- \verb|`|FROM payment\verb|`|: Indicates that the main table we are querying is the \verb|`|payment\verb|`| table. \\
- \verb|`|INNER JOIN customer ON payment.customer\_id = customer.customer\_id\verb|`|: This joins the \verb|`|payment\verb|`| table with the \verb|`|customer\verb|`| table based on the \verb|`|customer\_id\verb|`| present in both tables, allowing us to access customer details related to each payment. \\
- \verb|`|WHERE customer.first\_name = 'MARY' AND customer.last\_name = 'SMITH'\verb|`|: These conditions filter the results to include only payments made by a customer named Mary Smith. \\
- \verb|`|AND payment.amount > 4.99\verb|`|: This additional condition specifies that we only want payments where the amount is greater than 4.99.
\\ \\
This query will count the payments made by Mary Smith that exceed the specified amount.

\end{tabularx}
\end{tcolorbox}
\caption{A qualitative example of \texttt{gpt-4o-mini-2024-07-18} generated Chain-of-Thought for an instance of Bird Train set. \textbf{SQL in the CoT can be different from original gold SQL in the dataset}. }
\label{tab:synthesisCoT}
\end{table*}

%% file: Tables/ExampleCoTSFT.tex
\begin{table*}[htbp]
\centering
\begin{tcolorbox}[
    colback=white, 
    colframe=black, 
    title=Response, 
    fonttitle=\bfseries, 
    width=\textwidth, 
    boxrule=1pt, 
    rounded corners, 
]

\begin{tabularx}{\textwidth}{X}
\textbf{Database Prompt:} Table district, columns = [ district.a11 ( integer | comment : average salary | values : 12541 , 8507 ) ,
district.a3 ( text | comment : region | values : Prague , \underline{central Bohemia} ) , district.district\_id ( integer | primary
key | comment : location of branch | values : 1 , 2 ) ... \\
\textbf{Question:} How many male customers who are living in North Bohemia have average salary greater than 8000? \\ 
\textbf{External Knowledge:} Male means that gender = 'M'; a3 refers to region; a11 pertains to average salary. \\
\hline
\\ 

To answer the question of how many male customers living in North Bohemia have an average salary greater than 8000, we need to join the \verb|`|district\verb|`| and \verb|`|client\verb|`| tables based on the \verb|`|district\_id\verb|`|. We will then filter the results based on the specified conditions: the region being 'North Bohemia', the gender being 'M', and the average salary being greater than 8000. \\ \\

Here's the SQL query that accomplishes this: \\ \\

\verb|`|\verb|`|\verb|`|SQL \\
SELECT COUNT(client.client\_id) \\
FROM district \\
INNER JOIN client ON district.district\_id = client.district\_id \\
WHERE district.a3 = 'North Bohemia' \\ 
AND client.gender = 'M' \\ 
AND district.a11 > 8000; \\ 
\verb|`|\verb|`|\verb|`| \\ \\ 

\#\#\# Explanation: \\
1. **SELECT COUNT(client.client\_id)**: This part of the query counts the number of male clients that meet the specified criteria. 
  
2. **FROM district**: We start our query from the \verb|`|district\verb|`| table since we need to filter based on the region and average salary.

3. **INNER JOIN client ON district.district\_id = client.district\_id**: We perform an inner join between the \verb|`|district\verb|`| and \verb|`|client\verb|`| tables using the \verb|`|district\_id\verb|`| as the key. This allows us to access the client information associated with each district.

4. **WHERE district.a3 = 'North Bohemia'**: This condition filters the results to only include districts in the 'North Bohemia' region.

5. **AND client.gender = 'M'**: This condition filters the results to only include male clients.

6. **AND district.a11 > 8000**: This condition filters the results to only include districts where the average salary is greater than 8000.

\\
By executing this query, we will obtain the count of male customers living in North Bohemia with an average salary greater than 8000.

\end{tabularx}
\end{tcolorbox}
\caption{The full response Syn CoT SFT Qwen2.5-7B-Instruct generated for a sample question from Bird Dev. \textbf{The response is wrong due to entity mismatch for the region 'north Bohemia'}.}
\label{tab:cotResponseSFT}
\end{table*}

%% file: Tables/ExampleCoTDPO.tex
\begin{table*}[htbp]
\centering
\begin{tcolorbox}[
    colback=white, 
    colframe=black, 
    title=Response, 
    fonttitle=\bfseries, 
    width=\textwidth, 
    boxrule=1pt, 
    rounded corners, 
]

\begin{tabularx}{\textwidth}{X}
\textbf{Database Prompt:} Table district, columns = [ district.a11 ( integer | comment : average salary | values : 12541 , 8507 ) ,
district.a3 ( text | comment : region | values : Prague , \underline{central Bohemia} ) , district.district\_id ( integer | primary
key | comment : location of branch | values : 1 , 2 ) ... \\
\textbf{Question:} How many male customers who are living in North Bohemia have average salary greater than 8000? \\ 
\textbf{External Knowledge:} Male means that gender = 'M'; a3 refers to region; a11 pertains to average salary. \\
\hline
\\ 

To answer the question of how many male customers living in North Bohemia have an average salary greater than 8000, we need to join the \verb|`|client\verb|`| table with the \verb|`|district\verb|`| table based on the \verb|`|district\_id\verb|`|. We will then apply the necessary filters for the region (\verb|`|a3\verb|`|), gender (\verb|`|gender\verb|`|), and average salary (\verb|`|a11\verb|`|).
\\ \\ 
Here's the breakdown of the SQL query:
\\ \\ 
1. **SELECT count(client.client\_id)**: We want to count the number of male clients, so we use the `count` function on `client.client\_id`. \\

2. **FROM district INNER JOIN client ON district.district\_id = client.district\_id**: We perform an inner join between the \verb|`|district\verb|`| and \verb|`|client\verb|`| tables on the \verb|`|district\_id\verb|`| column, which is common to both tables. This allows us to access the district information for each client. \\

3. **WHERE district.a3 = 'north Bohemia'**: We filter the results to include only those clients who live in the North Bohemia region. \\

4. **AND client.gender = 'M'**: We further filter to include only male clients. \\

5. **AND district.a11 > 8000**: Finally, we add a condition to ensure that we only count clients whose average salary is greater than 8000. 
\\ \\
Putting this all together, the SQL query is as follows:
\\ \\ 
\verb|`|\verb|`|\verb|`|SQL \\
SELECT count(client.client\_id) \\ 
FROM district \\
INNER JOIN client ON district.district\_id = client.district\_id \\
WHERE district.a3 = 'north Bohemia' \\
AND client.gender = 'M' \\
AND district.a11 > 8000; \\
\verb|`|\verb|`|\verb|`| 
\\ \\
This query will return the count of male clients residing in North Bohemia with an average salary exceeding 8000.

\end{tabularx}
\end{tcolorbox}
\caption{The full response Syn CoT DPO Qwen2.5-7B-Instruct generated for a sample question from Bird Dev. \textbf{The model is able to infer entity format of 'north Bohemia' from value examples given in database prompt}.}
\label{tab:cotResponseDPO}
\end{table*}

%% file: Tables/MainExp_Spider.tex
\definecolor{darkgreen}{RGB}{0,150,0}
\begin{table*}[t!]
    \centering
\begin{adjustbox}{max width=\textwidth}
    \begin{tabular}{c  c | c c | c c | c c | c}
        \toprule
         & \multirow{3}{*}{\textbf{Model}} & \multicolumn{6}{c}{\textbf{Spider Dev}} & \\ \cline{3-9}
         & & \multicolumn{2}{c}{Greedy} & \multicolumn{2}{c}{Pass@1} & \multicolumn{2}{c |}{Maj@16} & \multirow{2}{*}{$\Delta$} \\ \cline{3-8}
         & & SFT & DPO & SFT & DPO & SFT & DPO & \\ \midrule
        \multicolumn{9}{c}{\textbf{Execution Accuracy (EX)}} \\ \midrule
        
        & Deepseek-coder-6.7b-instruct & 81.6 & 81.0 (\textcolor{darkgreen}{-0.6}) & 80.0  & 80.4 (\textcolor{red}{+0.4}) & \text{82.5} & 81.9 (\textcolor{darkgreen}{-0.6}) & -\\ 
        & Qwen2.5-7B-Instruct & 79.8 & 79.7 (\textcolor{darkgreen}{-0.1}) & 78.4  & 78.6 (\textcolor{red}{+0.2}) & \text{81.8} & 81.6 (\textcolor{darkgreen}{-0.2}) & -\\ 
        \multirow{-3}{*}{\textbf{Vanilla}} & CodeS-7b & 79.3 & 79.4 (\textcolor{red}{+0.1}) & 78.0  & 78.2 (\textcolor{red}{+0.2}) & \text{80.9} & 80.4 (\textcolor{darkgreen}{-0.5}) & - \\
        
        \rowcolor{cyan!20}
        & Deepseek-coder-6.7b-instruct & 80.0 & 82.1 (\textcolor{red}{+2.1}) & 79.4  & 81.3 (\textcolor{red}{+1.9}) & 82.8 & \text{83.8} (\textcolor{red}{+1.0}) & 82.5 $\rightarrow$ 83.8 (\textbf{\textcolor{red}{+1.3}}) \\ 
        \rowcolor{cyan!20}
        & Qwen2.5-7B-Instruct & 79.1 & 80.5 (\textcolor{red}{+1.4}) & 76.8 & 78.9 (\textcolor{red}{+2.1}) & 80.4 & \text{82.6} (\textcolor{red}{+2.2}) & 81.8 $\rightarrow$ 82.6 (\textbf{\textcolor{red}{+0.8}})\\ 
        \rowcolor{cyan!20}
        \multirow{-3}{*}{\textbf{Syn CoT}} & CodeS-7b & 76.7 & 80.0 (\textcolor{red}{+3.3}) & 75.5  & 78.8 (\textcolor{red}{+3.3}) &  78.3 & \text{82.3} (\textcolor{red}{+4.0}) & 80.9 $\rightarrow$ 82.3 (\textbf{\textcolor{red}{+1.4}}) \\
        
        \midrule
        \multicolumn{9}{c}{\textbf{Test Suite (TS)}} \\ \midrule
        
        & Deepseek-coder-6.7b-instruct & 80.5 & 80.7 (\textcolor{red}{+0.2}) & 79.1 & 79.6 (\textcolor{red}{+0.5}) & \text{81.1} & 80.7 (\textcolor{darkgreen}{-0.4}) & - \\
        & Qwen2.5-7B-Instruct & 78.5 & 78.9 (\textcolor{red}{+0.4})& 77.2 & 77.1 (\textcolor{darkgreen}{+0.1})& \text{79.5} & 79.3 (\textcolor{darkgreen}{-0.2}) & - \\
        \multirow{-3}{*}{\textbf{Vanilla}} & CodeS-7b & 76.9 & 77.1 (\textcolor{red}{+0.2}) & 75.4 & 75.7 (\textcolor{red}{+0.3}) & \text{78.1} & 77.7 (\textcolor{darkgreen}{-0.4}) & - \\
        
        \rowcolor{cyan!20}
        & Deepseek-coder-6.7b-instruct & 78.3 & 80.3 (\textcolor{red}{+2.0}) & 76.9 & 78.9 (\textcolor{red}{+2.0}) & 79.7 & \text{81.6} (\textcolor{red}{+1.9}) & 81.1 $\rightarrow$ 81.6 (\textbf{\textcolor{red}{+0.5}}) \\
        \rowcolor{cyan!20}
        & Qwen2.5-7B-Instruct & 77.5 & 77.9 (\textcolor{red}{+0.4})& 75.0 & 76.6 (\textcolor{red}{+1.6})& 78.6 & \text{80.2} (\textcolor{red}{+1.6}) & 79.5 $\rightarrow$ 80.2 (\textbf{\textcolor{red}{+0.7}})\\
        \rowcolor{cyan!20}
        \multirow{-3}{*}{\textbf{Syn CoT}} & CodeS-7b & 74.8 & 77.0 (\textcolor{red}{+2.2}) & 73.1 & 75.3 (\textcolor{red}{+2.2}) & 76.1 & \text{77.8} (\textcolor{red}{+1.7}) & 78.1 $\rightarrow$ 77.8 (\textbf{\textcolor{darkgreen}{-0.3}}) \\
        \bottomrule
    \end{tabular}
\end{adjustbox}
    \caption{Model performance on the Spider development set. \textbf{Vanilla}: SFT and DPO on the original Spider training set; \textbf{Syn CoT}: SFT and DPO on the CoT-enhanced training set; \textbf{$\Delta$}: The performance difference in EX/TS between ``Syn CoT + DPO'' and ``Vanilla + SFT'' when using the same base model.\textbf{We make preference dataset with EX since TS in not available to train set.}}
    \label{tab:spiderDev}
\end{table*}


%% file: Tables/MainExp_SpiderOther.tex
\definecolor{darkgreen}{RGB}{0,150,0}
\begin{table*}[t!]
    \centering
\begin{adjustbox}{max width=\textwidth}
    \begin{tabular}{c  c | c c | c c | c c | c}
        \toprule
         & \multirow{3}{*}{\textbf{Model}} & \multicolumn{6}{c}{\textbf{Spider Variants Dev (EX)}} & \\ \cline{3-9}
         & & \multicolumn{2}{c}{Greedy} & \multicolumn{2}{c}{Pass@1} & \multicolumn{2}{c |}{Maj@16} & \multirow{2}{*}{$\Delta$EX} \\ \cline{3-8}
         & & SFT & DPO & SFT & DPO & SFT & DPO & \\ \midrule
        \multicolumn{9}{c}{\textbf{Spider-Syn}} \\ \midrule
        
        & Deepseek-coder-6.7b-instruct & 73.8 & 73.5 (\textcolor{darkgreen}{-0.3}) & 72.5 & 72.4 (\textcolor{darkgreen}{-0.1}) & 75.2 & \text{75.5} (\textcolor{red}{+0.3}) & -\\ 
        & Qwen2.5-7B-Instruct & 71.5 & 72.0 (\textcolor{red}{+0.5}) & 69.6  & 69.8 (\textcolor{red}{+0.2}) & \text{73.7} & 73.6 (\textcolor{darkgreen}{-0.1}) & -\\ 
        \multirow{-3}{*}{\textbf{Vanilla}} & CodeS-7b & 69.1 & 69.4 (\textcolor{red}{+0.3}) & 67.1  & 67.6 (\textcolor{red}{+0.5}) & 71.1 & \text{71.2} (\textcolor{red}{+0.1}) & - \\
 
        \rowcolor{cyan!20}
        & Deepseek-coder-6.7b-instruct & 70.7 & 71.7 (\textcolor{red}{+1.0}) & 70.1  & 71.3 (\textcolor{red}{+1.2}) & 76.7 & \text{76.8} (\textcolor{red}{+0.1}) & 75.2 $\rightarrow$ 76.8 (\textbf{\textcolor{red}{+1.6}}) \\ 
        \rowcolor{cyan!20}
        & Qwen2.5-7B-Instruct & 69.7 & 71.2 (\textcolor{red}{+1.5}) & 67.9 & 70.5 (\textcolor{red}{+2.6}) & 74.1 & \text{76.2} (\textcolor{red}{+2.1}) & 73.7 $\rightarrow$ 76.2 (\textbf{\textcolor{red}{+2.5}})\\ 
        \rowcolor{cyan!20}
        \multirow{-3}{*}{\textbf{Syn CoT}} & CodeS-7b & 64.4 & 69.1 (\textcolor{red}{+4.7}) & 63.6  & 67.5 (\textcolor{red}{+3.9}) & 69.8 & \text{71.4} (\textcolor{red}{+1.6}) & 71.1 $\rightarrow$ 71.4 (\textbf{\textcolor{red}{+0.3}}) \\
        
        \midrule
        \multicolumn{9}{c}{\textbf{Spider-Realistic}} \\ \midrule
        
        & Deepseek-coder-6.7b-instruct & 77.2 & 77.2 \phantom{(\textcolor{red}{+0.2})} & 77.0 & 76.5 (\textcolor{darkgreen}{-0.5}) & \text{78.9} & 78.3 (\textcolor{darkgreen}{-0.6}) & - \\
        & Qwen2.5-7B-Instruct & 75.4 & 75.4 \phantom{(\textcolor{red}{+0.4})}& 73.7 & 74.2 (\textcolor{red}{+0.5})& 76.6 & \text{77.4} (\textcolor{red}{+0.8}) & - \\
        \multirow{-3}{*}{\textbf{Vanilla}} & CodeS-7b & 73.8 & 74.0 (\textcolor{red}{+0.2}) & 73.0 & 72.9 (\textcolor{darkgreen}{-0.1}) & \text{76.4} & 75.8 (\textcolor{darkgreen}{-0.6}) & - \\
        
        \rowcolor{cyan!20}
        & Deepseek-coder-6.7b-instruct & 79.7 & 80.7 (\textcolor{red}{+1.0}) & 77.4 & 78.7 (\textcolor{red}{+1.3}) & 80.9 & \text{82.7} (\textcolor{red}{+1.8}) & 78.9 $\rightarrow$ 82.7 (\textbf{\textcolor{red}{+3.8}}) \\
        \rowcolor{cyan!20}
        & Qwen2.5-7B-Instruct & 76.0 & 77.4 (\textcolor{red}{+1.4})& 73.9 & 75.1 (\textcolor{red}{+1.2})& 78.3 & \text{79.1} (\textcolor{red}{+0.8}) & 76.6 $\rightarrow$ 79.1 (\textbf{\textcolor{red}{+2.5}})\\
        \rowcolor{cyan!20}
        \multirow{-3}{*}{\textbf{Syn CoT}} & CodeS-7b & 73.8 & 76.4 (\textcolor{red}{+2.6}) & 71.5 & 73.3 (\textcolor{red}{+1.8}) & 76.6 & \text{78.5} (\textcolor{red}{+1.9}) & 76.4 $\rightarrow$ 78.5 (\textbf{\textcolor{red}{+2.1}}) \\

        \midrule
        \multicolumn{9}{c}{\textbf{Spider-DK}} \\ \midrule

        & Deepseek-coder-6.7b-instruct & 69.0 & 69.8 (\textcolor{darkgreen}{-0.2}) & 68.2 & 68.5 (\textcolor{red}{+0.3}) & \text{70.3} & 70.1 (\textcolor{darkgreen}{-0.2}) & - \\
        & Qwen2.5-7B-Instruct & 70.5 & 70.3 (\textcolor{darkgreen}{-0.2})& 69.0 & 69.4 (\textcolor{red}{+0.4})& \text{73.8} & 73.6 (\textcolor{darkgreen}{-0.2}) & - \\
        \multirow{-3}{*}{\textbf{Vanilla}} & CodeS-7b & 67.5 & 67.5 \phantom{(\textcolor{red}{+0.2})} & 66.3 & 66.6 (\textcolor{red}{+0.3}) & \text{69.2} & 68.6 (\textcolor{darkgreen}{-0.6}) & - \\

        \rowcolor{cyan!20}
        & Deepseek-coder-6.7b-instruct & 69.7 & 72.1 (\textcolor{red}{+2.4}) & 68.4 & 71.2 (\textcolor{red}{+2.8}) & 72.9 & \text{75.1} (\textcolor{red}{+2.2}) & 70.3 $\rightarrow$ 75.1 (\textbf{\textcolor{red}{+4.8}}) \\
        \rowcolor{cyan!20}
        & Qwen2.5-7B-Instruct & 67.5 & 69.0 (\textcolor{red}{+1.5})& 65.1 & 67.2 (\textcolor{red}{+2.1})& 70.7 & \text{72.9} (\textcolor{red}{+2.2}) & 73.8 $\rightarrow$ 72.9 (\textbf{\textcolor{darkgreen}{-0.9}})\\
        \rowcolor{cyan!20}
        \multirow{-3}{*}{\textbf{Syn CoT}} & CodeS-7b & 62.6 & 67.7 (\textcolor{red}{+5.1}) & 61.9 & 67.0 (\textcolor{red}{+5.1}) & 67.5 & \text{72.1} (\textcolor{red}{+4.6}) & 69.2 $\rightarrow$ 72.1 (\textbf{\textcolor{red}{+2.9}}) \\
        
        \bottomrule
    \end{tabular}
\end{adjustbox}
    \caption{Model performance on Spider's variants (Spider-Syn, Spider-Realistiv, Spider-DK). \textbf{Vanilla}: SFT and DPO on the original Spider training set; \textbf{Syn CoT}: SFT and DPO on the CoT-enhanced training set; \textbf{$\Delta$EX}: The performance difference in EX between ``Syn CoT + DPO'' and ``Vanilla + SFT'' when using the same base model. \textbf{In this setting, we directly assess best checkpoint on Spider Dev.}}
    \label{tab:spiderVariants}
\end{table*}

%% file: Tables/MainExp_DrDB.tex
\definecolor{darkgreen}{RGB}{0,150,0}
\begin{table*}[t!]
    \centering
\begin{adjustbox}{max width=\textwidth}
\begin{tabular}{c | l | c c | c c | c c }
        \toprule
         \multirow{3}{*}{\textbf{Type}} & \multirow{3}{*}{\textbf{Metrics}} & \multicolumn{6}{c}{\textbf{Syn CoT Models}} \\ \cline{3-8}
         & & \multicolumn{2}{c}{DSC (6.7B)} & \multicolumn{2}{c}{Qwen (7B)} & \multicolumn{2}{c}{CodeS (7B)} \\ \cline{3-8}
         & & SFT & DPO & SFT & DPO & SFT & DPO \\ \midrule
        \multicolumn{8}{c}{\textbf{DB Perturbations}} \\ \midrule

        \multirow{3}{*}{schema-synonym} & Greedy & 66.6 & 68.8 (\textcolor{red}{+2.2}) & 63.1 & 65.1 (\textcolor{red}{+2.0}) & 57.5 & 62.2 (\textcolor{red}{+4.7}) \\
        & Pass@1 & 63.8 & 66.7 (\textcolor{red}{+2.9}) & 60.7 & 63.3 (\textcolor{red}{+2.6}) & 55.0 & 60.8 (\textcolor{red}{+5.8}) \\
        & Maj@K & 69.7 & 71.7 (\textcolor{red}{+2.0}) & 68.1 & 69.3 (\textcolor{red}{+1.2}) & 63.7 & 66.2 (\textcolor{red}{+2.5}) \\
         \midrule

        \multirow{3}{*}{schema-abbreviation} & Greedy & 74.8 & 77.9 (\textcolor{red}{+3.1}) & 70.7 & 73.3 (\textcolor{red}{+2.6}) & 69.4 & 74.9 (\textcolor{red}{+5.5}) \\
\bottomrule
    \end{tabular}
\end{adjustbox}
    \caption{Syn CoT model performance on DB perturbations of Dr.Spider dataset. Names of base models are abbreviated. \textbf{DSC (6.7B)}: Deepseek-coder-6.7b-instruct; \textbf{Qwen (7B)}: Qwen2.5-7B-Instruct; \textbf{CodeS (7B)}: CodeS-7b. \textbf{In this setting, we directly assess best checkpoint on Spider Dev.}}
    \label{tab:drSpiderDBCoT}
\end{table*}

\definecolor{darkgreen}{RGB}{0,150,0}
\begin{table*}[t!]
    \centering
\begin{adjustbox}{max width=\textwidth}
    \begin{tabular}{c | l | c c | c c | c c }
        \toprule
         \multirow{3}{*}{\textbf{Type}} & \multirow{3}{*}{\textbf{Metrics}} & \multicolumn{6}{c}{\textbf{Vanilla Models}} \\ \cline{3-8}
         & & \multicolumn{2}{c}{DSC (6.7B)} & \multicolumn{2}{c}{Qwen (7B)} & \multicolumn{2}{c}{CodeS (7B)} \\ \cline{3-8}
         & & SFT & DPO & SFT & DPO & SFT & DPO \\ \midrule
        \multicolumn{8}{c}{\textbf{DB Perturbations}} \\ \midrule

        \multirow{3}{*}{schema-synonym} & Greedy & 66.3 & 66.4 (\textcolor{red}{+0.1}) & 63.3 & 63.5 (\textcolor{red}{+0.2}) & 62.0 & 62.2 (\textcolor{red}{+0.2}) \\
        & Pass@1 & 65.6 & 65.6 (\textcolor{red}{+0.0}) & 61.1 & 61.2 (\textcolor{red}{+0.1}) & 59.7 & 60.5 (\textcolor{red}{+0.8}) \\
        & Maj@K & 69.0 & 68.6 (\textcolor{darkgreen}{-0.4}) & 67.9 & 67.8 (\textcolor{darkgreen}{-0.1}) & 65.7 & 65.9 (\textcolor{red}{+0.2}) \\ \midrule

        \multirow{3}{*}{schema-abbreviation} & Greedy & 77.0 & 77.3 (\textcolor{red}{+0.3}) & 75.3 & 75.7 (\textcolor{red}{+0.4}) & 71.3 & 71.3 \phantom{(+0.3)} \\
\bottomrule
    \end{tabular}
\end{adjustbox}
    \caption{Vanilla model performance on DB perturbations of Dr.Spider dataset. Names of base models are abbreviated. \textbf{DSC (6.7B)}: Deepseek-coder-6.7b-instruct; \textbf{Qwen (7B)}: Qwen2.5-7B-Instruct; \textbf{CodeS (7B)}: CodeS-7b. \textbf{In this setting, we directly assess best checkpoint on Spider Dev.}} 
    \label{tab:drSpiderDBVanilla}
\end{table*}

%% file: Tables/MainExp_DrNLQ.tex
\definecolor{darkgreen}{RGB}{0,150,0}
\begin{table*}[t!]
    \centering
\begin{adjustbox}{max width=\textwidth}
    \begin{tabular}{c | l | c c | c c | c c }
        \toprule
         \multirow{3}{*}{\textbf{Type}} & \multirow{3}{*}{\textbf{Metrics}} & \multicolumn{6}{c}{\textbf{Syn CoT Models}} \\ \cline{3-8}
         & & \multicolumn{2}{c}{DSC (6.7B)} & \multicolumn{2}{c}{Qwen (7B)} & \multicolumn{2}{c}{CodeS (7B)} \\ \cline{3-8}
         & & SFT & DPO & SFT & DPO & SFT & DPO \\ \midrule
        \multicolumn{8}{c}{\textbf{NLQ Perturbations}} \\ \midrule

        \multirow{3}{*}{keyword-synonym} & Greedy & 67.4 & 71.2 (\textcolor{red}{+3.8}) & 66.7 & 68.9 (\textcolor{red}{+2.2}) & 62.5 & 71.9 (\textcolor{red}{+9.4}) \\
        & Pass@1 & 66.9 & 70.4 (\textcolor{red}{+3.5}) & 65.6 & 68.5 (\textcolor{red}{+2.9}) & 60.6 & 68.8 (\textcolor{red}{+8.2}) \\
        & Maj@K & 69.9 & 73.5 (\textcolor{red}{+3.6}) & 69.7 & 71.5 (\textcolor{red}{+1.8}) & 66.7 & 72.8 (\textcolor{red}{+6.1}) \\ \midrule

        \multirow{3}{*}{keyword-carrier} & Greedy & 76.4 & 80.7 (\textcolor{red}{+4.3}) & 77.2 & 78.9 (\textcolor{red}{+1.7}) & 77.4 & 84.5 (\textcolor{red}{+7.1}) \\
\bottomrule
    \end{tabular}
\end{adjustbox}
    \caption{Syn CoT model performance on NLQ perturbations of Dr.Spider dataset. Names of base models are abbreviated. \textbf{DSC (6.7B)}: Deepseek-coder-6.7b-instruct; \textbf{Qwen (7B)}: Qwen2.5-7B-Instruct; \textbf{CodeS (7B)}: CodeS-7b. \textbf{In this setting, we directly assess best checkpoint on Spider Dev.}} 
    \label{tab:drSpiderNLQCoT}
\end{table*}

\definecolor{darkgreen}{RGB}{0,150,0}
\begin{table*}[t!]
    \centering
\begin{adjustbox}{max width=\textwidth}
    \begin{tabular}{c | l | c c | c c | c c }
        \toprule
         \multirow{3}{*}{\textbf{Type}} & \multirow{3}{*}{\textbf{Metrics}} & \multicolumn{6}{c}{\textbf{Vanilla Models}} \\ \cline{3-8}
         & & \multicolumn{2}{c}{DSC (6.7B)} & \multicolumn{2}{c}{Qwen (7B)} & \multicolumn{2}{c}{CodeS (7B)} \\ \cline{3-8}
         & & SFT & DPO & SFT & DPO & SFT & DPO \\ \midrule
        \multicolumn{8}{c}{\textbf{NLQ Perturbations}} \\ \midrule

        \multirow{3}{*}{keyword-synonym} & Greedy & 69.6 & 69.8 (\textcolor{red}{+0.2}) & 68.2 & 68.3 (\textcolor{red}{+0.1}) & 66.8 & 67.7 (\textcolor{red}{+0.9}) \\
        & Pass@1 & 68.7 & 68.8 (\textcolor{red}{+0.1}) & 67.1 & 67.0 (\textcolor{darkgreen}{-0.1}) & 65.3 & 66.0 (\textcolor{red}{+0.7}) \\
        & Maj@K & 70.9 & 70.4 (\textcolor{darkgreen}{-0.5}) & 69.4 & 69.5 (\textcolor{red}{+0.1}) & 68.9 & 68.5 (\textcolor{darkgreen}{-0.4}) \\
        \midrule

        \multirow{3}{*}{keyword-carrier} & Greedy & 80.7 & 81.5 (\textcolor{red}{+0.8}) & 80.2 & 80.5 (\textcolor{red}{+0.3}) & 81.7 & 81.2 (\textcolor{darkgreen}{-0.5}) \\
\bottomrule
    \end{tabular}
\end{adjustbox}
    \caption{Vanilla model performance on NLQ perturbations of Dr.Spider dataset. Names of base models are abbreviated. \textbf{DSC (6.7B)}: Deepseek-coder-6.7b-instruct; \textbf{Qwen (7B)}: Qwen2.5-7B-Instruct; \textbf{CodeS (7B)}: CodeS-7b. \textbf{In this setting, we directly assess best checkpoint on Spider Dev.}} 
    \label{tab:drSpiderNLQVanilla}
\end{table*}

%% file: Tables/MainExp_DrSQL.tex
\definecolor{darkgreen}{RGB}{0,150,0}
\begin{table*}[t!]
    \centering
\begin{adjustbox}{max width=\textwidth}
    \begin{tabular}{c | l | c c | c c | c c }
        \toprule
         \multirow{3}{*}{\textbf{Type}} & \multirow{3}{*}{\textbf{Metrics}} & \multicolumn{6}{c}{\textbf{Syn CoT Models}} \\ \cline{3-8}
         & & \multicolumn{2}{c}{DSC (6.7B)} & \multicolumn{2}{c}{Qwen (7B)} & \multicolumn{2}{c}{CodeS (7B)} \\ \cline{3-8}
         & & SFT & DPO & SFT & DPO & SFT & DPO \\ \midrule
        \multicolumn{8}{c}{\textbf{SQL Perturbations}} \\ \midrule

        \multirow{3}{*}{comparison} & Greedy & 68.0 & 74.2 (\textcolor{red}{+6.2}) & 57.9 & 64.6 (\textcolor{red}{+6.7}) & 68.0 & 75.3 (\textcolor{red}{+7.3}) \\
& Pass@1 & 68.0 & 73.5 (\textcolor{red}{+5.5}) & 60.8 & 66.5 (\textcolor{red}{+5.7}) & 66.0 & 74.2 (\textcolor{red}{+8.2}) \\
& Maj@K & 75.3 & 77.5 (\textcolor{red}{+2.2}) & 62.9 & 70.2 (\textcolor{red}{+7.3}) & 76.4 & 78.7 (\textcolor{red}{+2.3}) \\
 \midrule

        \multirow{3}{*}{sort-order} & Greedy & 71.4 & 78.6 (\textcolor{red}{+7.2}) & 69.3 & 73.4 (\textcolor{red}{+4.1}) & 63.0 & 74.5 (\textcolor{red}{+11.5}) \\
\bottomrule
    \end{tabular}
\end{adjustbox}
    \caption{Syn CoT model performance on SQL perturbations of Dr.Spider dataset. Names of base models are abbreviated. \textbf{DSC (6.7B)}: Deepseek-coder-6.7b-instruct; \textbf{Qwen (7B)}: Qwen2.5-7B-Instruct; \textbf{CodeS (7B)}: CodeS-7b. \textbf{In this setting, we directly assess best checkpoint on Spider Dev.}} 
    \label{tab:drSpiderSQLCoT}
\end{table*}

\definecolor{darkgreen}{RGB}{0,150,0}
\begin{table*}[t!]
    \centering
\begin{adjustbox}{max width=\textwidth}
    \begin{tabular}{c | l | c c | c c | c c }
        \toprule
         \multirow{3}{*}{\textbf{Type}} & \multirow{3}{*}{\textbf{Metrics}} & \multicolumn{6}{c}{\textbf{Vanilla Models}} \\ \cline{3-8}
         & & \multicolumn{2}{c}{DSC (6.7B)} & \multicolumn{2}{c}{Qwen (7B)} & \multicolumn{2}{c}{CodeS (7B)} \\ \cline{3-8}
         & & SFT & DPO & SFT & DPO & SFT & DPO \\ \midrule
        \multicolumn{8}{c}{\textbf{SQL Perturbations}} \\ \midrule

        \multirow{3}{*}{comparison} & Greedy & 70.2 & 69.7 (\textcolor{darkgreen}{-0.5}) & 69.1 & 69.1 \phantom{(+0.3)} & 71.9 & 72.5 (\textcolor{red}{+0.6}) \\
& Pass@1 & 69.7 & 69.6 (\textcolor{darkgreen}{-0.1}) & 67.7 & 68.3 (\textcolor{red}{+0.6}) & 68.9 & 69.5 (\textcolor{red}{+0.6}) \\
& Maj@K & 73.6 & 73.6 \phantom{(+0.3)} & 71.9 & 71.9 \phantom{(+0.3)} & 72.5 & 73.0 (\textcolor{red}{+0.5}) \\
 \midrule

        \multirow{3}{*}{sort-order} & Greedy & 79.2 & 79.2 \phantom{(+0.3)} & 75.5 & 76.0 (\textcolor{red}{+0.5}) & 78.6 & 77.6 (\textcolor{darkgreen}{-1.0}) \\
\bottomrule
    \end{tabular}
\end{adjustbox}
    \caption{Vanilla model performance on SQL perturbations of Dr.Spider dataset. Names of base models are abbreviated. \textbf{DSC (6.7B)}: Deepseek-coder-6.7b-instruct; \textbf{Qwen (7B)}: Qwen2.5-7B-Instruct; \textbf{CodeS (7B)}: CodeS-7b. \textbf{In this setting, we directly assess best checkpoint on Spider Dev.}}
    \label{tab:drSpiderSQLVanilla}
\end{table*}

%% file: Tables/ErrorCategories.tex
\definecolor{darkgreen}{RGB}{0,150,0}
\begin{table*}[t]
\centering
\begin{adjustbox}{max width=\textwidth}
\begin{tabular}{c p{5.5cm} l p{6cm} }
\toprule
\textbf{Category} & \textbf{Description} & \textbf{Type} & \textbf{Description}\\
\midrule
\multirow{1}{*}{External Knowledge} 
    & \textbf{Neglect of external knowledge} directly causes the failure.
    & [A1] EK 
    & \textbf{Neglect of external knowledge} directly causes the failure.
    \\
\midrule
\multirow{6}{*}{Schema Linking} 
    & \multirow{6}{=}{The SQL fails to match the question with its \textbf{concerning table and columns.}}
    & [B1] Table 
    & The SQL selects \textbf{irrelevant tables, or misses necessary tables}. 
    \\
    & & [B2] JOIN & Selects right table, but \textbf{joins on the wrong key}.
    \\
    & & [B3] Column & The SQL fails to understand \textbf{critical columns}. 
    \\
    & & [B4] Hallucination & \textbf{Make up table or column} that does not exist in database.
    \\
    & & [B5] Condition & \textbf{Misunderstands/misses out} the required condition.
    \\
    & & [B6] NULL/DISTINCT & Possible \textbf{replication or null data} affect query result.
    \\
\midrule
\multirow{2}{*}{Value Retrieval}
    & \multirow{2}{=}{Fails to match the condition with its \textbf{storage format} in  database.}
    & [C1] String/Number 
    & \textbf{Mismatch of entity format}.
    \\
    & & [C2] Date & \textbf{Manipulation of Date} is wrong.
    \\
\midrule
\multirow{3}{*}{Operation}
    & \multirow{3}{=}{\textbf{Misunderstands required operation} in the question.}
    & [D1] Mathematical Formula 
    & The \textbf{formula} is not correspond to the original question.
    \\
    & & [D2] Aggregation & Wrong use of \textbf{aggregation function or GROUP BY keyword}.
    \\
    & & [D3] Complex Operation & \textbf{Unsusal complex operation (CASE, RANK(), IIF(), ...)} is required in query.
    \\
\midrule
\multirow{4}{*}{Information}
    & \multirow{4}{=}{The SQL generally satisfies requirement of the question, but fails to \textbf{organize information} in right way.}
    & [E1] Redundant/Incomplete 
    & Extra \textbf{filter/ranking conditions returned} following queried information.
    \\
    & & [E2] Column Sequence & Returned columns are listed in \textbf{wrong order}.
    \\
    & & [E3] ORDER BY/LIMIT & Does not follow instruction to \textbf{limit} records according to certain criteria.
    \\
    & & [E4] Format & The \textbf{format} of returned records is not as required.
    \\
\midrule
Syntax Error 
    & \textbf{Inexecutatble SQL}.
    & [F1] Syntax 
    & \textbf{Inexecutable SQL}.
    \\ 
\bottomrule
\end{tabular}
\end{adjustbox}
\caption{Categories and Types we used in error analysis and their corresponding descriptions.}
\label{table:ErrorCategoryDescription}
\end{table*}

%% file: Tables/ErrorSamples.tex
\definecolor{darkgreen}{RGB}{0,150,0}
\newcommand{\bt}[1]{\textbf{\texttt{#1}}}
\newcommand{\gh}[1]{\textbf{\textcolor{darkgreen}{#1}}}
\newcommand{\rh}[1]{\textbf{\textcolor{red}{#1}}}

\begin{table*}[t]
\centering
\begin{adjustbox}{max width=\textwidth}
\begin{tabular}{p{4.75cm} p{5.5cm} p{5.5cm} p{2.5cm}}
\toprule
\textbf{Sample Question with EK} & \textbf{Gold SQL} & \textbf{Predicted SQL} & \textbf{Error Analysis}\\
\midrule
\multicolumn{4}{c}{[A1] EK} \\
\midrule
 Identify the atoms belong to the molecule with \rh{ID between TR010 to TR050} that contain the element carbon. EK: carbon refers to element = 'c'; \rh{between TR010 to TR050 refers to substr(molecule\_id, 3, 3)>=10 \bt{AND} substr(molecule\_id, 3, 3) <= 50}
 & \raggedright \bt{SELECT} T.atom\_id \bt{FROM} atom \bt{AS} T \bt{WHERE} T.molecule\_id \rh{\bt{BETWEEN} 'TR010' \bt{AND} 'TR050' \bt{AND} T.element = 'c'}
 & \raggedright \bt{SELECT} atom\_id \bt{FROM} atom \bt{WHERE} \gh{molecule\_id \bt{LIKE} 'TR\%' \bt{AND} element = 'c'}
 & The SQL selects all molecule with a 'TR' prefix instead. \\
\bottomrule
\end{tabular}
\end{adjustbox}
\caption{Sample errors and analysis for types in the External Knowledge category.}
\label{table:ErrorSampleEK}
\end{table*}

\begin{table*}[t]
\centering
\begin{adjustbox}{max width=\textwidth}
\begin{tabular}{p{4.75cm} p{5.5cm} p{5.5cm} p{2.5cm}}
\toprule
\textbf{Sample Question with EK} & \textbf{Gold SQL} & \textbf{Predicted SQL} & \textbf{Error Analysis}\\
\midrule
\multicolumn{4}{c}{[B1] Table} \\
\midrule
How many accounts who have region in Prague are eligible for \rh{loans}? EK: A3 contains the data of region & \raggedright
\bt{SELECT}  \bt{COUNT(}T1.account\_id\bt{)} \bt{FROM} account \bt{AS} T1 \rh{\bt{INNER} \bt{JOIN} loan} \bt{AS} T2 \bt{ON} T1.account\_id = T2.account\_id \bt{INNER} \bt{JOIN} district \bt{AS} T3 \bt{ON} T1.district\_id = T3.district\_id \bt{WHERE} T3.A3 = 'Prague' & \raggedright \bt{SELECT} \bt{COUNT(}account.account\_id\bt{)} \bt{FROM} account \bt{INNER} \bt{JOIN} district \bt{ON} account.district\_id = district.district\_id \bt{WHERE} district.a3 = 'Prague' & The loan table is need to check for eligibility. \\
\midrule
\multicolumn{4}{c}{[B2] JOIN} \\
\midrule
State the \rh{district} and region for \rh{loan} ID '4990'. EK: A2, A3 contains the information about district and region respectively. & \raggedright \bt{SELECT} T2.A2, T2.A3 \bt{FROM} account \bt{AS} T1 \bt{INNER} \bt{JOIN} district \bt{AS} T2 \bt{ON} \rh{T1.district\_id = T2.district\_id} \bt{INNER JOIN} loan \bt{AS} T3 \bt{ON} \rh{T1.account\_id = T3.account\_id} \bt{WHERE} T3.loan\_id = 4990 & \raggedright \bt{SELECT} district.a2, district.a3 \bt{FROM} district \bt{INNER JOIN} loan \bt{ON} \gh{district.district\_id = loan.account\_id} \bt{WHERE} loan.loan\_id = 4990; & District table and loan table can't join together without account table. \\
\midrule
\multicolumn{4}{c}{[B3] Column} \\
\midrule
In which \rh{mailing street address} can you find the school that has the lowest average score in reading? Also give the school's name. & \raggedright \bt{SELECT} T2.\rh{MailStreet}, T2.School \bt{FROM} satscores \bt{AS} T1 \bt{INNER JOIN} schools \bt{AS} T2 \bt{ON} T1.cds = T2.CDSCode \bt{WHERE} T1.AvgScrRead \bt{IS NOT NULL ORDER BY} T1.AvgScrRead \bt{ASC LIMIT 1} & \raggedright \bt{SELECT} schools.\gh{street}, satscores.sname \bt{FROM} satscores \bt{INNER JOIN} schools \bt{ON} satscores.cds = schools.cdscode \bt{ORDER BY} satscores.avgscrread \bt{ASC LIMIT} 1; & The question ask for mailing street address, not the street address. \\
\midrule
\multicolumn{4}{c}{[B4] Hallucination} \\
\midrule
What is the average math score of the school with the lowest average score for all subjects, and \rh{in which county} is it located? EK: Average score for all subjects can be computed by AvgScrMath + AvgScrRead + AvgScrWrite & \raggedright \bt{SELECT} T1.AvgScrMath, \rh{T2.County} \bt{FROM} satscores \bt{AS} T1 \bt{INNER JOIN} \rh{schools \bt{AS} T2} \bt{ON} T1.cds = T2.CDSCode \bt{WHERE} T1.AvgScrMath \bt{IS NOT NULL ORDER BY} T1.AvgScrMath + T1.AvgScrRead + T1.AvgScrWrite \bt{ASC LIMIT} 1	& \raggedright \bt{SELECT} avgscrmath, \gh{county \bt{FROM} satscores} \bt{ORDER BY} avgscrmath + avgscrread + avgscrwrite \bt{ASC LIMIT} 1	& Country is a column of schools table. \\
\midrule
\multicolumn{4}{c}{[B5] Condition} \\
\midrule
Please list the zip code of all the charter \rh{schools in Fresno County Office of Education}.	EK: Charter schools refers to `Charter School (Y/N)` = 1 in the table fprm & \raggedright \bt{SELECT} T2.Zip \bt{FROM} frpm \bt{AS} T1 \bt{INNER JOIN} schools \bt{AS} T2 \bt{ON} T1.CDSCode = T2.CDSCode \bt{WHERE} \rh{T1.`District Name` = 'Fresno County Office of Education'} \bt{AND} T1.`Charter School (Y/N)` = 1 & \raggedright \bt{SELECT DISTINCT} schools.zip \bt{FROM} schools \bt{INNER JOIN} frpm \bt{ON} schools.cdscode = frpm.cdscode \bt{WHERE} \gh{schools.county = 'Fresno'} \bt{AND} frpm.`charter school (y/n)` = 1; & 'Fresno County Office of Education' is  a district name, the SQL mistakenly treats it as county name. \\
\midrule
\multicolumn{4}{c}{[B6] NULL/DISTINCT} \\
\midrule
\rh{How many molecules} have a triple bond type?	EK : triple bond refers to bond\_type = '\#'; & \raggedright \rh{\bt{SELECT COUNT(DISTINCT} T.molecule\_id\bt{)}} \bt{FROM} bond \bt{AS} T \bt{WHERE} T.bond\_type = '\#' & \raggedright \gh{\bt{SELECT COUNT(}*\bt{)}} \bt{FROM} bond \bt{WHERE} bond\_type = '\#'; & There may be multiple records for a molecule in the bond table.
\\ \midrule
\rh{Which school} in Contra Costa has the highest number of test takers? & \raggedright \bt{SELECT} \rh{sname} \bt{FROM} satscores \bt{WHERE} cname = 'Contra Costa' \bt{AND} \rh{sname \bt{IS NOT NULL}} \bt{ORDER BY} NumTstTakr \bt{DESC LIMIT} 1 & \raggedright \bt{SELECT} \gh{sname} \bt{FROM} satscores \bt{WHERE} cname = 'Contra Costa' \bt{ORDER BY} numtsttakr \bt{DESC LIMIT} 1 & Sname column may have missing values.
\\
\bottomrule
\end{tabular}
\end{adjustbox}
\caption{Sample errors and analysis for types in the Schema Linking category.}
\label{table:ErrorSampleSchema}
\end{table*}

\begin{table*}[t]
\centering
\begin{adjustbox}{max width=\textwidth}
\begin{tabular}{p{4.75cm} p{5.5cm} p{5.5cm} p{2.5cm}}
\toprule
\textbf{Sample Question with EK} & \textbf{Gold SQL} & \textbf{Predicted SQL} & \textbf{Error Analysis}\\
\midrule
\multicolumn{4}{c}{[C1] String/Number} \\
\midrule
List out the code for drivers who have \rh{nationality in America}. & \raggedright \bt{SELECT} code \bt{FROM} drivers \bt{WHERE} \rh{Nationality = 'American'} & \raggedright \bt{SELECT} code \bt{FROM} drivers \bt{WHERE} \gh{nationality = 'America'}; & Although the question mentions 'America', in the nationality column, the corresponding entity in database is 'American'.  \\
\midrule
\multicolumn{4}{c}{[C2] Date} \\
\midrule
How many percent of 'Gold' cards were \rh{issued prior to 1998?} EK: Percent of Gold = [ count(type = 'gold' and issued date < 1998) / count(all cards)] * 100\%
 & \raggedright \bt{SELECT CAST( SUM(}type = 'gold' \bt{AND} \rh{\bt{STRFTIME(}'\%Y', issued\bt{)} < '1998'\bt{)}} \bt{AS REAL)} * 100 / \bt{COUNT(}card\_id\bt{) FROM} card
 & \raggedright \bt{SELECT CAST( SUM( CASE WHEN} TYPE = 'gold' \bt{AND} \gh{issued < '1998'} \bt{THEN} 1 \bt{ELSE} 0 \bt{END) AS REAL)} * 100 / \bt{count(}card\_id\bt{) FROM} card
 & The issued date is in format of yyyy-mm-dd  \\
\bottomrule
\end{tabular}
\end{adjustbox}
\caption{Sample errors and analysis for types in the Value Retrieval category.}
\label{table:ErrorSampleValue}
\end{table*}

\begin{table*}[t]
\centering
\begin{adjustbox}{max width=\textwidth}
\begin{tabular}{p{4.75cm} p{5.5cm} p{5.5cm} p{2.5cm}}
\toprule
\textbf{Sample Question with EK} & \textbf{Gold SQL} & \textbf{Predicted SQL} & \textbf{Error Analysis}\\
\midrule
\multicolumn{4}{c}{[D1] Mathematical Formula} \\
\midrule
 For loan amount less than USD100,000, \rh{what is the percentage of accounts that is still running with no issue.} EK: Status = 'C' stands for running contract, ok so far; Percentage of accounts by condition = [(total(amount) \& condition) / (total amount)] * 100.
 & \raggedright \bt{SELECT} \rh{\bt{CAST(SUM(}status = 'C'\bt{) AS} REAL\bt{)} * 100 / \bt{COUNT(}account\_id\bt{)}} \bt{FROM} loan \bt{WHERE} amount < 100000
 & \raggedright \bt{SELECT} \gh{\bt{(SUM(CASE WHEN} amount < 100000 \bt{AND} status = 'C' \bt{THEN} amount \bt{ELSE} 0 \bt{END)} * 100.0 / \bt{SUM(CASE WHEN} status = 'C' \bt{THEN} amount \bt{ELSE} 0 \bt{END))}} \bt{AS} percentage \bt{FROM} loan
 & The SQL tries to answer: For accounts that is still running with no issue, what is the percentage of loan amount less than USD 100,000? \\
\midrule
\multicolumn{4}{c}{[D2] Aggregation} \\
\midrule
 \rh{How many owner and disponent} dispositions are there from account number 1 to account number 10?  
 & \raggedright \bt{SELECT} \rh{\bt{SUM(}type = 'OWNER'\bt{)} ,\bt{ SUM(}type = 'DISPONENT'\bt{)}} \bt{FROM} disp \bt{WHERE} account\_id \bt{BETWEEN} 1 \bt{AND} 10
 & \raggedright \bt{SELECT} \gh{\bt{COUNT(}*\bt{)}} \bt{FROM} disp \bt{WHERE} account\_id \bt{BETWEEN} 1 \bt{AND} 10 \bt{AND} \gh{type \bt{IN (}'OWNER', 'DISPONENT'\bt{)}}
 &  Should count owener and disponent seperately. \\
\midrule
\multicolumn{4}{c}{[D3] Complex Operation} \\
\midrule
 Which molecule does the atom TR001\_10 belong to? \rh{Please state whether this molecule is carcinogenic or not.} EK: TR001\_10 is the atom id; label = '+' mean molecules are carcinogenic
 & \raggedright \bt{SELECT} T2.molecule\_id , \rh{IIF(T2.label = '+', 'YES', 'NO')} \bt{AS} flag\_carcinogenic \bt{FROM} atom \bt{AS} T1 \bt{INNER JOIN} molecule \bt{AS} T2 \bt{ON} T1.molecule\_id = T2.molecule\_id \bt{WHERE} T1.atom\_id = 'TR001\_10'
 & \raggedright \bt{SELECT DISTINCT} m.label \bt{FROM} molecule m \bt{INNER JOIN} atom a \bt{ON} m.molecule\_id = a.molecule\_id \bt{WHERE} a.atom\_id = 'TR001\_10'
 & The question ask for a judgement, which is more complex than ordinary query. \\

\bottomrule
\end{tabular}
\end{adjustbox}
\caption{Sample errors and analysis for types in the Operation category.}
\label{table:ErrorSampleOperation}
\end{table*}

\begin{table*}[t]
\centering
\begin{adjustbox}{max width=\textwidth}
\begin{tabular}{p{4.75cm} p{5.5cm} p{5.5cm} p{2.5cm}}
\toprule
\textbf{Sample Question with EK} & \textbf{Gold SQL} & \textbf{Predicted SQL} & \textbf{Error Analysis}\\
\midrule
\multicolumn{4}{c}{[E1] Redundant/Incomplete} \\
\midrule
 What are the \rh{labels for TR000, TR001 and TR002}? EK: TR000, TR001 and TR002 are molecule id;
 & \raggedright \bt{SELECT} \rh{molecule\_id, T.label} \bt{FROM} molecule \bt{AS} T \bt{WHERE} T.molecule\_id \bt{IN (}'TR000', 'TR001', 'TR002'\bt{)}
 & \raggedright \bt{SELECT} \gh{label} \bt{FROM} molecule \bt{WHERE} molecule\_id \bt{IN (}'TR000', 'TR001', 'TR002'\bt{)};
 &  Although the question asks for labels, without id, one can not tell them apart. \\
\midrule
\multicolumn{4}{c}{[E2] Column Sequence} \\
\midrule
 What are the \rh{elements} of the toxicology \rh{and label} of molecule TR060? EK: TR060 is the molecule id; 
 & \raggedright \bt{SELECT DISTINCT} \rh{T1.element, T2.label} \bt{FROM} atom \bt{AS} T1 \bt{INNER JOIN} molecule \bt{AS} T2 \bt{ON} T1.molecule\_id = T2.molecule\_id \bt{WHERE} T2.molecule\_id = 'TR060'
 & \raggedright \bt{SELECT DISTINCT} \gh{m.label, a.element} \bt{FROM} molecule m \bt{JOIN} atom a \bt{ON} m.molecule\_id = a.molecule\_id \bt{WHERE} m.molecule\_id = 'TR060';
 & Label should come after element as required. \\
\midrule
\multicolumn{4}{c}{[E3] ORDER BY/LIMIT} \\
\midrule
 For the \rh{first client} who opened his/her account in Prague, what is his/her account ID? EK: A3 stands for region names
 & \raggedright \bt{SELECT} T1.account\_id \bt{FROM} account \bt{AS} T1 \bt{INNER JOIN} district \bt{AS} T2 \bt{ON} T1.district\_id = T2.district\_id \bt{WHERE} T2.A3 = 'Prague' \rh{\bt{ORDER BY} T1.date \bt{ASC LIMIT} 1}
 & \raggedright \bt{SELECT} account.account\_id \bt{FROM} account \bt{INNER JOIN} district \bt{ON} account.district\_id = district.district\_id \bt{WHERE} district.a3 = 'Prague' \gh{\bt{LIMIT} 1}
 & First client refers to the account with smallest timestamp. \\
\midrule
\multicolumn{4}{c}{[E4] Format} \\
\midrule
 Name the ID of the patient who is born on the April 1st, 1982. \rh{Is his/her alkaliphophatase (ALP) within normal range}? EK: alkaliphophatase (ALP) within normal range refers to ALP < 300
 & \raggedright \bt{SELECT} T1.ID , \rh{\bt{CASE WHEN} T2.ALP < 300 \bt{THEN} 'normal' \bt{ELSE} 'abNormal' \bt{END}} \bt{FROM} Patient \bt{AS} T1 \bt{INNER JOIN} Laboratory \bt{AS} T2 \bt{ON} T1.ID = T2.ID \bt{WHERE} T1.Birthday = '1982-04-01'
 & \raggedright \bt{SELECT} \gh{\bt{CASE WHEN} laboratory.alp < 300 \bt{THEN} 'YES' \bt{ELSE} 'NO' \bt{END}}, patient.id \bt{FROM} patient \bt{INNER JOIN} laboratory \bt{ON} patient.id = laboratory.id \bt{WHERE} patient.birthday = '1982-04-01'
 & The desired answer is 'normal' or 'abNormal'. \\

\bottomrule
\end{tabular}
\end{adjustbox}
\caption{Sample errors and analysis for types in the Information category.}
\label{table:ErrorSampleInfo}
\end{table*}

\begin{table*}[t]
\centering
\begin{adjustbox}{max width=\textwidth}
\begin{tabular}{p{4.75cm} p{5.5cm} p{5.5cm} p{2.5cm}}
\toprule
\textbf{Sample Question with EK} & \textbf{Gold SQL} & \textbf{Predicted SQL} & \textbf{Error Analysis}\\
\midrule
\multicolumn{4}{c}{[F1] Syntax} \\
\midrule
 For the owner user of \rh{post} No. 65041, what is his/her reputation points? EK: post no. 65041 refers to Id = '65041'; reputation point refers to Reputation
 & \raggedright \bt{SELECT} T1.Reputation \bt{FROM} users \bt{AS} T1 \bt{INNER JOIN} \rh{posts} \bt{AS} T2 \bt{ON} T1.Id = T2.OwnerUserId \bt{WHERE} T2.Id = 65041
 & \raggedright \bt{SELECT} users.reputation \bt{FROM} users \bt{INNER} \gh{\bt{JOIN}\_posts} \bt{ON} users.id =\gh{\_posts}.owneruserid \gh{\bt{WHERE}\_posts}.id = 65041
 & No seperation between keywords and the post table.  \\
\bottomrule
\end{tabular}
\end{adjustbox}
\caption{Sample errors and analysis for types in the Syntax Error category.}
\label{table:ErrorSampleSyntax}
\end{table*}

%% file: Tables/ErrorAnalysisRanking.tex
\definecolor{darkgreen}{RGB}{0,150,0}
\begin{table*}[t]
\centering
\begin{adjustbox}{max width=\textwidth}
\begin{tabular}{ l | c | c | c }
\toprule

\textbf{Type}
& \makecell{\textbf{Vanilla DPO} \\ \textbf{Fix (\%)}} 
& \makecell{\textbf{Syn CoT DPO} \\ \textbf{Fix (\%)}}
& \textbf{$\Delta$(\%)$\downarrow$}  \\
\midrule
    \textbf{[E2] Column Sequence}
    & 0 (0/5)
    & \textbf{42.9 (3/7)}
    & \textbf{\textcolor{red}{+42.9}}
    \\
    \textbf{[A1] EK} 
    & 0.0 (0/3)
    & \textbf{37.5 (3/8)} 
    & \textbf{\textcolor{red}{+37.5}}
    \\
    \textbf{[B6] NULL/DISTINCT}
    & 9.7 (3/31)
    & \textbf{40.0 (12/30)}
    & \textbf{\textcolor{red}{+30.3}}
    \\
    \text{[C1] String/Number} 
    & 4.5 (1/22) 
    & 21.1 (4/19) 
    & \textbf{\textcolor{red}{+16.6}}
    \\
    \textbf{[B2] JOIN}
    & 15.6 (12/77)
    & \textbf{32.1 (18/56)}
    & \textbf{\textcolor{red}{+16.5}} 
    \\
    \text{[D2] Aggregation}
    & 6.7 (5/75)
    & 18.2 (12/66) 
    & \textbf{\textcolor{red}{+11.5}}
    \\
    \text{[E1] Redundant/Incomplete} 
    & 11.8 (4/34) 
    & 19.2 (5/26) 
    & \textbf{\textcolor{red}{+7.4}}
    \\
    \textbf{[C2] Date}
    & 23.1 (6/26)
    & \textbf{30.4 (7/23)}
    & \textbf{\textcolor{red}{+7.3}}
    \\
    \text{\underline{[D3] Complex Operation}}
    & 5.6 (1/18)
    & \underline{12.5 (3/24)}
    & \textbf{\textcolor{red}{+6.9}}
    \\
    \text{[B5] Condition}
    & 16.7 (10/60)
    & 23.2 (16/69) 
    & \textbf{\textcolor{red}{+6.5}}
    \\
    \text{\underline{[B3] Column}}
    & 10.3 (7/68)
    & \underline{16.1 (10/62)}
    & \textbf{\textcolor{red}{+5.8}}
    \\
    \text{[D1] Mathematical Formula} 
    & 13.3 (6/45)
    & 18.2 (8/44) 
    & \textbf{\textcolor{red}{+4.9}}
    \\
    \textbf{[B4] Hallucination}
    & 23.7 (14/59)
    & \textbf{27.2 (28/102)}
    & \textbf{\textcolor{red}{+3.5}} 
    \\
    \text{\underline{[E3] ORDER BY/LIMIT}}
    & 9.1 (1/11)
    & \underline{12.5 (1/8)}
    & \textbf{\textcolor{red}{+3.4}}
    \\
    \text{\underline{[B1] Table}}
    & 13.0 (12/92) 
    & \underline{15.9 (11/69)}
    & \textbf{\textcolor{red}{+2.9}}
    \\
    \text{\underline{[F1] Syntax}}
    & 14.3 (2/14) 
    & \underline{14.3 (2/14)}
    & \textbf{\textcolor{darkgreen}{-1.0}}
    \\
    \textbf{[E4] Format}
    & 66.7 (2/3)
    & \textbf{33.3 (2/6)}
    & \textbf{\textcolor{darkgreen}{-33.4}}
    \\
\bottomrule
\end{tabular}
\end{adjustbox}
\caption{Comparison of Vanilla and Syn CoT DPO correction capability across error types on Bird development set (greedy), results are arranged in descending order of fix rate difference. Base model is Qwen2.5-7B-Instruct. For Syn CoT, Error types with fix rates over $25\%$ are bolded, and error types with the five lowest fix rates are underlined.}
\label{tab:fixRateRank}
\end{table*}

%% file: Tables/ErrorAnalysisNewError.tex
\definecolor{darkgreen}{RGB}{0,150,0}
\begin{table*}[t]
\centering
\begin{adjustbox}{max width=\textwidth}
\begin{tabular}{c p{3.5cm} l | c | c | c | c }
\toprule
\textbf{Category} & \textbf{Description} & \textbf{Type} & \multicolumn{2}{c}{\makecell{\textbf{Vanilla DPO} \\ \textbf{New Error (\#)}}} \vline & \multicolumn{2}{c}{\makecell{\textbf{Syn CoT DPO} \\ \textbf{New Error (\#)}}} \\
\midrule
\multirow{1}{*}{External Knowledge} 
    & Neglect of hints
    & [A1] EK 
    & 0
    & 0 
    & 1
    & 1 \\
\midrule
\multirow{6}{*}{Schema Linking} 
    & \multirow{6}{=}{Fails to match the question with its concerning table and columns}
    & [B1] Table 
    & \multirow{6}{*}{32}
    & 9
    & \multirow{6}{*}{44}
    & 5 
    \\
    & & [B2] JOIN &
    & 5 &
    & 6
    \\
    & & [B3] Column &
    & 5 &
    & 7
    \\
    & & [B4] Hallucination &
    & 7 &
    & 19
    \\
    & & [B5] Condition &
    & 6 &
    & 3 
    \\
    & & [B6] NULL/DISTINCT &
    & 0 &
    & 4
    \\
\midrule
\multirow{2}{*}{Value Retrieval}
    & \multirow{2}{=}{Mismatch of condition with its storage format}
    & [C1] String/Number 
    & \multirow{2}{*}{5}
    & 2
    & \multirow{2}{*}{4}
    & 1
    \\
    & & [C2] Date &
    & 3 &
    & 3
    \\
\midrule
\multirow{3}{*}{Operation}
    & \multirow{3}{=}{Misunderstands required operation in the question.}
    & [D1] Mathematical Formula 
    & \multirow{3}{*}{15}
    & 4
    & \multirow{3}{*}{14}
    & 3
    \\
    & & [D2] Aggregation &
    & 10 &
    & 9 
    \\
    & & [D3] Complex Operation &
    & 1 &
    & 5 
    \\
\midrule
\multirow{4}{*}{Information}
    & \multirow{4}{=}{Fails to organize information in the right way}
    & [E1] Redundant/Incomplete 
    & \multirow{4}{*}{8}
    & 4 
    & \multirow{4}{*}{7}
    & 6 
    \\
    & & [E2] Column Sequence &
    & 2 &
    & 0
    \\
    & & [E3] ORDER BY/LIMIT &
    & 0 &
    & 1 
    \\
    & & [E4] Format &
    & 2 &
    & 0
    \\
\midrule
Syntax Error 
    & Inexecutatble SQL
    & [F1] Syntax 
    & 2
    & 2
    & 2
    & 2
    \\
\bottomrule
\end{tabular}
\end{adjustbox}
\caption{Comparison of Vanilla and Syn CoT new emerging errors in DPO phase across error categories and types on Bird development set (greedy). The base model is Qwen2.5-7B-Instruct.}
\label{tab:newErrorStat}
\end{table*}

%% file: Tables/ExampleHackingPatterns.tex
\begin{table*}[htbp]
\centering
\begin{tcolorbox}[
    colback=white, 
    colframe=black, 
    title=Reward Hacking Example, 
    fonttitle=\bfseries, 
    width=\textwidth, 
    boxrule=1pt, 
    rounded corners, 
]

\begin{tabularx}{\textwidth}{X}
\textbf{Question:} Among the superheroes with the race of god/eternal, how many of them are male?\\ 
\textbf{Gold SQL:} SELECT COUNT(*) FROM superhero AS T1 INNER JOIN race AS T2 ON T1.race\_id = T2.id INNER JOIN gender AS T3 ON T3.id = T1.gender\_id WHERE T1.race\_id = 21 AND T1.gender\_id = 1 \\
\hline
\\ \textbf{Response:} SELECT count(superhero.id) FROM gender INNER JOIN superhero ON gender.id = superhero.gender\_id WHERE{\color{red}\begin{CJK}{UTF8}{gbsn}英雄种族为\end{CJK}}god/eternal AND{\color{red}\begin{CJK}{UTF8}{gbsn}英雄性别为男\end{CJK}}
\end{tabularx}
\end{tcolorbox}
\caption{DPO model uses Chinese to state the conditions (Note that there are no Chinese character included in training IOs), which cause syntax errors. Interestingly, \begin{CJK}{UTF8}{gbsn}英雄种族为\end{CJK} means ``the race of superhero is'', and \begin{CJK}{UTF8}{gbsn}英雄性别为男\end{CJK} means ``the gender of superhero is male'', which are semantically correct. Similar cases are common, and the alternative language is not restricted to Chinese, we also see Farsi and Egyptian in the generation.}
\label{tab:egRH1}
\end{table*}

\begin{table*}[htbp]
\centering
\begin{tcolorbox}[
    colback=white, 
    colframe=black, 
    title=Reward Hacking Example, 
    fonttitle=\bfseries, 
    width=\textwidth, 
    boxrule=1pt, 
    rounded corners, 
]

\begin{tabularx}{\textwidth}{X}
\textbf{Question:} Calculate the average height of all neutral superheroes.\\ 
\textbf{Gold SQL:} SELECT AVG(T1.height\_cm) FROM superhero AS T1 INNER JOIN alignment AS T2 ON T1.alignment\_id = T2.id WHERE T2.alignment = 'Neutral' \\
\hline
\\ \textbf{Response:} SELECT avg(superhero.height\_cm) FROM alignment INNER JOIN superhero ON alignment.idossalhero.alignment\_id WHERE {\color{red}alignmentalignmentalignmentalignmentalignmentalignmentalignmentalignmentalignmentalignmentalignmenta}...
\end{tabularx}
\end{tcolorbox}
\caption{DPO model starts to repeat certain word (``alignment'' in this case), which is a common reward hacking pattern when DPO is used as alignment algorithm.}
\label{tab:egRH2}
\end{table*}

\begin{table*}[htbp]
\centering
\begin{tcolorbox}[
    colback=white, 
    colframe=black, 
    title=Reward Hacking Example, 
    fonttitle=\bfseries, 
    width=\textwidth, 
    boxrule=1pt, 
    rounded corners, 
]

\begin{tabularx}{\textwidth}{X}
\textbf{Question:} For the patient with the highest lactate dehydrogenase in the normal range, when was his or her data first recorded? \\ 
\textbf{Gold SQL:} SELECT T1.`First Date` FROM Patient AS T1 INNER JOIN Laboratory AS T2 ON T1.ID = T2.ID WHERE T2.LDH < 500 ORDER BY T2.LDH ASC LIMIT 1 \\
\hline
\\ \textbf{Response:} SELECT min(patient.`first date`) FROM patient INNER JOIN laboratory ON patient.id = laboratory.id WHERE laboratory.ldh < 500 OR {\color{red} (ldh IS NULL AND NOT AND ISnumeric(laboratory.ldh))}
\end{tabularx}
\end{tcolorbox}
\caption{DPO model struggles to produce SQLs with complete structure (e.g. unmatched parentheses, missing spaces, wrong usage of keywords), which is another common reward hacking pattern.}
\label{tab:egRH3}
\end{table*}